%% file: main.tex
\documentclass[lettersize, journal]{IEEEtran}
\PassOptionsToPackage{table}{xcolor}

\usepackage{cite}
\usepackage{graphicx}
\usepackage{textcomp}
\usepackage[sort,compress,numbers]{natbib}
\usepackage[hyphens]{url}  
\usepackage{xcolor}
\usepackage{subcaption}
\usepackage{wrapfig}
\usepackage{multirow}
\usepackage{rotating}
\usepackage{float}
\usepackage{xspace}
\usepackage{nicematrix}
\usepackage{makecell}
\usepackage{nicefrac}
\usepackage{enumitem}
\usepackage{booktabs}
\usepackage{microtype}      
\usepackage{amsmath}
\usepackage{hyperref}
\usepackage{amssymb}
\usepackage{amsfonts}
\usepackage{bbm}
\usepackage{tikz}
\usepackage{pgfplots}
\usepackage{amsthm}
\usepackage{multirow}
\usepackage{colortbl} 
\usepackage{arydshln}
\usepackage{siunitx}
\usepackage{cleveref}
\usepackage{arydshln}
\usepackage{amssymb}

\usepackage{comment}
\usepackage{pgfplots}
\usetikzlibrary{
        patterns,
        pgfplots.fillbetween,
    }
    \pgfplotsset{
        compat=1.15,
    }
\usepackage{bm}

\usepackage{algorithm}
\usepackage{algorithmic}

\input{packages}
\input{commands}

\Crefname{algocf}{Algorithm}{Algorithms}

\begin{document}

\title{AttackBench: Evaluating Gradient-based Attacks for Adversarial Examples}

\author{
\IEEEauthorblockN{Antonio Emanuele Cinà\IEEEauthorrefmark{1}\textsuperscript{\textsection}, 
Jérôme Rony\IEEEauthorrefmark{2}\textsuperscript{\textsection}, 
Maura Pintor\IEEEauthorrefmark{3}\IEEEauthorrefmark{10},
Luca Demetrio\IEEEauthorrefmark{1}\IEEEauthorrefmark{10},\\
Ambra Demontis\IEEEauthorrefmark{3},
Battista Biggio\IEEEauthorrefmark{3}\IEEEauthorrefmark{10},
Ismail Ben Ayed\IEEEauthorrefmark{2},
Fabio Roli\IEEEauthorrefmark{1}\IEEEauthorrefmark{3}}

\IEEEauthorblockA{\IEEEauthorrefmark{1}University of Genoa} 
\IEEEauthorblockA{\IEEEauthorrefmark{2}ÉTS Montréal}
\IEEEauthorblockA{\IEEEauthorrefmark{3}University of Cagliari 
}
\IEEEauthorblockA{\IEEEauthorrefmark{10}Pluribus One} 

Email: \IEEEauthorrefmark{1} antonio.cina@unige.it,  \IEEEauthorrefmark{2} jerome.rony.1@etsmtl.net,
}

\maketitle
\begingroup\renewcommand\thefootnote{\textsection}
\footnotetext{Equal contribution}
\endgroup

\begin{abstract}
Adversarial examples are typically optimized with gradient-based attacks. While novel attacks are continuously proposed, each is shown to outperform its predecessors using different experimental setups, hyperparameter settings, and number of forward and backward calls to the target models. This provides overly-optimistic and even biased evaluations that may unfairly favor one particular attack over the others. 
In this work, we aim to overcome these limitations by proposing AttackBench, \ie the first evaluation framework that enables a fair comparison among different attacks. To this end, we first propose a categorization of gradient-based attacks, identifying their main components and differences.
We then introduce our framework, which evaluates their effectiveness and efficiency. We measure these characteristics by (i) defining an \textit{optimality} metric that quantifies how close an attack is to the optimal solution, and (ii) limiting the number of forward and backward queries to the model, such that all attacks are compared within a given maximum query budget.
Our extensive experimental analysis compares more than $100$ attack implementations with a total of over $800$ different configurations against CIFAR-10 and ImageNet models, highlighting that only very few attacks outperform all the competing approaches.
Within this analysis, we shed light on several implementation issues that prevent many attacks from finding better solutions or running at all. We release AttackBench as a publicly-available benchmark, aiming to continuously update it to include and evaluate novel gradient-based attacks for optimizing adversarial examples.
\end{abstract}

\begin{IEEEkeywords}
Evasion attacks, benchmark, computer vision
\end{IEEEkeywords}

\input{introduction}
\input{gradient-attacks}

\input{attackbench}

\input{experiments}
\input{related}
\input{conclusions}

\section*{Acknowledgments} This work has been partly supported by the EU-funded Horizon Europe projects ELSA (GA no. 101070617) and Sec4AI4Sec (GA no. 101120393); by project TESTABLE (GA no. 101019206); by project SERICS (PE00000014) under the MUR National Recovery and Resilience Plan funded by the European Union - NextGenerationEU; and by European Union—NextGenerationEU (National Sustainable Mobility Center CN00000023, Italian Ministry of University and Research Decree n. 1033—17/06/2022, Spoke 10).

\bibliographystyle{IEEEtranN}
\bibliography{biblio}
\input{bio}
\clearpage
\appendix
\input{appendix/appendix}

\end{document}

%% file: packages.tex
\usepackage{cite}
\usepackage{graphicx}
\usepackage{textcomp}
\usepackage{xcolor}
\usepackage{subcaption}

\usepackage[algo2e,noend,linesnumbered,ruled,vlined]{algorithm2e}
\usepackage{mathtools}
\usepackage{multirow}
\usepackage{rotating}
\usepackage{float}
\usepackage{xspace}
\usepackage{nicematrix}
\usepackage{makecell}
\usepackage{nicefrac}
\usepackage{enumitem}
\usepackage{booktabs}
\usepackage{microtype}      
\usepackage{amsmath}
\usepackage{amssymb}
\usepackage{amsfonts}
\usepackage{tikz}
\usepackage{pgfplots}
\usepackage{amsthm}
\usepackage{multirow}
\usepackage{colortbl} 
\usepackage{arydshln}
\usepackage{siunitx}
\usepackage{cleveref}
\usepackage{arydshln}
\usepackage{amssymb}
\usepackage{comment}
\usepgfplotslibrary{fillbetween} 
\usetikzlibrary{patterns}

%% file: commands.tex
\newcommand{\vct}[1]{\ensuremath{\boldsymbol{\mathbf{#1}}}}

\newcommand{\set}[1]{\ensuremath{\mathcal{#1}}}
\newcommand{\con}[1]{\ensuremath{\mathsf{#1}}}

\DeclareMathOperator*{\argmax}{arg\,max}
\DeclareMathOperator*{\minimize}{minimize}

\DeclareMathOperator*{\subjectto}{subject~to}

\newcommand{\myparagraph}[1]{\noindent \textbf{#1}}
\newcommand{\mysubparagraph}[1]{\noindent \textit{#1}}

\makeatletter
\DeclareRobustCommand\onedot{\futurelet\@let@token\@onedot}
\def\@onedot{\ifx\@let@token.\else.\null\fi\xspace}

\def\eg{\emph{e.g}\onedot} 
\def\ie{\emph{i.e}\onedot}

\def\etal{\emph{et al}\onedot}
\makeatother

\newcommand{\ellzero}{$\ell_0$\xspace}
\newcommand{\ellone}{$\ell_1$\xspace}
\newcommand{\elltwo}{$\ell_2$\xspace}
\newcommand{\ellinf}{$\ell_{\infty}$\xspace}

\newcommand{\cw}{CW\xspace}
\newcommand{\pgd}{PGD\xspace}
\newcommand{\bb}{BB\xspace}
\newcommand{\fab}{FAB\xspace}
\newcommand{\ddn}{DDN\xspace}
\newcommand{\fmn}{FMN\xspace}
\newcommand{\alma}{ALMA\xspace}
\newcommand{\apgd}{APGD\xspace}
\newcommand{\apgdlone}{APGD-\ensuremath{\ell_1}\xspace}

\newcommand{\apgdt}{APGD$_t$\xspace}
\newcommand{\ead}{EAD\xspace}
\newcommand{\sfool}{SparseFool\xspace}
\newcommand{\dfool}{DeepFool\xspace}
\newcommand{\bim}{BIM\xspace}

\newcommand{\jsma}{JSMA\xspace}
\newcommand{\fgsm}{FGSM\xspace}
\newcommand{\fgm}{FGM\xspace}
\newcommand{\tr}{TrustRegion\xspace}
\newcommand{\pdgd}{PDGD\xspace}
\newcommand{\pdpgd}{PDPGD\xspace}
\newcommand{\vfga}{VFGA\xspace}
\newcommand{\deepfool}{\dfool}
\newcommand{\sigmazero}{\ensuremath{\sigma}-zero\xspace}
\newcommand{\pgdlzero}{PGD-$\ell_0$\xspace}

\newcommand{\crossentropy}{\texttt{NCE}\xspace}
\newcommand{\logitloss}{\texttt{DL}\xspace}
\newcommand{\dlr}{\texttt{DLR}\xspace}

\newcommand{\maxloss}{\texttt{FixedBudget}\xspace}
\newcommand{\minnorm}{\texttt{MinNorm}\xspace}

\newcommand{\ab}{AttackBench\xspace}
\newcommand{\areacurve}[1]{\ensuremath{\text{AUREC}_{#1}(\epszero)}\xspace}
\newcommand{\attack}{\ensuremath{a}\xspace}
\newcommand{\oneattack}{\ensuremath{\attack^{i}}\xspace}
\newcommand{\bestattack}{\ensuremath{\attack^{\star}}\xspace}
\newcommand{\pertsize}{\ensuremath{\varepsilon}\xspace}
\newcommand{\epszero}{\ensuremath{\pertsize_{0}}\xspace}
\newcommand{\ellp}{$\ell_p$\xspace}

\newcommand{\nattacks}{102\xspace}
\newcommand{\nalgo}{20\xspace}
\newcommand{\nbugs}{5\xspace}

\newcommand{\accuracy}{\ensuremath{\rho}\xspace}
\newcommand{\budgetsize}{\ensuremath{\varepsilon}\xspace}
\newcommand{\robustaccuracy}{\ensuremath{\accuracy(\pertsize)}\xspace}
\newcommand{\attackindex}{\ensuremath{i}}
\newcommand{\Aopt}{{\ensuremath{\accuracy^{\ast}_{\vct\theta}(\epsilon)}}\xspace}
\newcommand{\asr}{\ensuremath{\text{ASR}}}

\newcommand{\lOptimality}{\ensuremath{\xi^\attackindex_{\vct\theta}}\xspace}
\newcommand{\lOptimalityIndex}[1]{\ensuremath{\xi^{#1}_{\vct\theta}}\xspace}
\newcommand{\lOptimalitym}{\ensuremath{\xi^\attackindex_{\vct\theta_m}}\xspace}
\newcommand{\gOptimality}{\ensuremath{\xi^\attackindex}\xspace}

\newcommand{\lo}{\textit{LO}\xspace}
\newcommand{\go}{\textit{GO}\xspace}

\newcommand{\tcheck}{\checkmark}

\theoremstyle{definition}

\newcommand{\standardcifar}{C1\xspace}
\newcommand{\zhang}{C2\xspace}
\newcommand{\stutz}{C3\xspace}
\newcommand{\xiao}{C4\xspace}
\newcommand{\wang}{C5\xspace}

\newcommand{\standardimagenet}{I1\xspace}
\newcommand{\wong}{I2\xspace}
\newcommand{\salman}{I3\xspace}
\newcommand{\debenedetti}{I4\xspace}

\newcommand{\advlibshort}{{\small{AL}}}
\newcommand{\originalshort}{{\small{O}}}
\newcommand{\foolboxshort}{{\small{FB}}}
\newcommand{\artshort}{{\small{Art}}}
\newcommand{\cleverhansshort}{{\small{CH}}}
\newcommand{\torchattacksshort}{{\small{TA}}}
\newcommand{\deeprobustshort}{{\small{DR}}}

%% file: introduction.tex
\section{Introduction}
In recent years, we have been witnessing a proliferation of gradient-based attacks, aiming to constantly improve efficacy and efficiency while optimizing adversarial examples~\cite{carlini17-sp,chen2018ead,rony2019decoupling,pintor2021fast,Cina2024SigmaZero}.
Despite these advancements, the comparisons between each newly-proposed attack and its predecessors have been systematically conducted using different experimental settings, i.e., (i) models and performance metrics, (ii) attack implementations, and (iii) computational budgets,
hindering reproducibility and an overall fair comparison of current attack algorithms. 
In particular,
(i) many comparisons have considered a different set of target models, and used different metrics to evaluate the attack success, using either the success rate at a predefined perturbation budget (e.g., $\epsilon = \nicefrac{8}{255}$)~\cite{Croce2019SparseAI} or the median perturbation size required to mislead the model~\cite{brendel2020accurate,pintor2021fast}. 
However, these settings neither allow a direct comparison of the performance of different attacks across different papers nor tell anything about the effectiveness of the attacks at different perturbation budgets. 
Many attacks (ii) have been re-implemented without ensuring consistency with their original versions, and some implementations are even flawed~\cite{carlini2019critique}.
We will indeed demonstrate in this paper that different implementations of the same attack can yield significantly different results, affecting the performance of state-of-the-art methods.
Attacks (iii) are not consistently compared using the same number of queries (i.e., the number of forward and backward passes performed on the target model), biasing the evaluation in favor of more computationally-demanding methods.
Nevertheless, limiting the number of iterations of each attack is insufficient to solve this issue. 
The reason is that some attacks require two queries per iteration (one forward and one backward)~\cite{pintor2021fast,rony2019decoupling}, while others use additional subroutines like restarts~\cite{croce2020minimally,Croce2019SparseAI} or run internal hyperparameter optimizations~\cite{chen2018ead,carlini17-sp}, resulting in multiple queries per iteration.
These factors have led to either overestimating or underestimating the performance of certain attacks~\cite{pintor2022indicators,carlini2019critique}. Consequently, an in-depth investigation into the effectiveness of gradient-based attacks in a fair and reproducible experimental setting is missing. 

To address the above limitations, we propose \ab: a unified benchmark that evaluates adversarial attacks under a consistent setup and imposes a maximum computational budget for the attack (i.e., the maximum number of queries they can use). 
We design \ab as a five-stage procedure (see Fig.~\ref{fig:abstract_atkbench}) to discover the attacks that find the best minimally-perturbed adversarial examples with fewer queries to the models.
\ab ranks the attacks based on a novel metric, called \emph{optimality}, that quantifies how close each attack is to the empirically-optimal solution (estimated by ensembling all the attacks tested). 
The optimality metric evaluates the effectiveness of gradient-based attacks across various perturbation budgets, avoiding pointwise evaluations at fixed perturbation budgets, and promoting attacks that consistently find smaller perturbations to evade the target model. 
\ab also compares the attacks in their computational efficiency, measuring their runtime within the given maximum computational budget. 
We then use \ab to perform an extensive benchmark analysis that compares \nalgo attacks (listed in \autoref{table:adv_attacks_categorization}). 
For each attack, we consider all the implementations available among popular adversarial attack libraries and the original authors' code when available.
We empirically test a total of \nattacks techniques, re-evaluating them in terms of their runtime, success rate and perturbation size, as well as our newly introduced \emph{optimality} metric.
While developing our benchmark, we uncovered further insights, including suboptimal implementations and source code errors that prevent some attacks from completing their runs correctly.
To foster reproducible results, we open source \ab and will provide an online leaderboard, enabling researchers to easily assess the performance of newly-proposed attacks against competing strategies in a unified setting.

We summarize our contributions as follows: 
(i) we propose \ab, a fair benchmark for adversarial attacks that evaluates them on consistent models, data, and computational budgets, and introduce the novel \emph{optimality} metric to rank them based on the quality of the adversarial examples they generate;
(ii) we extensively test \nattacks attacks and we rank them according to our novel metric (Sec.\ref{sec:experiments}); and lastly (iii) we highlight \nbugs programming errors inside the code of some attacks we have considered, and we present inconsistent results of different implementation of the same attack (Sec.\ref{sec:experiments}).

%% file: gradient-attacks.tex
\section{Gradient-based Attacks}
\label{sec:gradient-attacks}
Let us assume, without loss of generality, that the input samples lie in a $\con d$-dimensional (bounded) space, \ie $\vct x \in [0, 1]^\con d$, and that their labels are denoted with $y \in \{1, ..., C\}$. Then, the predicted label of a trained model parameterized by $\vct \theta$ can be denoted with $\hat y = f(\vct x, \vct \theta)$, while the confidence value (logit) for class $c$ can be denoted with $f_c(\vct x, \vct \theta)$ and the softmax-rescaled logits with $z_c(\vct x, \vct \theta)$.
The predicted label can thus be computed also as $\hat y = \argmax_c f_c(\vct x, \vct \theta)$.
Under this setting, finding an adversarial example amounts to solving the following multi-objective optimization:
\begin{eqnarray}
    \label{eq:pareto} \minimize_{\vct \delta} &&  \left ( L(\vct x + \vct \delta, y; \vct \theta) ,  \| \vct \delta \|_p \right ) \, , \\
    \label{eq:constr_pareto}\subjectto && \vct x + \vct \delta \in [0,1]^\con d \, ,
\end{eqnarray}
where $L(\vct x + \vct \delta, y; \vct \theta)$ is a loss defining the misclassification objective, and $\vct \delta$ is the perturbation optimized to find an adversarial example $\vct x^\prime = \vct x+\vct \delta$ within the feasible domain.
The loss function $L: \mathbb{R}^\con d \times \mathbb{R} \to \mathbb{R}$ defines the objective of the attack so that $L$ is large when the input is correctly classified, whereas it is lower when the model predicts a wrong label for $\vct x$.\footnote{For targeted attacks, one can minimize $L(\vct x + \vct \delta, y_t; \vct \theta)$, being $y_t \neq y$ the label of the target class.}
The majority of existing attacks now leverage the Negative Cross-Entropy (\crossentropy) loss, the Difference of Logits (\logitloss)~\cite{carlini17-sp}, or the Difference of Logits Ratio (\dlr)~\cite{croce2020reliable}. 
The second objective in Eq.~\eqref{eq:pareto} is expressed as a constraint on the size of the perturbation $||\vct \delta||_p$, formulated through the usage of $\ell_p$ norms.
Typically, \ellzero, \ellone, \elltwo, and \ellinf norms are used, yielding sparse to increasingly dense adversarial perturbations.
The box constraint in Eq.~\eqref{eq:constr_pareto} ensures that the sample remains within the input space of the model, \ie $\vct x + \vct \delta \in [0, 1]^\con d$. 

The optimization problem expressed in Eq.~\eqref{eq:pareto} presents an inherent tradeoff: minimizing $L$ favors the computation of adversarial examples with large misclassification confidence, but also a large perturbation size, whereas minimizing $\| \vct \delta \|_p$ penalizes larger perturbations at the expense of decreasing the misclassification confidence.
We can thus define two main families of attacks, where one aims at finding the inputs that cause the maximum error within a given perturbation budget \pertsize (\maxloss)~\cite{biggio13-ecml,madry18-iclr}, and the counterpart that searches for the smallest perturbation needed to achieve misclassification (\minnorm)~\cite{szegedy14-iclr-intriguing,brendel2020accurate}.

\paragraph{Limitations of Attack Evaluation Metrics.} 
We discuss here the main metrics that are normally used to evaluate the effectiveness of \maxloss and \minnorm attacks.
For \maxloss attacks, the Attack Success Rate $\asr_a(\pertsize)$ is used to determine the success rate of attack $a$ when the perturbation budget is at most \pertsize~\cite{madry18-iclr,Croce2019SparseAI,Croce2021MindTB,croce2020minimally}. It is defined as:
\begin{equation}
     \asr_a(\pertsize)= \frac{1}{|\set D |} \sum_{(\vct x, y) \in \set D} \mathbb I( d_{\vct x} \leq \pertsize) \, ,
\end{equation}
where $d_{\vct x} = \| \vct x_{\rm adv} - \vct x\|_p$ is the perturbation size of the adversarial examples that successfully mislead the prediction of the target model, and $\mathbb I(\cdot)$ is the indicator function, which returns $1$ if its argument is true, and $0$ otherwise.

Evaluations for \minnorm attacks focus on the mean~\cite{carlini17-sp,chen2018ead} or median~\cite{brendel2020accurate,Cina2024SigmaZero} values of the adversarial perturbation. 
The median in particular indicates the perturbation budget at which 50\% of attacks are successful. 

Unfortunately, these metrics suffer from some limitations. In particular, (i) the $\asr_a$ and the median do not allow a direct comparison of the performance of \maxloss and \minnorm attacks.
Moreover, (ii) these metrics do not capture whether an attack is succeeding or not at larger perturbation budgets, which is instead useful to understand whether the attack is failing due, e.g., to implementation issues or for being ineffective against the specific model~\cite{carlini2019evaluating,pintor2022indicators}.
Finally, (iii) these metrics do not provide any insights on how close or far the results of each attack are from the best possible (though unknown) solution. While newly-proposed attacks are normally compared with a bunch of competing approaches, as we will see, these evaluations still fall short when it comes to assess the effectiveness of each attack with respect to the optimal solution.

%% file: attackbench.tex
\section{The AttackBench Framework}
\label{sec:attackbench}

\begin{figure*}[t]
    \centering
    \includegraphics[width=0.99\linewidth]{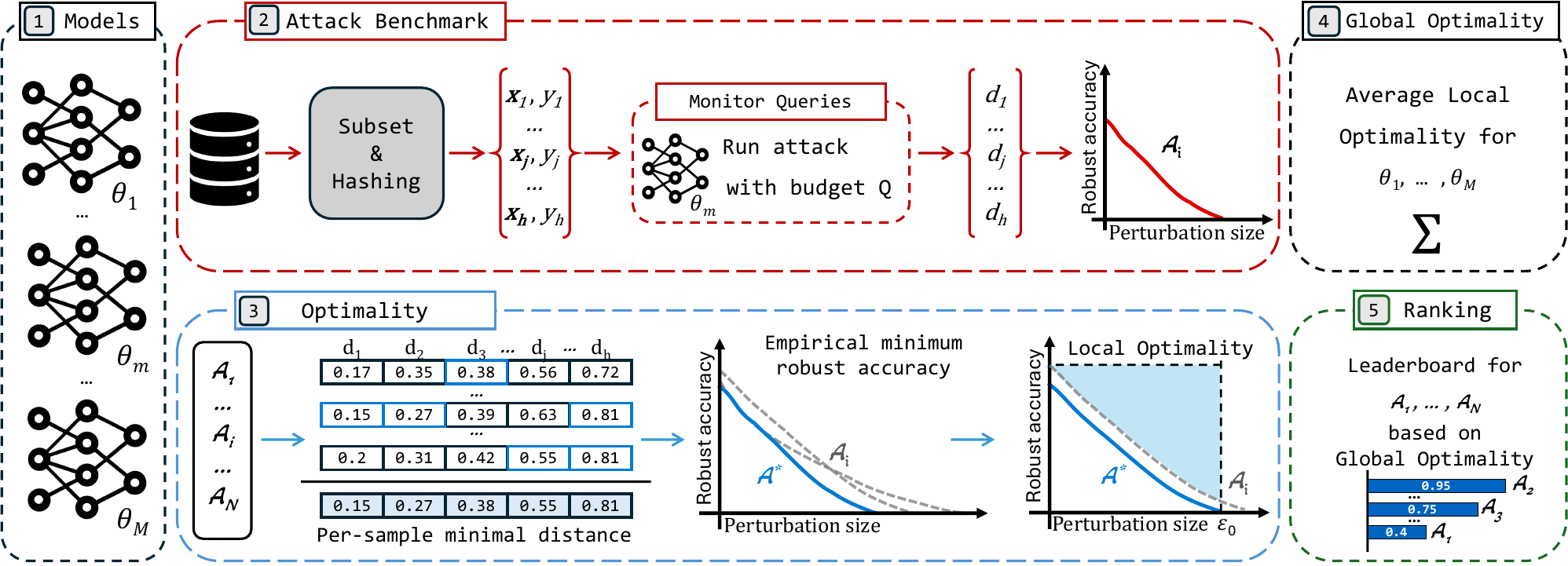}
    \caption{A comprehensive overview of the five stages of \ab. Each attack is tested in fair conditions and ranked through the optimality metric. The best attack is the one that produces minimally-perturbed adversarial examples with fewer queries.}
    \label{fig:abstract_atkbench}
\end{figure*}
We now describe the stages of \ab, depicted in \Cref{fig:abstract_atkbench}.
In \Cref{sec:testing_attacks}, we detail stages (1-2), which define the benchmark environment setup for executing attacks under standardized conditions.  
We then present in \Cref{sec:optimality} stages (3-4-5), where we aggregate results from stage (2) to compute a novel \textit{optimality} metric, as shown in stage (3) of \Cref{fig:abstract_atkbench}, for ranking gradient-based attacks. 
Lastly, we explain how to integrate novel attacks into the benchmark and provide implementation details in Appendix~\ref{appendix:implementation}.

\subsection{Model Zoo and Attack Benchmarking}\label{sec:testing_attacks}
\myparagraph{Stage (1): Model Zoo}. 
It builds a \textit{zoo} of diverse models, encompassing robust and non-robust models, to evaluate attacks across various scenarios and minimize evaluation biases. The pool can be clearly expanded as novel defenses are released.\smallskip\smallskip
\input{pseudocode/testattack}

\myparagraph{Stage (2): Attack Benchmarking}. It consists of testing each attack with \Cref{algo:atkbenchmark_code}, against all models in the \ab zoo.
For each attack $a$, target model $\vct \theta$, and dataset $\set D$, \ab returns the best minimum distances $d^\star$ found by $a$ for each sample, as well as the number of forward and backward queries $q^\star$ utilized.  
In particular, Algorithm~\ref{algo:atkbenchmark_code} begins by initializing the optimal minimum distances $d^\star$ and $q^\star$. 
It then wraps the target model $\vct \theta$ within a custom object $\texttt{B}_{\vct \theta}$, which tracks the number of queries—both forward and backward passes—used by the attack to optimize each sample (\Cref{alg:benchmodel}). 
The algorithm then iterates over the input samples $\vct x$ and their labels $y$ in $\set D$. 
The loop starts by setting the number of queries for the current sample to zero (\Cref{alg:init-queries}), and then hashes the sample (\Cref{alg:hashing}) to ensure consistent result storage, even if the samples are processed in a different order.
The attack is then executed (\Cref{alg:attack_step}), while tracking the number of queries made to the target model. 
If the query budget $Q$ is exceeded, the attack is halted. 
Whether the attack exceeds $Q$ or not, \ab always returns the best adversarial example $\vct x_{\rm adv}$ found during the optimization. 
This approach improves upon many existing attack implementations, which typically return only the last sample, even though it may not be the optimal one~\cite{pintor2022indicators}.
The algorithm then calculates the distance $d^\star$ between the adversarial sample $\vct{x}_{\rm adv}$ and the source sample $\vct{x}$ (\Cref{alg:distance}), and retrieves the number of queries $q^\star$ used to optimize $\vct{x}_{\rm adv}$ from $\texttt{B}_{\vct{\theta}}$ (\Cref{alg:get_query}). Finally, the results ($d^\star$, $q^\star$) for all samples are returned and used to compute the robustness evaluation curve. 
Remarkably, \ab enables the benchmarking of novel attacks without altering their code or adjusting their parameters, avoiding misconfiguration issues. 
\ab operates by wrapping the model and continuously monitoring all queries made by the attack to provide a comprehensive view of the attack's performance and resource usage.
Technical details regarding the implementation of stage (2) are reported Appendix~\ref{appendix:implementation}.

\subsection{Evaluating and Ranking Attacks}
\label{sec:optimality}
\myparagraph{Stage (3): Optimality Score.} 
\ab compares gradient-based attacks by inspecting the robustness evaluation curves~\cite{biggio14-tkde,biggio18wild} they generate against the target models in the \emph{zoo}.
These curves illustrate how robust accuracy $\rho_a(\pertsize)$, i.e., 1 - $\asr_a(\pertsize)$, decreases as the perturbation budget \pertsize for attack $a$ increases. 
Some examples are shown in \Cref{fig:robust_accuracy_curves_2}. 
Note that when \pertsize equals zero, $\rho_a(0)=\rho$ corresponds to the model's \emph{clean accuracy}, meaning the accuracy on unperturbed samples. 
More effective attacks typically produce robustness evaluation curves closer to the origin. 
The Area Under the Robustness Evaluation Curve (AUREC) can be defined as:
\begin{equation}
    \text{AUREC}_{\attack}(\pertsize_0) = \int_0^{\pertsize_0}\rho_a(\pertsize) \mathrm{d}\pertsize, 
    \label{eq:ausec}
\end{equation}
where $\pertsize_0$ denotes the upper bound of the integration interval, which should be set to $\infty$ or to the smallest $\pertsize$ at which $\rho_a(\pertsize)=0$ to compute the complete area. 
The AUREC can be used to summarize the average attack effectiveness on the whole robustness evaluation curve, as illustrated in \Cref{fig:robust_accuracy_curves_2}.

However, AUREC values obtained from different models are not directly comparable, as the models tend to exhibit different \emph{clean accuracy} values.
As the outcome of our benchmark's stage (3), we overcome this issue by proposing a novel metric, called \emph{local optimality} and denoted with \lOptimality. 
To this end, we first define the best (empirical) attack \bestattack, by ensembling all the attacks tested in our benchmark. In particular,
for each sample, this amounts to setting the best minimum distance $d^\star$ of \bestattack as the smallest distance found among \emph{all} the considered attacks (see \Cref{fig:abstract_atkbench}). As we will see in our experiments, this provides the best possible empirical estimate of the optimal solution for each sample.\footnote{Let us remark that we are ensembling over 100 attacks in \ab, whereas previous evaluations normally consider less than tens of competing approaches without even ensembling them.} 
We then set $\epszero$ as the smallest value at which $\rho_\bestattack(\pertsize) = 0$, and compute \areacurve{\bestattack} from the robustness evaluation curve of \bestattack.

\mysubparagraph{Local optimality (\lo).}
The \lo metric, denoted as \lOptimality, 
computes the difference between the area under the curve of \oneattack and \bestattack against the target $\vct{\theta}$ to evaluate how much the first is suboptimal with respect to the second. Essentially, \lo indicates the loss in performance of \oneattack compared to \bestattack. 
In \Cref{fig:robust_accuracy_curves_2}, we illustrate an example of the \lo measure when comparing the robustness evaluation curve of \oneattack (red curve) with that of the best attack, \bestattack (blue curve). 
The \lo metric quantifies the area of the gray region above the red curve in \Cref{fig:robust_accuracy_curves_2}, inversely proportional to the area between the two curves.
When the area between these two curves is non-zero, it indicates that \oneattack is less optimal than \bestattack. 
The adversarial examples generated by \oneattack thus are generally sub-optimal, meaning they succeed only with larger perturbations compared to the empirical lower bound found by \bestattack.
\input{figure_area}
Formally: 
\begin{equation}
    \lOptimality = \frac{\accuracy\cdot\epszero - \areacurve{\oneattack}}{\accuracy\cdot\epszero - \areacurve{\bestattack}}\,,
    \label{eq:local_optimality}
\end{equation}
where the box defined by $\accuracy \cdot \epszero$ serves to normalize the \lo measure with respect to the clean accuracy $\accuracy$ and robustness $\epszero$ of the target model $\vct \theta$. 
In more detail, when a target model exhibits high robustness (i.e., large $\epszero$) or accuracy (i.e., large $\accuracy$), evading the target model becomes inherently more challenging, leading to less severe penalties for the optimality of the attack. 
Conversely, when the target model is more vulnerable or less accurate, the curve disparity becomes more relevant, and the impact on the optimality of the attacks is more pronounced. 
Furthermore, this term ensures that the \lo measure is bounded in $[0, 1]$ with $\lOptimality=1$ when \oneattack performs exactly as \bestattack, thus finds always the smallest perturbation to evade the target model for each sample. 
By contrast, when $\lOptimality=0$, the attack fails to find a successful perturbation with $\pertsize \leq \varepsilon_0$, thus resulting in a curve that matches the box and therefore $\areacurve{\oneattack} = \accuracy\cdot \epszero$.
The \lo measure thus offers a clear scale to interpret and compare the performance of gradient-based attacks across different models.
Lastly, stage (3) of \ab uses a broad range of attacks to construct \bestattack and assess, with the \lo measure, how far an attack is from an ideal top performer across multiple perturbation budgets.\smallskip\smallskip

\myparagraph{Stage (4): Global Optimality.}
The \lo measure evaluates the effectiveness of an attack against a single target model, $\vct \theta_m$. 
However, ranking attacks with respect to a single model increases the risk of overfitting the benchmark. An attack might be specifically tailored for a particular model $\vct \theta_m$, but have poor generalization and lower performance across different models.
To this end, \ab prevents potential bias by evaluating the attacks against a diverse set of robust models and aggregating the results to derive a global score, namely \textit{global optimality}, as the output of step (4). 

\mysubparagraph{Global Optimality (\go).}
The \go measure quantifies the average \lo of an attack, denoted as \oneattack, concerning a set of target models $\mathcal{M}=\{\vct{\theta}_1,\dots, \vct\theta_M\}$. We define the \go as:
\begin{equation}
    \gOptimality = \frac{1}{|\set{M}|}\sum\limits_{\vct\theta_m \in \set{M}}\lOptimalitym\,.\label{eq:global_optimality}
\end{equation}
The \go score for an attack corresponds to the average \lo measure obtained across a set $\mathcal{M}$ of distinct models. Bounded in $[0, 1]$ as well, the \go score for an attack \oneattack equals $1$ only when \oneattack consistently identifies the minimal adversarial perturbations for all input samples across every model in $\mathcal{M}$, meaning \oneattack performs as well as \bestattack in every case (i.e., $\oneattack=\bestattack$). \smallskip

\myparagraph{Stage (5): Ranking.} Lastly, \ab ranks attacks by their \go, grouping them based on the $\ell_p$ norm threat model they consider and the target model.
Remarkably, the \ab \textit{optimality} ranking can be continuously and efficiently updated as new attacks emerge, without the need to re-run previous ones. 
When a new attack is uploaded, \ab updates the best minimum distances $d^\star$ of \bestattack and refreshes the leaderboard for all attacks by recalculating the \lo score employing the stored robustness evaluation curves (additional details in \Cref{appendix:new-attacks}).

\input{pseudocode/attackbenchmark}

%% file: pseudocode/testattack.tex
\begin{algorithm2e}[!t]
	\caption{Attack Benchmarking} \label{algo:atkbenchmark_code}
    \SetKwInOut{Input}{Input}
    \SetKwInOut{Output}{Output}
    \SetKwComment{Comment}{$\triangleright$\ }{}

    \SetAlgoLined
    \DontPrintSemicolon
   \Input{$\attack$, the attack algorithm; $\vct \theta$, the target model; $\set D$, the test dataset; $Q$, the query budget; and $p \in \{0,1,2,\infty\}$, the perturbation model.}
	\Output{Min. distances $d^{\star}$ and required queries $ q^\star$.}
    
    $d^{\star} \gets \texttt{\{\}}, \quad q^{\star} \gets \texttt{\{\}}$


    $\texttt{B}_{\vct \theta} \gets \texttt{BenchModel}(\vct \theta, Q)$ \label{alg:benchmodel}   
    
    \For {$(\vct x, y) \in \set{D}$}{

        $\texttt{B}_{\vct \theta}$.\texttt{init\_queries()}\label{alg:init-queries}
        
        $h_x \gets \texttt{hash}(\vct x)$\label{alg:hashing}
        
        $\vct x_{\rm adv} \gets \attack(\vct x, y, \texttt{B}_{\vct \theta})$ \label{alg:attack_step}

        $d^{\star}\{h_x\} \gets \|\vct x - \vct x_{\rm adv}\|_p $ \label{alg:distance}

        $q^{\star}\{h_x\} \gets \texttt{B}_{\vct \theta}.\texttt{num\_queries}()$\label{alg:get_query}

     }
    \bfseries return $d^{\star}, q^{\star}$
\end{algorithm2e}

%% file: figure_area.tex
\begin{figure}[t]
\pgfdeclareplotmark{mymark}
{%
\path[fill=white,postaction={pattern = north east lines, pattern color=red}] (-\pgfplotmarksize,-\pgfplotmarksize) rectangle (\pgfplotmarksize,\pgfplotmarksize);
}
    \centering
    \begin{tikzpicture}
        \begin{axis}[
            axis x line=middle, axis y line=middle, samples=1000,clip=false,
            xmin=0, xmax=5, domain=0:4.9, 
            ymin=0, ymax=1.05,
            xtick={0,...,4},
            xlabel=$\pertsize$, 
            ylabel=\robustaccuracy,
            x label style={at={(axis description cs:0.5,-0.1)},anchor=north},
            y label style={at={(axis description cs:-0.1,0.5)},anchor=south,rotate=90},
            width=.95\columnwidth, height=0.5\columnwidth,
            legend cell align={left},
            legend style={at={(1,0.65)}, anchor=east}
        ]
        \addplot[blue, 
                 thick, 
                 restrict x to domain=0:2.5, 
                 forget plot,
                 name path=Astar] {0.95 * exp(-x^2)};
        \draw[dashed, 
              blue, 
              thick] (axis cs:2.5,0) -- (axis cs:2.5,0.95);
        \node[anchor=west] at (axis cs: 2.55,0.3) {\textcolor{blue}{$\epszero=2.5$}};
        \node[anchor=center] at (axis cs: 4,1.02) {\accuracy = $95\%$};
        \addplot[red, 
                 thick, 
                 forget plot,
                 name path=Ai] {0.95 * exp(-x^2/3)};
        \addplot[black, 
                 dashed, 
                 forget plot,
                 name path=axes] {0.95};
        \addplot[black, 
                 dashed, 
                 forget plot,
                 opacity=0.,
                 name path=loweraxes] {0};
        \addplot[red, 
                 opacity=0.3,
                 pattern=north east lines,
                 pattern color=red
                 ] fill between[of=Ai and loweraxes,soft clip={domain=0:2.5}];
        \addplot[blue, 
                 opacity=0.3, 
                 pattern=grid, 
                 draw=black,
                 pattern color=blue
                 ] fill between[of=Astar and loweraxes,soft clip={domain=0:2.5}];
        \addplot[gray, 
                 opacity=0.3, 
                 pattern=crosshatch, 
                 draw=black,
                 pattern color=gray
                 ] fill between[of=axes and Ai,soft clip={domain=0:2.5}];
        \legend{
        \areacurve{\oneattack},
        \areacurve{\bestattack}
        }
    \end{axis}
    \end{tikzpicture}
    \caption{Robustness evaluation curves of \oneattack and \bestattack.}
    \label{fig:robust_accuracy_curves_2}
\end{figure}

%% file: pseudocode/attackbenchmark.tex
\begin{algorithm2e}[t]
 \SetKwInOut{Input}{Input}
    \SetKwInOut{Output}{Output}
    \SetKwComment{Comment}{$\triangleright$\ }{}
    \DontPrintSemicolon
    \LinesNumbered
    \setcounter{AlgoLine}{0}
	\caption{Computing Local Optimality}
	\label{alg:atkbenchmark_code}
   \Input{$\{\attack^{1},\dots, \attack^{N}\}$, list of attacks for benchmark; $\set D$, the validation dataset; $\vct \theta$, the target model; $Q$, query budget; $p$, threat model distance.}
	\Output{Local optimality measure $(\lOptimalityIndex{1}, \ldots, \lOptimalityIndex{n})$.}

    \For {$\oneattack \in \{\attack^{1}, \dots, \attack^{N}\}$}{
        $d_{i}, q_{i} \gets \texttt{TestAttack}(\oneattack, \set D, \vct \theta, Q, p)$\label{alg:test_attacks}
    }
    

    $\Aopt = \texttt{sample-wise-min}\{d_{1}, \dots, d_{N}\}$\label{alg:compile_results}
    
    \For {$\oneattack \in \{\attack^{1}, \dots, \attack^{n}\}$}{
        Get $\lOptimality$ for attack $\oneattack$ following \autoref{eq:local_optimality}\label{alg:local_optimality}
    }
 
	\bfseries return  $(\lOptimalityIndex{1}, \ldots, \lOptimalityIndex{n})$, $(q_1, \dots, q_n)$        
\end{algorithm2e}

%% file: experiments.tex
\section{Experiments}
\label{sec:experiments}
We now execute \ab to rate the adversarial attacks.
\subsection{Experimental Setup}\label{sec:exp_setup}
\myparagraph{Dataset.} We consider two popular datasets: CIFAR-10 \cite{Krizhevsky2009LearningML} and ImageNet \cite{deng2009imagenet}. 
We evaluate the performance of adversarial attacks on the entire CIFAR-10 test set, and on a random subset of $5\,000$ samples from the ImageNet validation set.
We use a batch size of $128$ for  CIFAR-10 and $32$ for ImageNet.

\myparagraph{Models.} 
\ab models \textit{zoo} utilizes a selection of both baseline and robust models to evaluate the effectiveness and adaptability of attacks against various model architectures and defense mechanisms.
For CIFAR-10, we include a baseline undefended WideResNet-28-10 (denoted as \standardcifar) from Robustbench~\cite{croce2020robustbench} and four robust models implementing the following defenses: a certified defense \cite{zhang2020towards}~(\zhang); adversarial training for generalization against unseen attacks \cite{Stutz2019CCAT}~(\stutz); gradient obfuscation defense \cite{Xiao20Defense}~(\xiao); and adversarial training with data augmentation \cite{Wang23Defense}~(\wang).
For ImageNet, we select four models from Robustbench~\cite{croce2020robustbench}: a pre-trained undefended ResNet-50, and three adversarially-trained models by \citet{Wong2020Defense}~(\wong), \citet{Salman2020Defense}~(\salman), and \citet{debenedetti2023Defense}~(\debenedetti).

\myparagraph{Adversarial Libraries.}
We integrate six publicly available adversarial attack libraries in \ab: FoolBox~(\foolboxshort)~\cite{rauber2017foolboxnative}, CleverHans~(\cleverhansshort)~\cite{papernot2018cleverhans}, AdvLib~(\advlibshort)~\cite{Rony_Adversarial_Library}, ART~(\artshort)~\cite{Nicolae2018AdversarialRT}, TorchAttacks~(\torchattacksshort)~\cite{Kim2020TorchattacksA}, and DeepRobust~(\deeprobustshort)~\cite{li2020deeprobust}. 
We also include authors' original implementations of their attacks to verify implementation inconsistencies (or improvements) among the different versions. We denote them denoted with \originalshort.

\myparagraph{Attack Settings.}
We focus on the well-studied $\ell_p$ perturbation models with $p \in \{0, 1, 2, \infty\}$.
We consider all the available implementations of the attacks listed in \Cref{table:adv_attacks_categorization}. 
We only run the attacks in their untargeted objective.
\footnote{\apgdt runs \apgd with a targeted objective on multiple classes, retaining the best result as an untargeted solution.}

\myparagraph{Hyperparameters.} 
For each considered attack implementation, we employed the default hyperparameters.
We set the maximum number of forward and backward propagations $Q$ to $2\,000$.
For an attack that does a single forward prediction and backward gradient computation per optimization step, this corresponds to the common ``$1\,000$ steps budget'' found in several works~\cite{brendel2018decision, rony2019decoupling, pintor2021fast, Rony2020AugmentedLA}, sufficient for algorithm convergence~\cite{pintor2022indicators}. Exceptionally, most implementations of the \cw attack use a default number of steps equal to $10^4$ for multiple runs to find $c$, so we modify it to perform the search of the penalty weight and the attack iterations within the $2,000$ propagations.
Furthermore, as done in \cite{Rony2020AugmentedLA, Cina2024SigmaZero}, for \maxloss attacks we leverage a line-search strategy to find the smallest budget $\budgetsize^\star$ for which the attack can successfully find an adversarial perturbation. Further details on the line search are reported in \Cref{appendix:search-fixed-budget}.

\myparagraph{Evaluation Metrics.}
For each attack, we compute the Attack Success Rate (\asr) as the percentage of samples in $\set D$ that the attack converts into adversarial examples at $\varepsilon = \infty$ which, if different from zero, indicates that the attack is applied incorrectly~\cite{carlini2019evaluating}, and the \textit{local} and \textit{global optimality} scores. Additionally, we track the computational effort of each attack expressed in terms of execution time and the number of forward and backward propagations required. 
The execution time is measured on a shared compute cluster equipped with NVIDIA V100 SXM2 GPU (16GB memory).

\subsection{Experimental Results}
\label{sec:exp_results}
We present the benchmark results of \ab on a total of $\textbf{815}$ experiments, encompassing the $102$ state-of-the-art gradient-based attacks implementations listed in \Cref{table:adv_attacks_categorization} (see \Cref{sec:advattacks}).  
For the $5$ CIFAR-10 models, we execute $6, 21, 42$, and $33$ implementations for the $\ell_0$, $\ell_1$, $\ell_2$, and $\ell_\infty$ perturbation models respectively, obtaining a total of $510$ experiments. For the $4$ ImageNet models, given the high data dimensionality, we reduced the load by excluding some implementations that were suboptimal in the CIFAR-10 experiments.
Therefore, we execute $6, 10, 9,$ and $10$ attack implementations for $\ell_0$, $\ell_1$, $\ell_2$, and $\ell_\infty$, obtaining $140$ experiments.\smallskip

\begin{figure*}[!htb]
    \centering
    \includegraphics[width=0.245\textwidth]{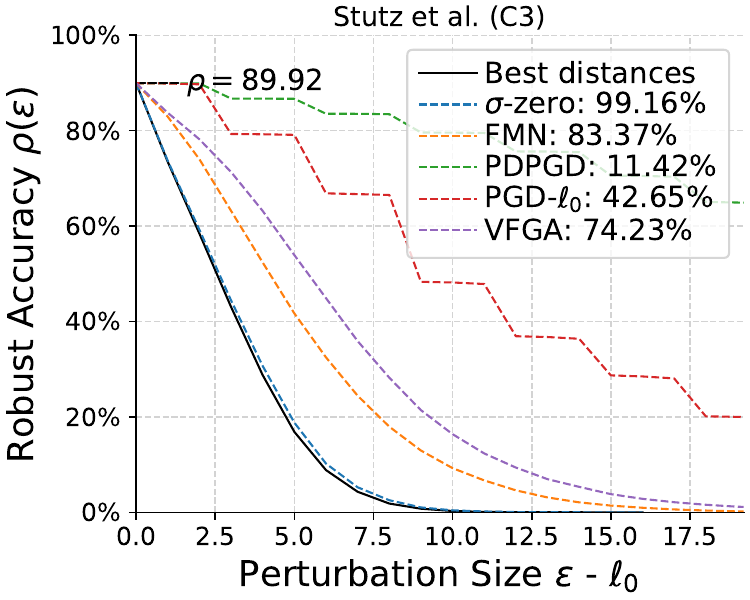}
    \includegraphics[width=0.245\textwidth]{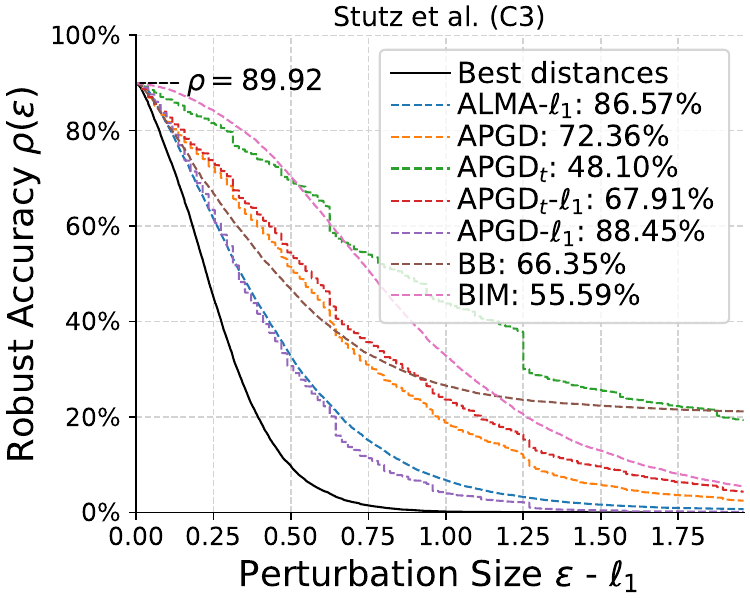}
    \includegraphics[width=0.245\textwidth]{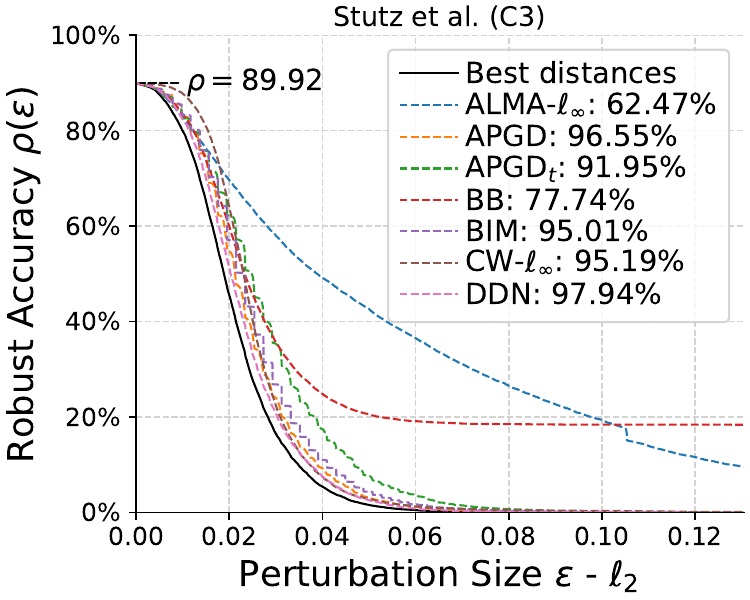}
    \includegraphics[width=0.245\textwidth]{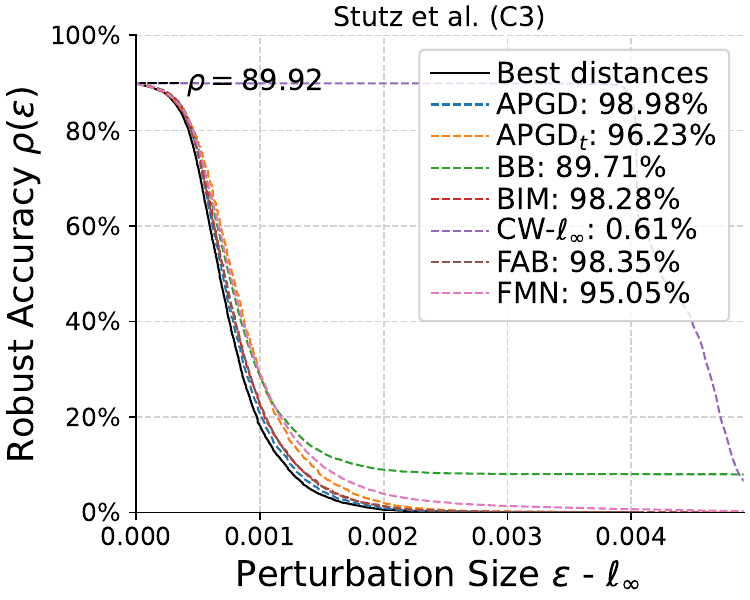}
    \caption{Robustness evaluation curves for the $7$ best $\ell_0$, $\ell_1$, $\ell_2$, and $\ell_\infty$-norm attacks against \stutz~\cite{Stutz2019CCAT}.
    }
    \label{fig:security_evaluation_curves_cifar_stutz}
\end{figure*}

\myparagraph{Attack Optimality.}
\Cref{tab:best_performing} shows the results obtained for the top-5 ranked attacks in each \ellp norm, according to our \go metric. 
Complementary, we report the $5$ worst-performing attacks in \Cref{tab:worst_performing}.
We defer to \Cref{apx:optimality_curves} additional results, including remaining attacks and singular model evaluations.
We also visually depict examples of the robustness evaluation curves obtained with the top-ranked attacks in \Cref{fig:security_evaluation_curves_cifar_stutz} computed against CIFAR-10 model \stutz~\cite{Stutz2019CCAT}.
We observe that the top-ranked attacks for the $\ell_0$ and $\ell_2$ norms are \sigmazero~\cite{Cina2024SigmaZero} and \ddn~\cite{rony2019decoupling} for both CIFAR-10 and ImageNet.
In their respective $\ell_p$ norms, these attacks achieve \go scores very close to 1, indicating they are nearly as effective as the empirical best attack \bestattack (blue curve in \Cref{fig:robust_accuracy_curves_2}, black curve in \Cref{fig:security_evaluation_curves_cifar_stutz}).
Regarding the $\ell_1$ and $\ell_\infty$ threat models, we observe the constant presence of \apgdlone~\cite{Croce2021MindTB}, \apgdt~\cite{croce2020reliable}, and \pdpgd~\cite{Matyasko2021PDPGDPP} in the top $5$ positions in both CIFAR-10 and ImageNet experiments. 
\pdpgd is, for example, the top-ranked attack in $\ell_1$ for CIFAR-10 and second on ImageNet by a small margin. 
In both cases, \apgd competes with \pdpgd. 
In $\ell_\infty$, we observe that \apgdt and \apgd demonstrate similar results, both contending for the lead. 
We highlight that \apgd, despite being conceived as a \maxloss attack, offers outstanding results also when configured as a \minnorm attack. 
This suggests that combining \apgd with the search strategy presented in \Cref{sec:exp_setup} is a viable strategy to efficiently find a small perturbation, even compared to \minnorm attacks.
Note finally that none of the tested attacks reaches $100\%$ of \go, suggesting that combining attacks, as done by~\citet{croce2020robustbench}, can lead to improved robustness evaluations.\smallskip



\myparagraph{Fast Attacks.} The top-ranked attacks in \Cref{tab:best_performing} do not always offer the best effectiveness/efficiency tradeoff. For instance, in the $\ell_1$ norm, \apgd achieves only a slightly better \go score than \pdpgd on ImageNet but is three times slower due to its original implementation and the integration of the $\epsilon$ search strategy. In the $\ell_0$ norm, the \vfga~\cite{Hajri2020StochasticSA} attack stands out as the fastest, requiring significantly fewer queries due to its early stopping mechanism. However, its ASR and \go largely drop for this attack when scaling to the ImageNet dataset, diminishing its reliability.\smallskip
\input{tables/best_attacks}

\myparagraph{Best and Worst Performing Implementations.} 
From \Cref{tab:best_performing} we observe that high-ranked attacks are frequently implemented in AdvLib and Foolbox libraries or are directly taken from the original repositories of the authors. 
Conversely, implementations from other libraries, such as the Art, Cleverhans, and Deeprobust, do not consistently appear among the top-performing attacks. 
In some cases, attacks integrated into these libraries even exhibit a decrease in performance.
For example, while \apgd implemented with AdvLib (as well as its original implementation) ranks among the best-performing attacks (see \Cref{tab:best_performing}) its implementation in the Art library ranks among the worst  (see \Cref{tab:worst_performing}). 
Specifically, for the CIFAR-10 dataset under the $\ell_1$ norm threat model, \go optimality drops from $90.9\%$ (original implementation) to $26\%$ (Art).
Upon inspecting the code, we found a key difference in the parameter controlling the number of random initializations for the attack: it is set to 5 in Art and 1 in both the original and AdvLib implementations.
Consequently, the restart mechanism of \apgd consumes more queries within the same perturbation budget $\pertsize$, leading to early termination of the attack without exploring smaller budgets. 
Another crucial difference is observed for \fab~\cite{githubAutoattackautoattackfab} and \cw~\cite{githubCWArtAttack} because they calculate the difference of the gradients of the two logits separately, thus requiring two backward passes rather than one. 
A final notable divergence in implementation is observed in the ImageNet results for the \apgd attack between the original repository and the AdvLib library. 
In the $\ell_2$ and $\ell_\infty$ norms, AdvLib's \apgd achieves \go scores of $95.9\%$ and $97\%$, respectively, compared to $36.2\%$ and $39.1\%$ for the original \apgd, representing substantial gaps of $59.7\%$ and $57.9\%$.
We also observed that while the original implementations used the \crossentropy loss, AdvLib defaults to the \dlr loss.\smallskip

\myparagraph{Ingredients for Optimal Attacks.}
We here identify key factors contributing to a high \go score in \Cref{tab:best_performing}. 
First, nearly all top-ranked attacks, including \sigmazero, \fmn, \pdpgd, \pgdlzero~\cite{Croce2019SparseAI}, \apgd, \apgdt, \bim~\cite{kurakin16adversarialexamples}, and \ddn, use normalization or linear projections on the gradient. These methods decouple the gradient's original size from the step size used in the updates~\cite{rony2019decoupling,pintor2021fast}.
Furthermore, top-ranked attacks (e.g., \sigmazero, \fmn, \pdpgd) are usually equipped with dynamic or adaptive step size schedulers, such as cosine annealing or reduce on plateau. 
None of the attacks securing the top rank in \Cref{tab:best_performing} adopts a fixed step size, confirming that dynamically reducing the step size across iterations and enhancing convergence stability contributes to achieving better optima.
Lastly, among the best-performing attacks, optimizers like Adam or momentum variants of gradient descent are only occasionally used.\smallskip

\myparagraph{Implementation Pitfalls.}
Through \ab, we found many coding errors across diverse libraries:
(i) the original \bb implementation crashes if not all samples are misclassified after initialization;
(ii) TorchAttacks DeepFool fails during batch processing if a sample has a label equal to the number of classes, resulting in an index out-of-bounds error;
(iii) Foolbox \fmn $\ell_0$ crashes with out-of-bounds errors during the projection step.
Lastly, per-sample $\pertsize$ evaluations in \maxloss attacks are supported only in AdvLib. 
\input{tables/worst_attacks}

%% file: tables/best_attacks.tex
\begin{table}[h!]
\centering
\label{tab:best_performing}
\setlength\tabcolsep{1.7pt}
\renewcommand{\arraystretch}{1}
\setlength{\dashlinedash}{5pt}
\setlength{\dashlinegap}{5pt}
\sisetup{table-auto-round}
{
\small{
\begin{tabular}{lllc*{2}{S[table-format=3.1, fixed-exponent=-2, drop-exponent=true, exponent-mode=fixed, drop-zero-decimal]}S[table-format=4]S[table-format=4]S[table-format=4.1]}
\toprule
 & \textbf{$\ell_p$} & \textbf{Attack} & \textbf{Library} & \textbf{ASR} & \textbf{GO} & \textbf{\#F} & \textbf{\#B} & \textbf{t(s)} \\ 
\midrule
\multirow{20}{*}{\begin{turn}{90}CIFAR-10\end{turn}} & \multirow{5}{*}{$\ell_0$} & \sigmazero & \originalshort & 1 & 0.984 & 998.8 & 998.8 & 292.20 \\
&& \fmn & \originalshort, \emph{\advlibshort} & 0.987 & 0.853 & 999.6 & 999.6 & 278.80 \\
&& \vfga & \advlibshort & 0.944 & 0.802 & 388.4 & 18.0 & 106.18 \\
&& \pgdlzero & \originalshort & 1 & 0.667 & 919.4 & 900.6 & 545.00 \\
&& \pdpgd & \advlibshort & 0.995 & 0.393 & 913.0 & 912.8 & 280.37\\
\cdashline{2-9}
& \multirow{5}{*}{$\ell_1$} & \pdpgd & \advlibshort & 0.998 & 0.932 & 995.0 & 995.0 & 279.60 \\
&& \apgdlone & \emph{\originalshort}, \advlibshort & 1 & 0.909 & 775.4 & 755.4 & 892.36 \\
&& \fmn & \emph{\originalshort}, \advlibshort, \foolboxshort & 0.979 & 0.904 & 1000.0 & 1000.0 & 275.98 \\
&& \apgdt & \emph{\originalshort}, \advlibshort & 1 & 0.854 & 577.0 & 536.0 & 860.59 \\
&& \ead & \foolboxshort & 1 & 0.7 & 923.0 & 923.0 & 276.72 \\
\cdashline{2-9}
&\multirow{5}{*}{$\ell_2$} & \ddn & \emph{\advlibshort}, \foolboxshort & 1 & 0.929 & 998.0 & 998.0 & 278.01 \\
&& \apgd & \emph{\originalshort}, \advlibshort & 1 & 0.929 & 775.4 & 755.4 & 709.24 \\
&& \apgdt & \emph{\originalshort}, \advlibshort & 1 & 0.922 & 522.4 & 481.6 & 641.82 \\
&& \pdgd & \advlibshort & 0.990 & 0.917 & 994.4 & 994.4 & 279.58 \\
&& \fmn & \originalshort, \emph{\advlibshort} & 0.995 & 0.908 & 998.2 & 998.2 & 275.34 \\
\cdashline{2-9}
& \multirow{5}{*}{$\ell_\infty$} & \apgdt & \emph{\originalshort}, \advlibshort & 1 & 0.976 & 629.0 & 583.8 & 626.14 \\
&& \apgd & \emph{\originalshort}, \advlibshort & 1 & 0.975 & 775.4 & 755.4 & 711.50 \\
&& \bim & \foolboxshort & 0.999 & 0.946 & 999.0 & 989.0 & 692.29 \\
&& \pgd & \advlibshort & 1 & 0.932 & 999.6 & 989.6 & 281.78 \\
&& \pdpgd & \advlibshort & 0.998 & 0.908 & 992.0 & 992.0 & 284.64 \\ 
\midrule
\midrule
\multirow{20}{*}{\begin{turn}{90}ImageNet\end{turn}} & \multirow{5}{*}{$\ell_0$} & \sigmazero & \originalshort & 1 & 0.998 & 999.0 & 999.0 & 345.17 \\
&& \fmn & \originalshort, \emph{\advlibshort} & 1 & 0.875 & 1000.0 & 1000.0 & 358.76 \\
&& \vfga & \advlibshort & 0.636 & 0.712 & 927.75 & 43.75 & 76.36 \\
&& \pdpgd & \advlibshort & 1 & 0.536 & 982.75 & 982.75 & 345.05 \\
&& \pgdlzero & \originalshort & 1 & 0.392 & 912.5 & 893.25 & 1680.1 \\
\cdashline{2-9}

& \multirow{5}{*}{$\ell_1$} & \apgd & \emph{\originalshort}, \advlibshort & 1 & 0.980 & 693.0 & 673.0 & 1085.9 \\
&& \pdpgd & \advlibshort & 1 & 0.973 & 999.3 & 999.3 & 339.24 \\
&& \apgdt & \emph{\originalshort}, \advlibshort & 1 & 0.948 & 579.3 & 538.3 & 990.13 \\
&& \alma & \advlibshort & 1 & 0.925 & 972.5 & 972.5 & 359.26 \\
&& \ead & \foolboxshort & 1 & 0.907 & 403.0 & 205.8 & 77.54 \\
\cdashline{2-9}
& \multirow{5}{*}{$\ell_2$} & \ddn & \advlibshort, \emph{\foolboxshort} & 1 & 0.986 & 1000.0 & 1000.0 & 335.99 \\
&& \apgdt & \advlibshort & 0.999 & 0.981 & 670.3 & 650.3 & 325.08 \\
&& \bim & \foolboxshort & 1 & 0.973 & 1000.0 & 990.0 & 1252.4 \\
&& \fmn & \originalshort, \emph{\advlibshort} & 0.996 & 0.971 & 999.8 & 999.8 & 362.75 \\
&& \apgd & \originalshort, \emph{\advlibshort} & 1 & 0.959 & 1000.0 & 990.0 & 333.54 \\
\cdashline{2-9}
& \multirow{5}{*}{$\ell_\infty$} &\apgdt & \advlibshort & 0.998 & 0.992 & 717.5 & 697.5 & 358.65 \\
&& \pdpgd & \advlibshort & 1 & 0.982 & 987.3 & 987.3 & 409.07 \\
&& \fmn & \emph{\originalshort}, \advlibshort, \foolboxshort & 1 & 0.982 & 999.8 & 999.8 & 335.15 \\
&& \bim & \foolboxshort & 1 & 0.980 & 1000.0 & 990.0 & 1293.7 \\
&& \apgd & \advlibshort & 1 & 0.970 & 1000.0 & 990.0 & 333.98 \\ 
\bottomrule
\end{tabular}
}}
\caption{Top-performing attacks. For each attack, we list the library implementations that yield similar results, highlighting the best one in italics and report its corresponding statics. We include the number of forward (\textbf{\#F}) and backward (\textbf{\#B}) passes, as well as the total runtime (\textbf{t(s)}) for each attack.
}
\end{table}

%% file: tables/worst_attacks.tex
\begin{table}[t]
\centering
\setlength\tabcolsep{1.3pt}
\renewcommand{\arraystretch}{0.95}
\setlength{\dashlinedash}{5pt}
\setlength{\dashlinegap}{5pt}
\label{tab:worst_performing}
\sisetup{table-auto-round}
{
\small{
\begin{tabular}{llc*{2}{S[table-format=3.1, fixed-exponent=-2, drop-exponent=true, exponent-mode=fixed, drop-zero-decimal]}S[table-format=4]S[table-format=4]S[table-format=4.1]}
\toprule
\textbf{$\ell_p$} & \textbf{Attack} & \textbf{Library} & \textbf{ASR} & \textbf{GO} & \textbf{\#F} & \textbf{\#B} & \textbf{t(s)} \\
\midrule
 & \pgd & \foolboxshort & 1 & 0.556 & 1000 & 990 & 715.0 \\
 & \ead & \artshort & 0.852 & 0.533 & 334 & 1665 & 295.7 \\
 & \fgm & \artshort, \emph{\foolboxshort} & 0.977 & 0.280 & 40 & 20 & 30.3 \\
 & \apgd & \artshort & 0.988 & 0.256 & 822 & 354 & 456.9 \\
\multirow{-5}{*}{$\ell_1$} & \bb & \foolboxshort & 0.380 & 0.38 & 623 & 36 & 119.4 \\
\cmidrule(l{0.5em}r{0.5em}){1-8}
 & \deepfool & \foolboxshort & 0.986 & 0.406 & 255.6 & 254.6 & 21.2 \\
 & \fgm & \artshort, \emph{\cleverhansshort}, \deeprobustshort, \foolboxshort & 0.976 & 0.379 & 41 & 20 & 28.1 \\
 & \deepfool & \artshort & 0.849 & 0.323 & 268.6 & 1340.8 & 317.8 \\
 & \bb & \foolboxshort & 0.383 & 0.309 & 623.6 & 36 & 112.1 \\
\multirow{-5}{*}{$\ell_2$} & \bim & \artshort & 0.957 & 0.226 & 807.6 & 781.8 & 322.2 \\
\cmidrule(l{0.5em}r{0.5em}){1-8}
 & \apgd & \artshort & 0.945 & 0.775 & 1037.2 & 504.4 & 390.0 \\
 & \fgsm & \emph{\torchattacksshort, \foolboxshort, \deeprobustshort, \cleverhansshort, \artshort} & 0.976 & 0.629 & 40 & 20 & 7.9 \\
 & \cw & \emph{\artshort}, AdvLib & 0.862 & 0.625 & 1320.6 & 640.4 & 2314.4 \\
 & \deepfool & \foolboxshort & 0.983 & 0.468 & 128.8 & 127.6 & 64.1 \\
\multirow{-5}{*}{$\ell_\infty$} & \bb & \foolboxshort & 0.429 & 0.320 & 806.2 & 134.8 & 139.0 \\
\bottomrule
\end{tabular}
}
\caption{Worst performing attacks on CIFAR-10. For each attack, we list the library implementations that yield similar results, highlighting the worst one in italics with its statistics. 
We include the number of forward (\textbf{\#F}) and backward (\textbf{\#B}) passes, as well as the total runtime (\textbf{t(s)}) for each attack. 
}

}
\end{table}

%% file: related.tex
\section{Related Work}

Although many gradient-based attacks for optimizing adversarial examples have been proposed, their performance is often evaluated independently in each study, increasing the risk of evaluation bias and hindering reproducibility.
These evaluations do not ensure that attack comparisons are conducted under the same fair experimental settings (e.g., consistent query budgets or data subsets).
Moreover, they do not usually compare \maxloss or \minnorm under the same evaluation metric, and they do not ensure consistency with the original implementation of the attack, which can induce to misleading conclusions~\cite{carlini2019critique}.

Researchers have also proposed some benchmarks to standardize evaluations of models' robustness.  
\citet{ling-19-deepsec} propose DeepSec, which tests 16 state-of-the-art adversarial attacks against 13 robust models computing 12 different metrics defined by the authors to evaluate the attacks' utility. 
However, DeepSec considers only a fixed arbitrary perturbation budget, and it does not provide a metric that can be used to rank the attacks' effectiveness.
Additionally, DeepSec has faced significant criticism for its implementation flaws~\cite{carlini2019critique}. 
Frameworks like \cite{croce2020robustbench,yinpeng20-cvpr,GUO2023-pr} focus on evaluating model robustness, not the attacks themselves, often overlooking factors like computational effort. 
Unlike \cite{croce2020robustbench,GUO2023-pr}, which evaluate only at a single, arbitrary perturbation budget, \citet{yinpeng20-cvpr} use robustness evaluation curves to compare model robustness, but this approach lacks an objective metric, relying instead on visual inspection of the curves.
Additionally, these curves are not directly comparable across models with different \emph{clean accuracy}, making it difficult to identify the most effective attacks.
In summary, these robustness evaluation benchmarks propose metrics that are not combinable across models to reflect the global performance of attacks, and they often ignore computational efficiency. 
\ab addresses these limitations by providing an evaluation framework that ranks adversarial attacks based on a single metric that accounts for their effectiveness and efficiency on a consistent experimental setup covering multiple robust models and attack implementations. 

%% file: conclusions.tex
\section{Conclusion and Future Work}
\label{sec:conclusions}
In this work, we propose \ab, a comprehensive benchmark to evaluate gradient-based attacks, which stands on top of the proposed \emph{optimality} metric. This novel measure is the first that allows ranking attack algorithms according to their effectiveness across their entire robustness evaluation curve. 
Employing \ab, we enable researchers to evaluate attacks effectiveness in fair conditions that ensure both reproducibility and impartial ranking.
Thanks to \ab, we can spotlight attacks that excel in most of the perturbation models while also alerting the presence of critical errors of libraries that prevent some strategies from properly running.
\ab, currently restricted to evaluating gradient-based attacks on image models, is a versatile methodology with the potential to be extended in future work to black-box attacks and other domains (e.g., malware).

%% file: bio.tex
\begin{IEEEbiography}[{\includegraphics[width=1in,height=1.25in,clip, keepaspectratio]{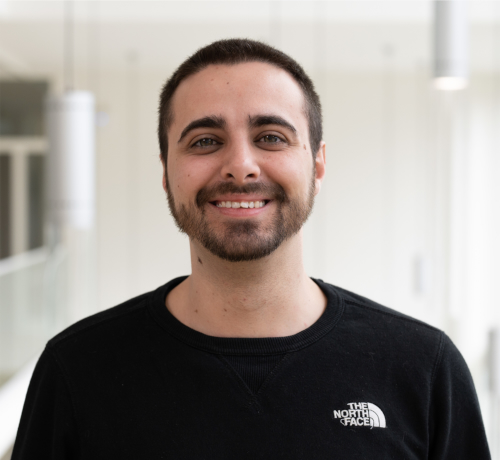}}]{Antonio Emanuele Cinà} (MSc 2019, PhD 2023) is an Assistant Professor at the University of Genoa, Italy. His research interests focus on the security of AI systems and the study of their trustworthiness.
In particular, he has studied possible vulnerabilities and emerging risks arising from AI caused by the malicious use of training data,
Finally, he works in the field of Generative AI, exploring how this technology can be integrated to optimize user applications.
\end{IEEEbiography}

\begin{IEEEbiography}[{\includegraphics[width=1in,height=1.25in,clip, keepaspectratio]{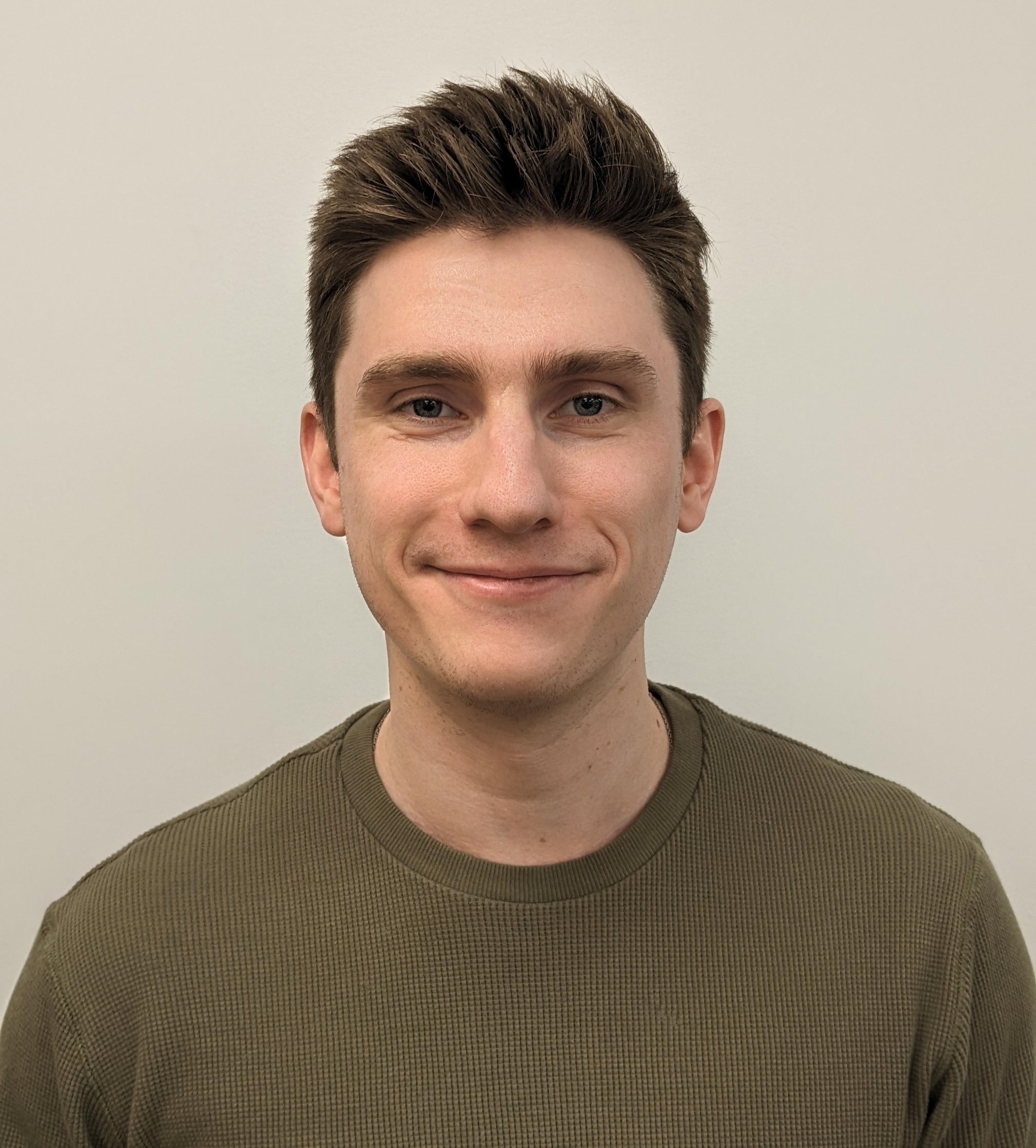}}]{Jérôme Rony} is a machine learning scientist at Synlico Inc. working on computational biology. He received his PhD from ÉTS Montréal in 2023, where his research focused on solving constrained optimization problems involving neural networks, which are central to designing safe and trustworthy models.
\end{IEEEbiography}

\begin{IEEEbiography}[{\includegraphics[width=1in,height=1.25in,clip, keepaspectratio]{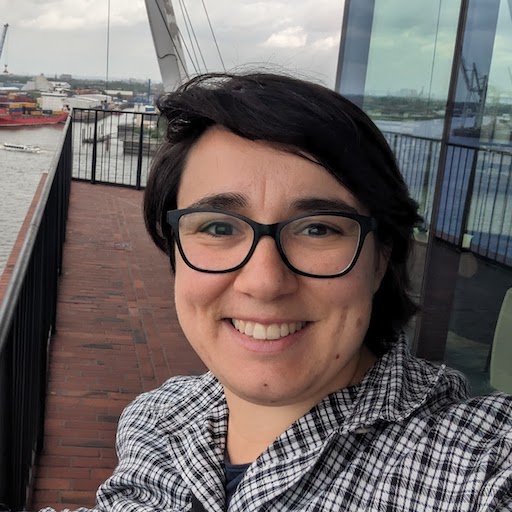}}]{Maura Pintor} is an Assistant Professor at the PRA Lab, in the Department of Electrical and Electronic Engineering of the University of Cagliari, Italy. She received her PhD in Electronic and Computer Engineering from the University of Cagliari in 2022. Her research focuses on machine learning security, with a particular focus on optimizing and debugging adversarial attacks.
\end{IEEEbiography}

\begin{IEEEbiography}[{\includegraphics[width=1in,height=1.25in,clip, keepaspectratio]{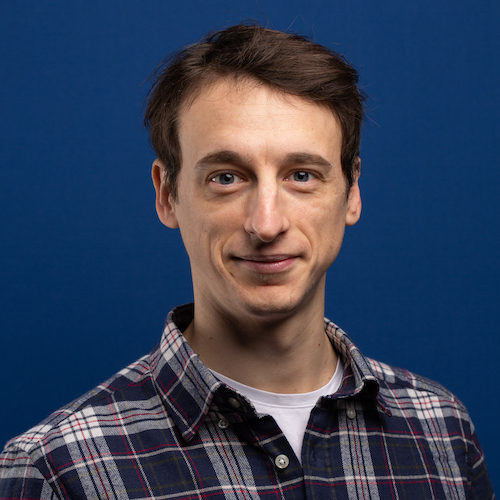}}]{Luca Demetrio} (MSc 2017, PhD 2021) is an Assistant Professor at the University of Genoa.
He is currently studying the security of Windows malware detectors implemented with Machine Learning techniques, and he is first author of papers published in top-tier journals (ACM TOPS, IEEE TIFS). He is part of the development team of SecML, and the maintainer of SecML Malware, a Python library for creating adversarial Windows malware.
\end{IEEEbiography}

\begin{IEEEbiography}[{\includegraphics[width=1in,height=1.25in,clip, keepaspectratio]{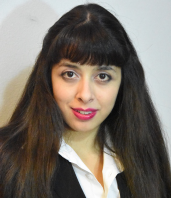}}]{Ambra Demontis} is an Assistant Professor at the University of Cagliari, Italy. She received her M.Sc. degree (Hons.) in Computer Science and her Ph.D. degree in Electronic Engineering and Computer Science from the University of Cagliari, Italy. Her research interests include secure machine learning, kernel methods, biometrics, and computer security.
\end{IEEEbiography}

\begin{IEEEbiography}[{\includegraphics[width=1in,height=1.25in,clip, keepaspectratio]{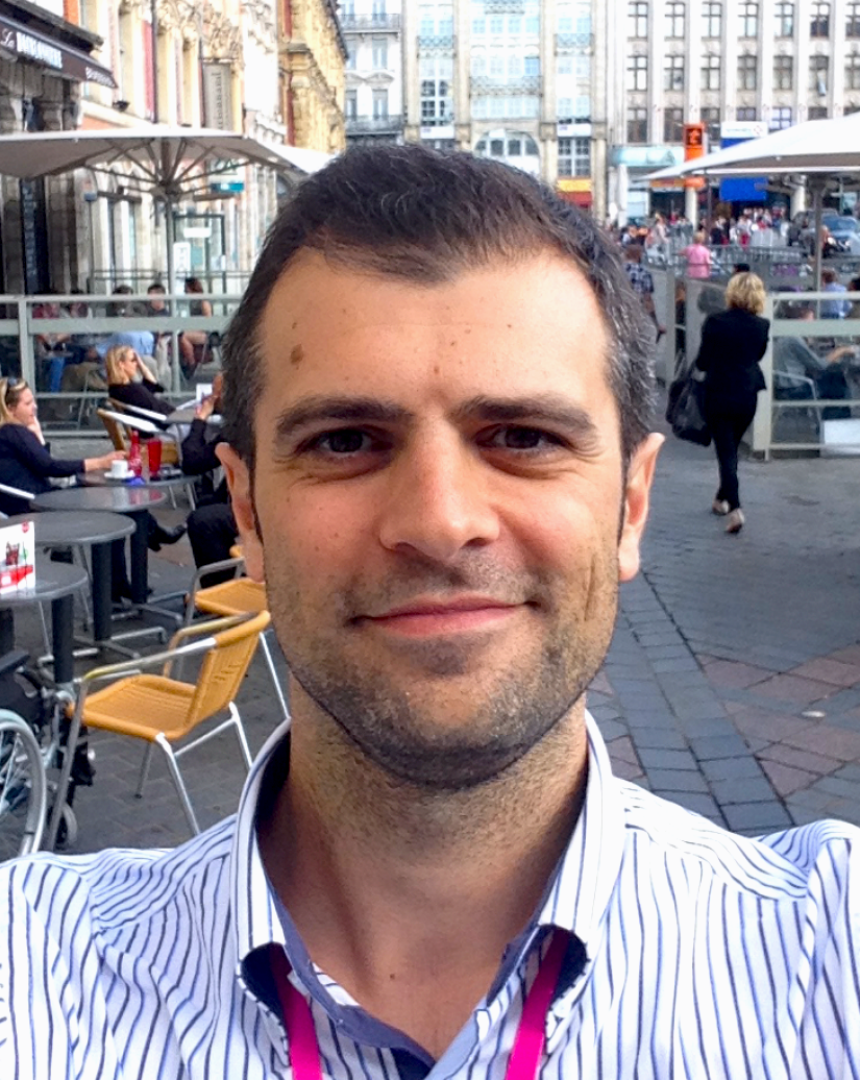}}]{Battista Biggio}
(MSc 2006, PhD 2010) is Full Professor at the University of Cagliari, Italy, and co-founder of the company Pluribus One. His research interests include adversarial machine learning and cybersecurity. He is the recipient of the 2022 ICML Test of Time Award, for his work on ``Poisoning Attacks against Support Vector Machines''. He is Senior Member of the IEEE and of the ACM, and Member of the IAPR and ELLIS.
\end{IEEEbiography}

\begin{IEEEbiography}[{\includegraphics[width=1in,height=1.25in,clip, keepaspectratio]{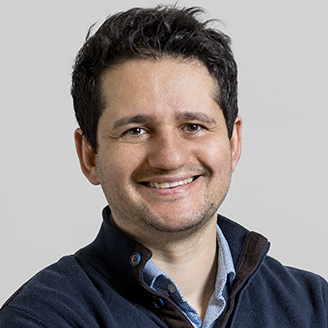}}]{Ismail Ben Ayed} is currently a full professor with ÉTS Montréal and affiliated with the CRCHUM. His interests are in computer vision, optimization, machine learning and medical image analysis algorithms. He authored more than 100 peer-reviewed papers, mostly published in the top venues, along with 2 books and 7 U.S. patents. His research has been covered in several visible media outlets, such as Radio Canada (CBC), Quebec Science Magazine and Canal du Savoir. His research team received several recent distinctions, such as the MIDL’19 best paper runner-up award and several top-ranking positions in internationally visible contests. He served on the program committee for MICCAI’15, ’17, and ’19 and as program chair for MIDL’20. He regularly serves as a reviewer for the main scientific journals of his field, and was selected several times among the top reviewers of prestigious conferences (such as CVPR’15 and NeurIPS’20).
\end{IEEEbiography}

\begin{IEEEbiography}[{\includegraphics[width=1in,height=1.25in,clip, keepaspectratio]{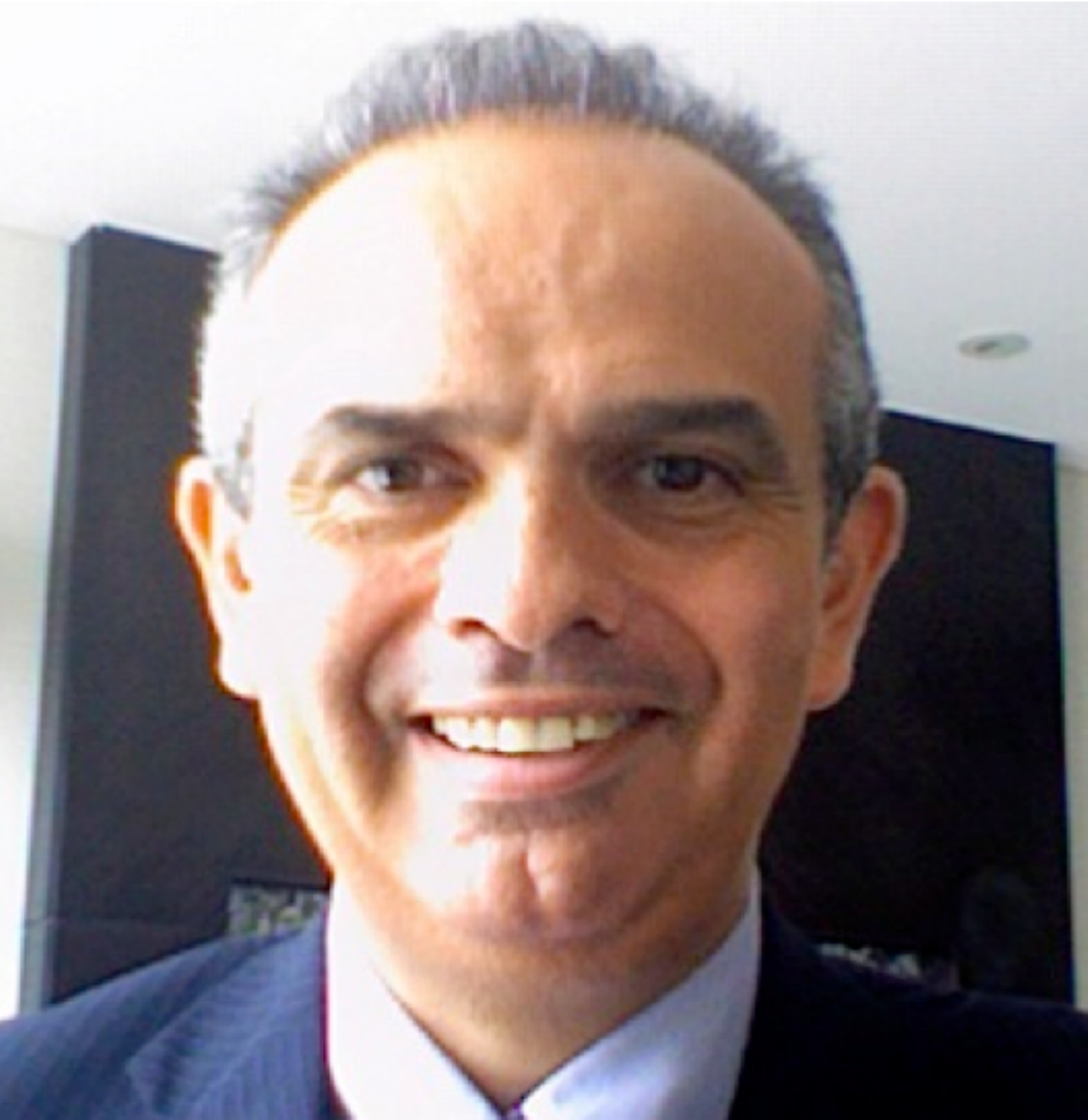}}]{Fabio Roli} is a Full Professor of Computer Science at the University of Genoa, Italy. He has been appointed Fellow of the IEEE and Fellow of the International Association for Pattern Recognition. He is a recipient of the Pierre Devijver Award for his contributions to statistical pattern recognition.
\end{IEEEbiography}

%% file: appendix/appendix.tex
\input{appendix/background}
\input{appendix/implementation}

\section{Additional Experiments}
In the following we report the remaining experimental results conducted with \ab.
In particular, we show the robustness evaluation curves, the top-5 $\ell-\inf$ performing attacks on CIFAR-10, and the tables showing the local optimality for all the considered implementations run on CIFAR-10. Upon request, we can add the same tables for ImageNet. 
\label{apx:optimality_curves}
We report in \cref{table:adv_attacks_categorization} the gradient-based attacks considered in this paper, the norms they support, and their supported library implementations.

\input{checklist}

\input{tables/attacks_libraries}

\begin{figure*}[!htb]
    \centering
    \includegraphics[width=0.245\textwidth]{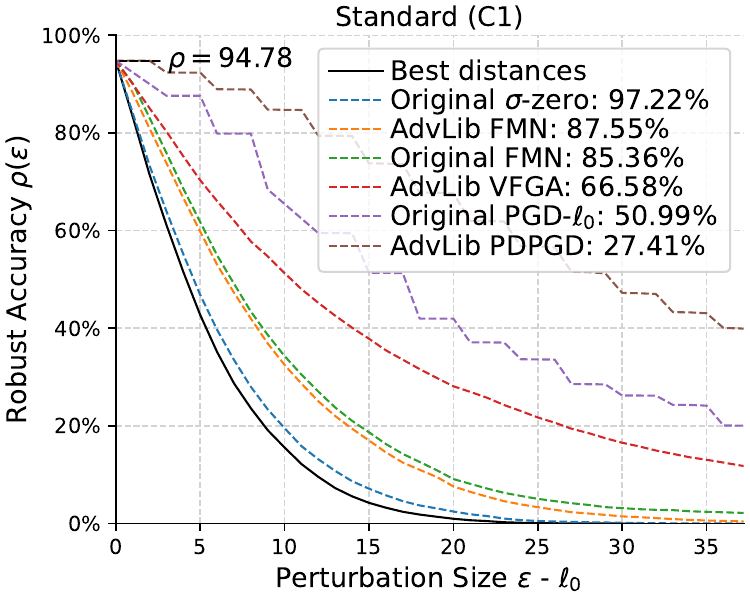}
    \includegraphics[width=0.245\textwidth]{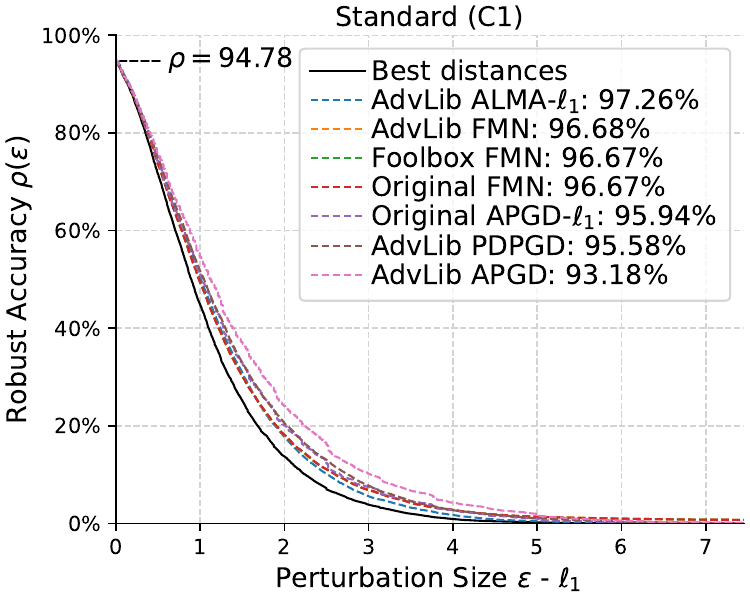}
    \includegraphics[width=0.245\textwidth]{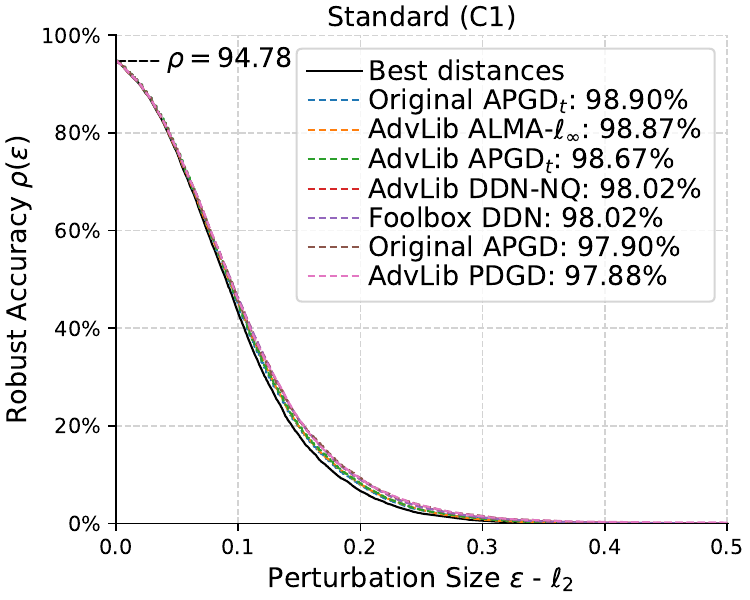}
    \includegraphics[width=0.245\textwidth]{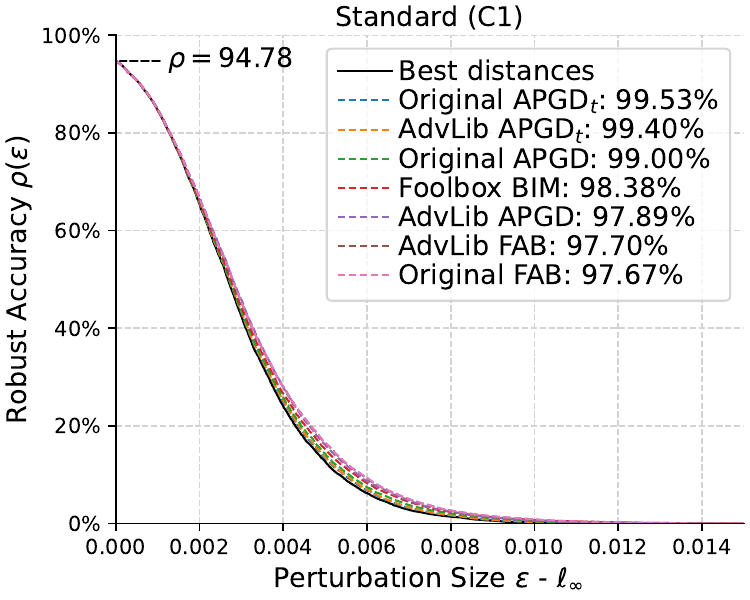}
    \caption{Robustness evaluation curves for the best $\ell_0$, $\ell_1$, $\ell_2$, and $\ell_\infty$-norm attacks against \standardcifar in CIFAR-10.}
    \label{fig:standard_vulnerable_security_evaluation_curves_cifar}
\end{figure*}

\begin{figure*}[!htb]
    \centering
    \includegraphics[width=0.245\textwidth]{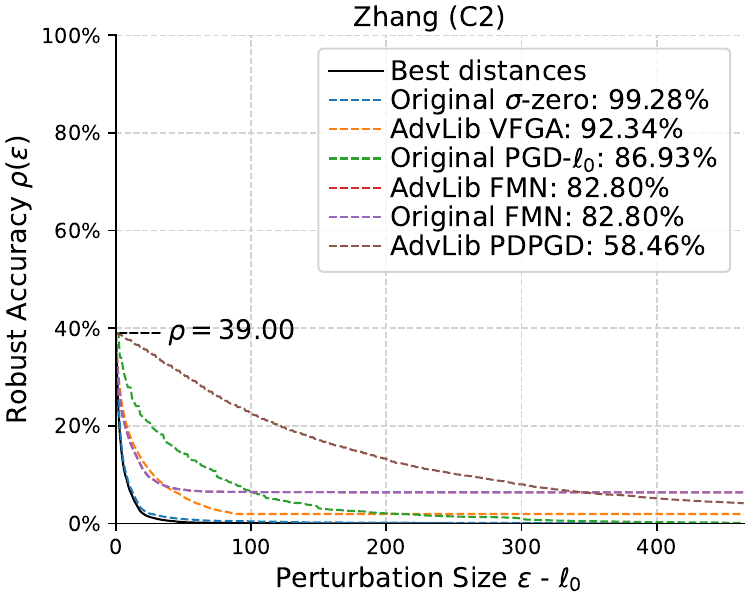}
    \includegraphics[width=0.245\textwidth]{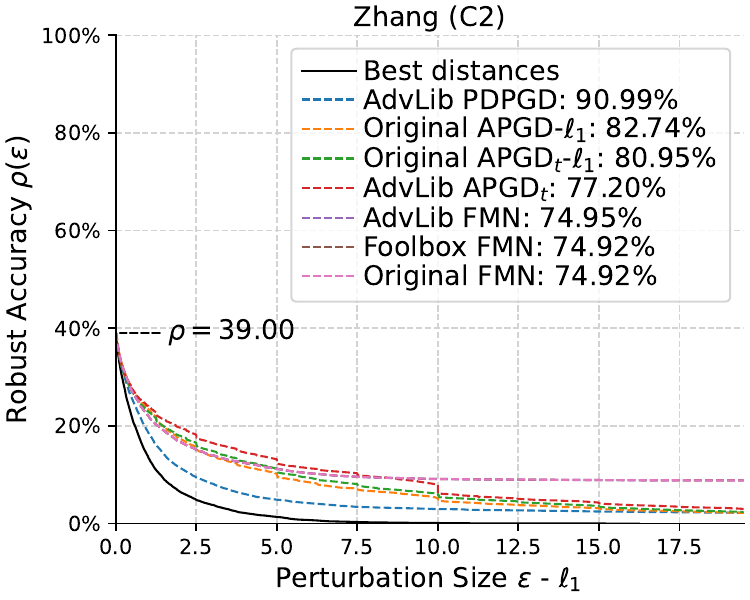}
    \includegraphics[width=0.245\textwidth]{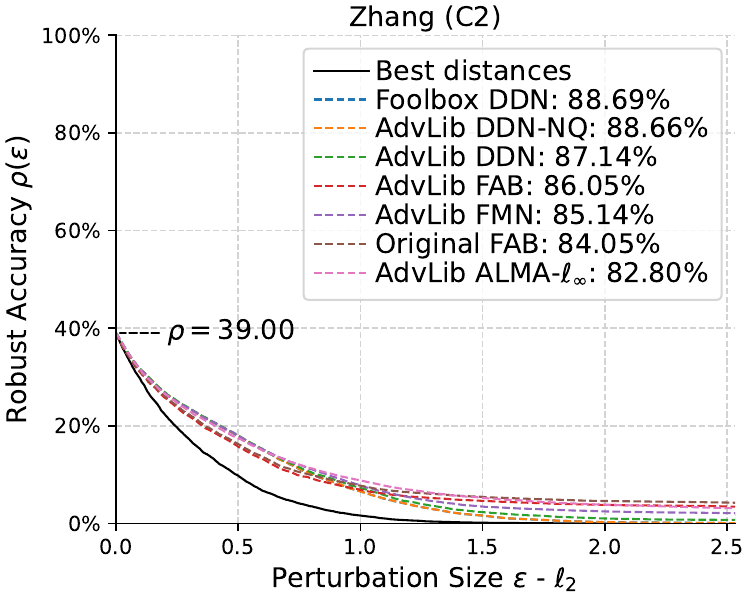}
    \includegraphics[width=0.245\textwidth]{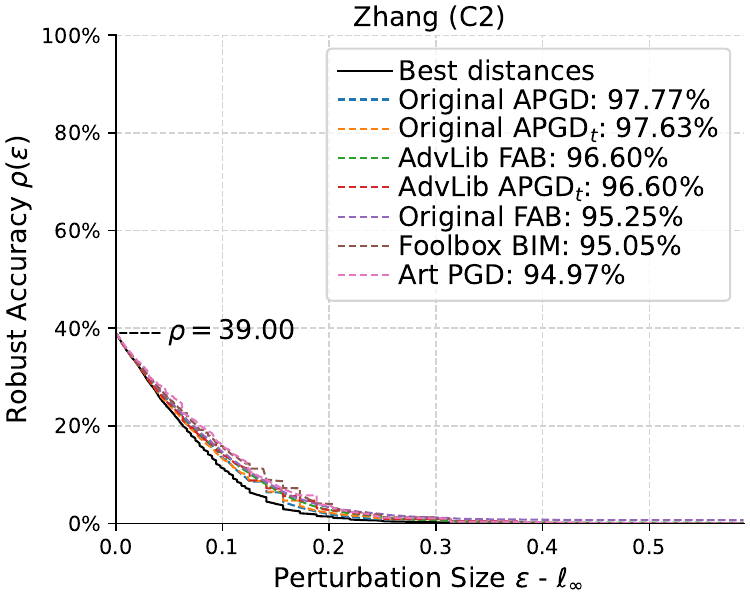}
    \caption{Robustness evaluation curves for the best $\ell_p$-norm attacks against \zhang in CIFAR-10.}
    \label{fig:zhang_security_evaluation_curves_cifar}
\end{figure*}

\begin{figure*}[!htb]
    \centering
    \includegraphics[width=0.245\textwidth]{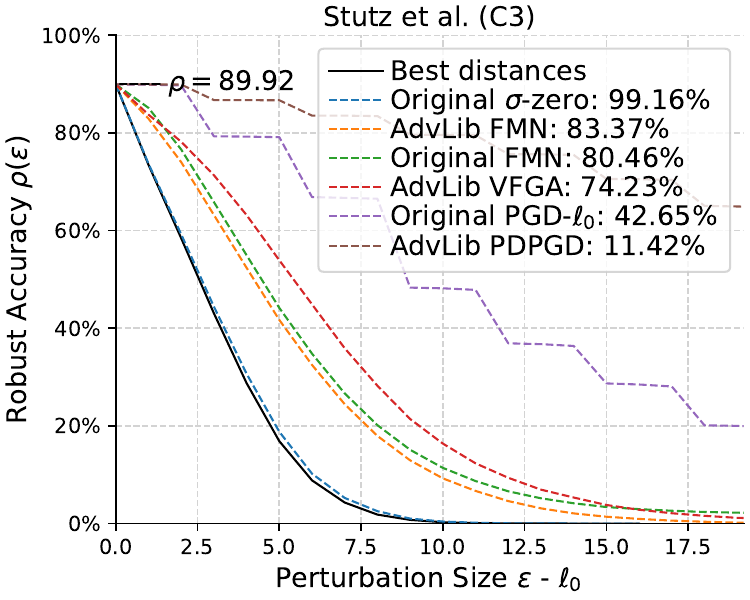}
        \includegraphics[width=0.245\textwidth]{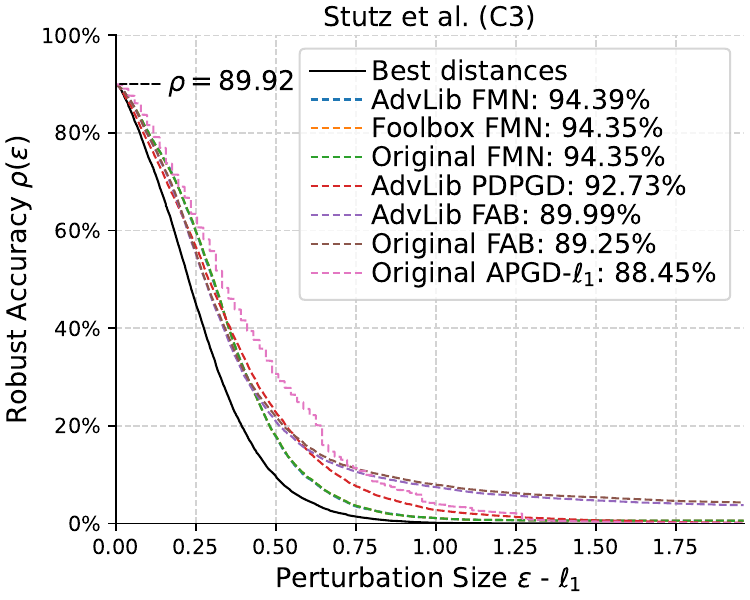}
    \includegraphics[width=0.245\textwidth]{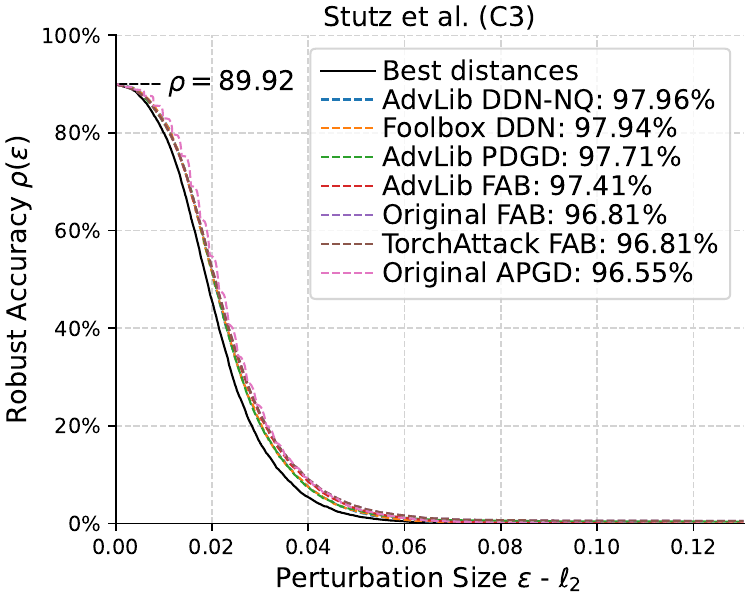}
    \includegraphics[width=0.245\textwidth]{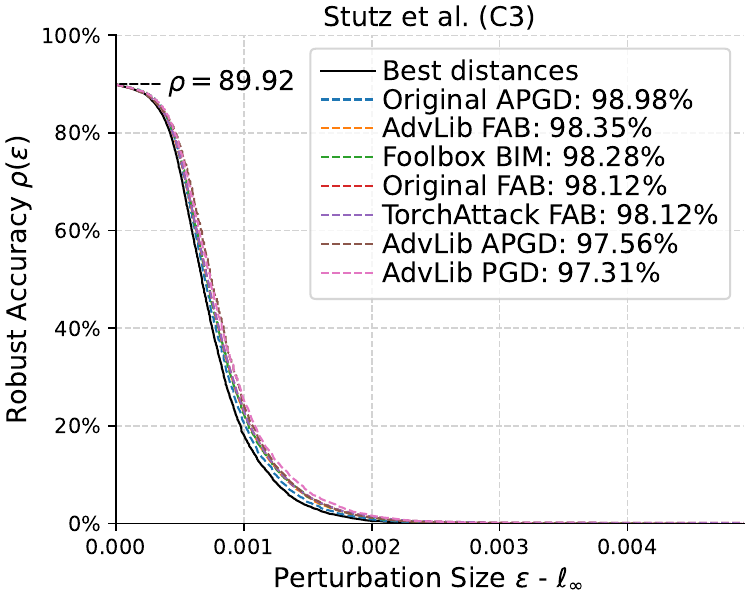}
    \caption{Robustness evaluation curves for the best $\ell_p$-norm attacks against \stutz in CIFAR-10.}
    \label{fig:stutz_security_evaluation_curves_cifar}
\end{figure*}

\begin{figure*}[!htb]
    \centering
    \includegraphics[width=0.245\textwidth]{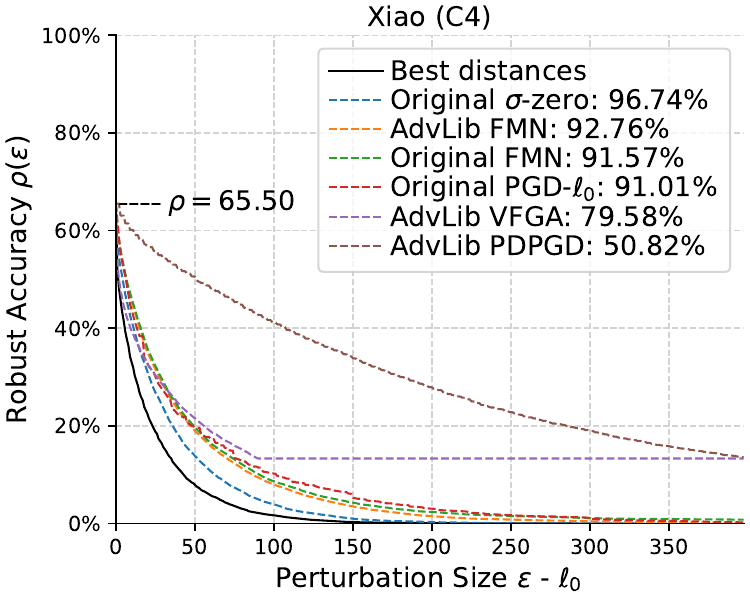}
    \includegraphics[width=0.245\textwidth]{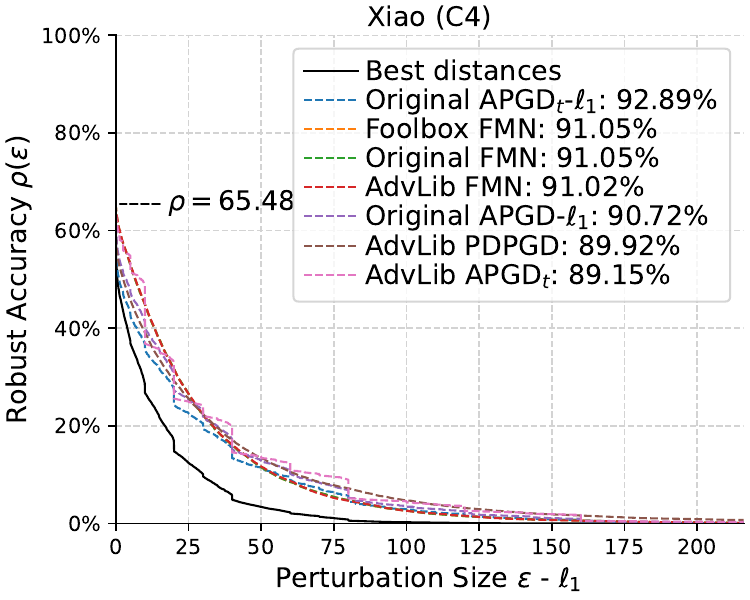}
    \includegraphics[width=0.245\textwidth]{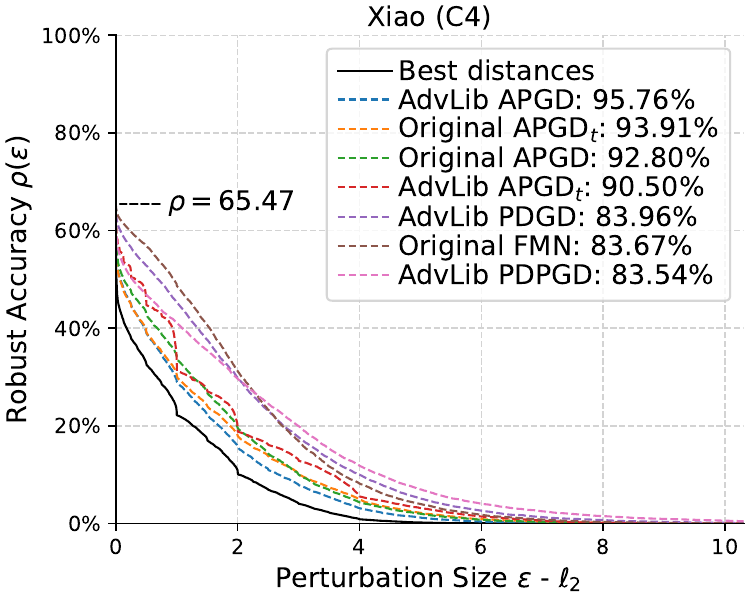}
    \includegraphics[width=0.245\textwidth]{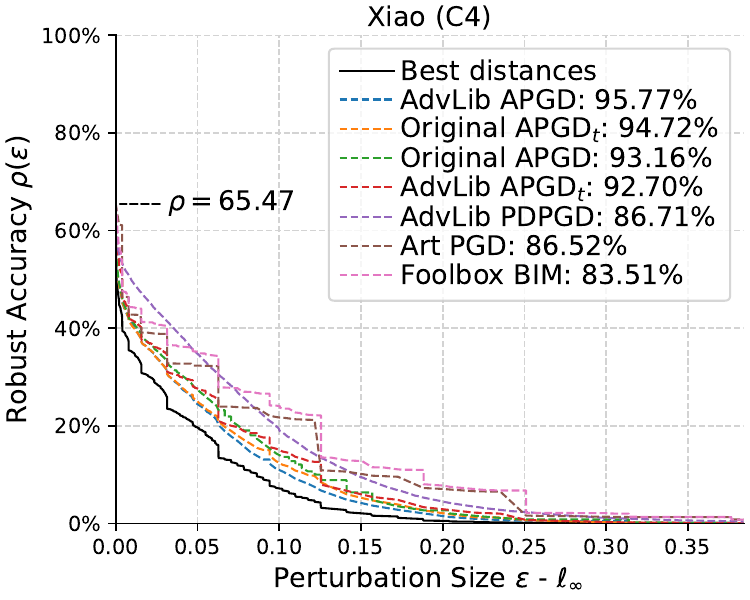}
    \caption{Robustness evaluation curves for the best $\ell_p$-norm attacks against \xiao in CIFAR-10.}
    \label{fig:xiao_security_evaluation_curves_cifar}
\end{figure*}

\begin{figure*}[!htb]
    \centering
    \includegraphics[width=0.245\textwidth]{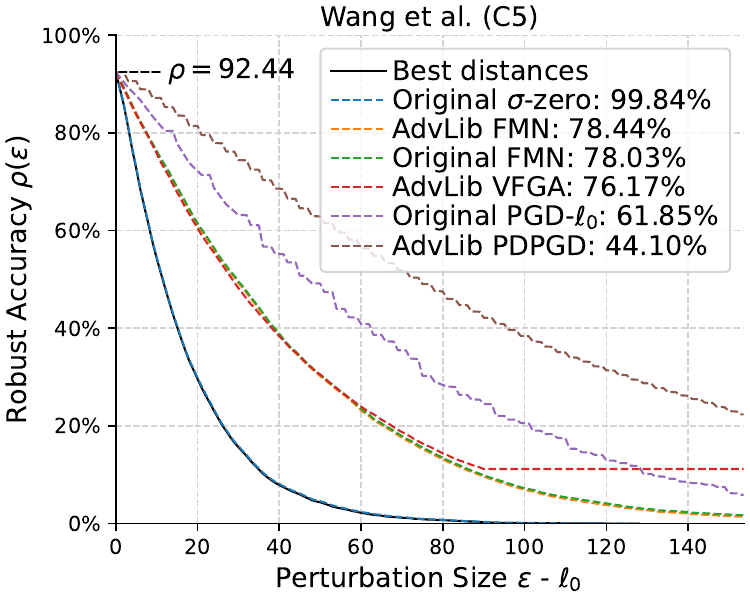}
    \includegraphics[width=0.245\textwidth]{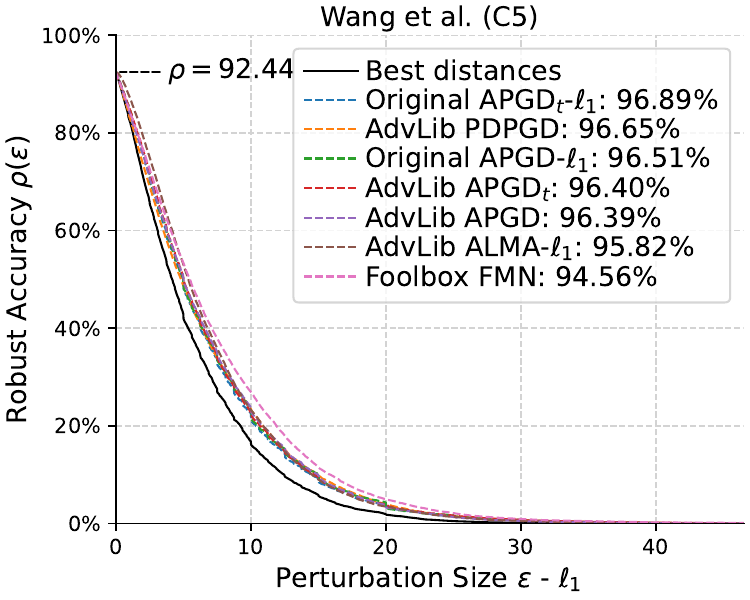}
    \includegraphics[width=0.245\textwidth]{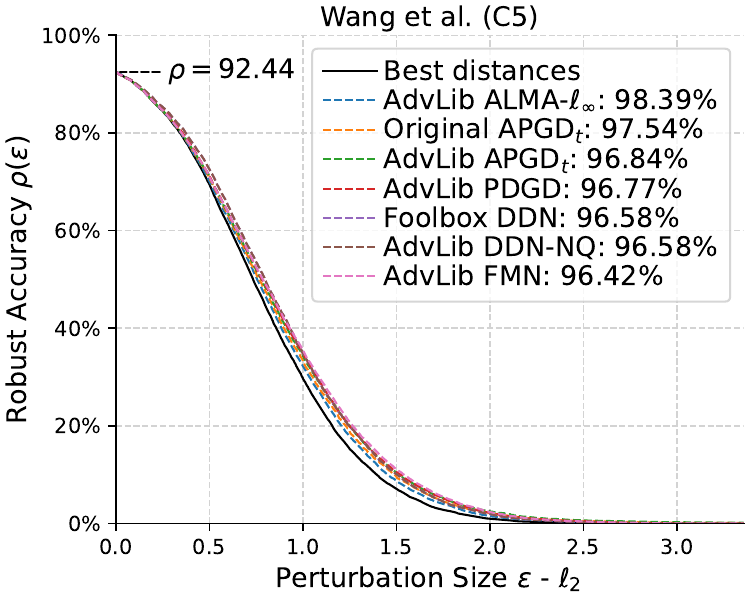}
    \includegraphics[width=0.245\textwidth]{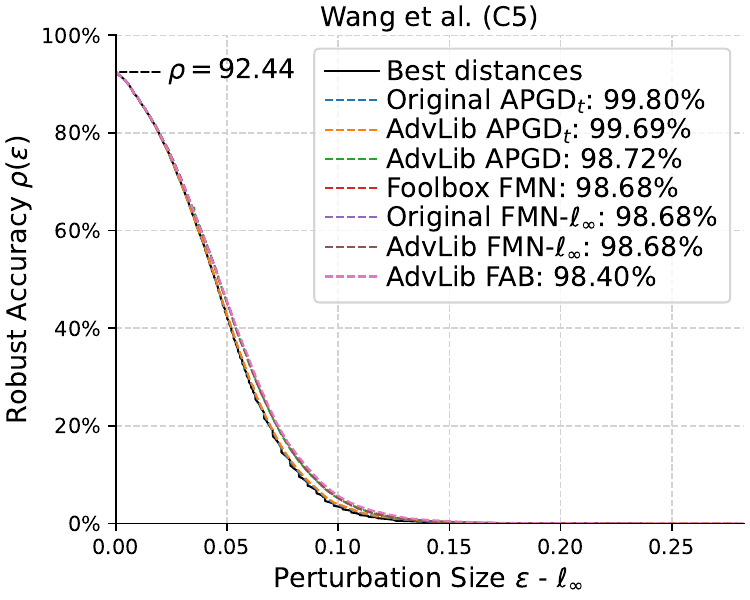}
    \caption{Robustness evaluation curves for the best $\ell_p$-norm attacks against \wang in CIFAR-10.}
    \label{fig:wang_security_evaluation_curves_cifar}
\end{figure*}

\begin{figure*}[!htb]
    \centering
    \includegraphics[width=0.245\textwidth]{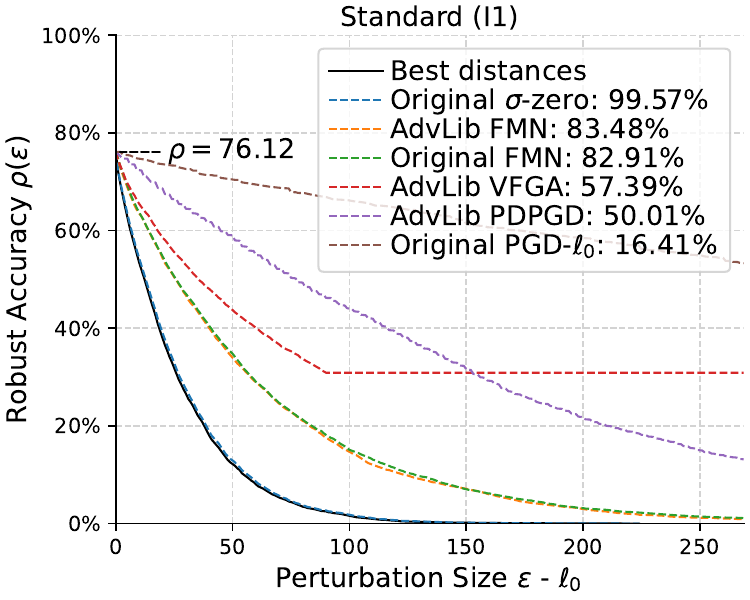}
    \includegraphics[width=0.245\textwidth]{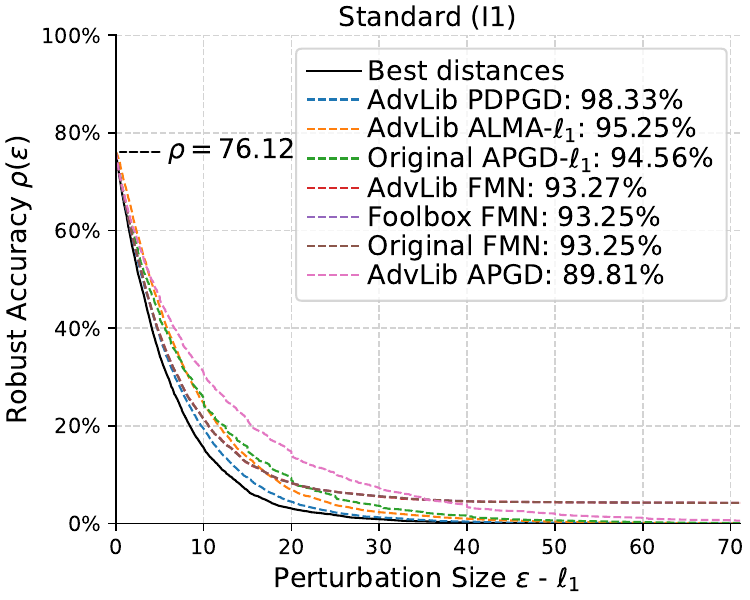}
    \includegraphics[width=0.245\textwidth]{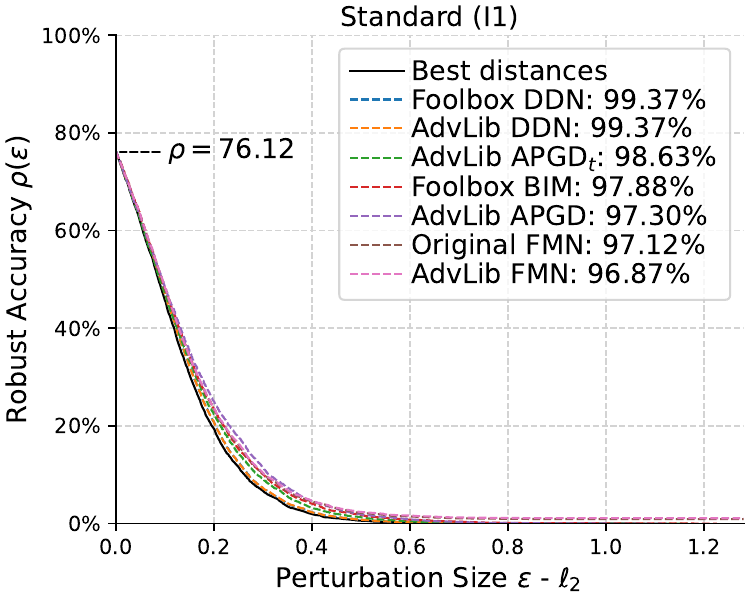}
    \includegraphics[width=0.245\textwidth]{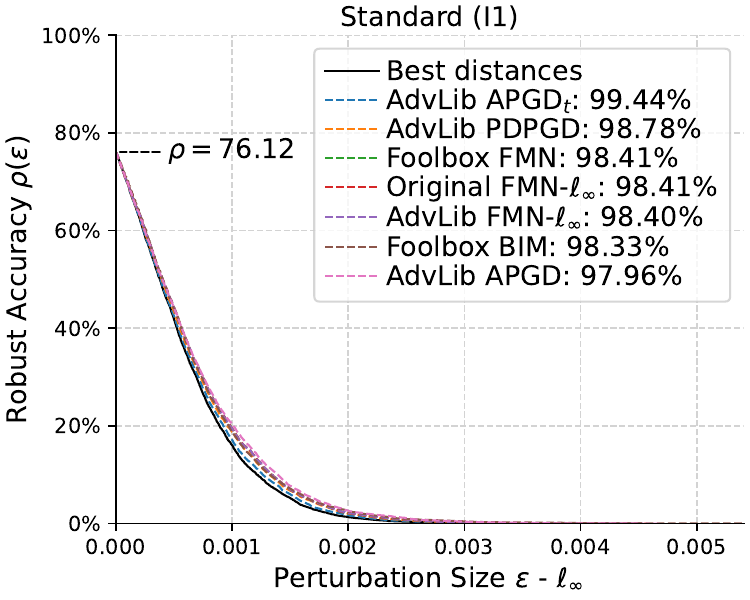}
    \caption{Robustness evaluation curves for the best $\ell_p$-norm attacks against \standardimagenet in ImageNet.}
    \label{fig:standard_security_evaluation_curves_imagenet}
\end{figure*}

\begin{figure*}[!htb]
    \centering
    \includegraphics[width=0.245\textwidth]{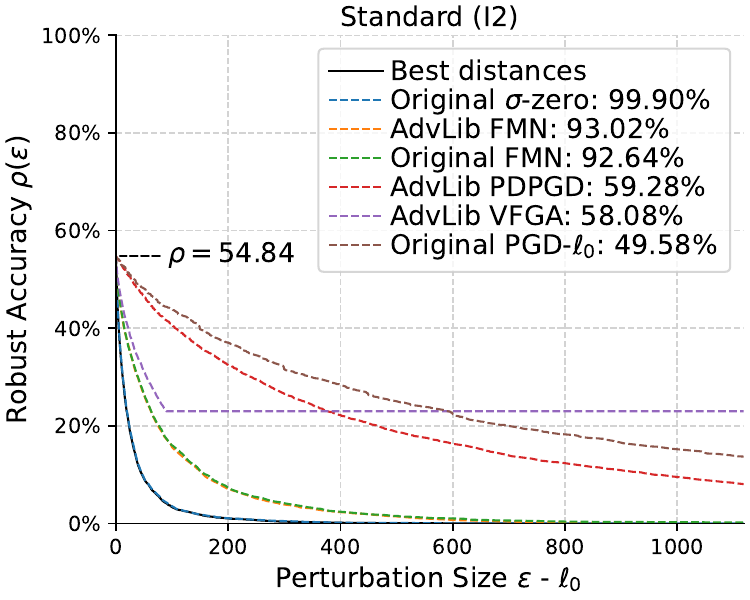}
    \includegraphics[width=0.245\textwidth]{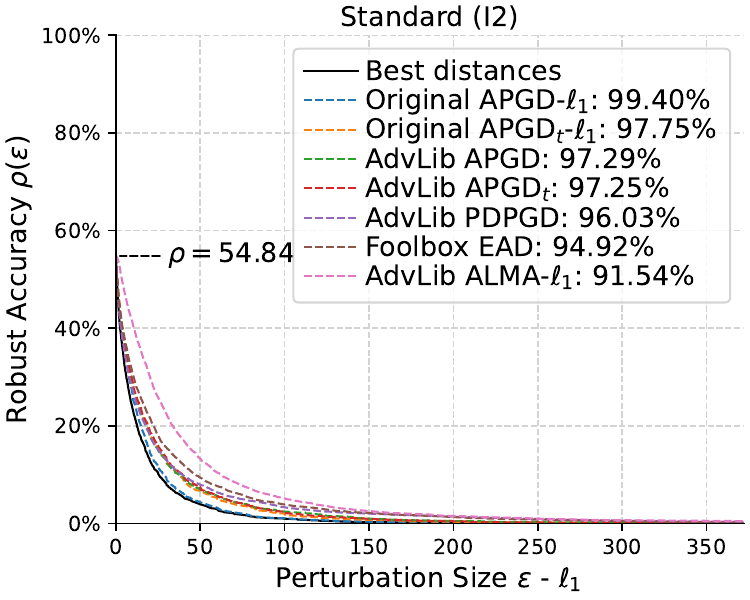}
    \includegraphics[width=0.245\textwidth]{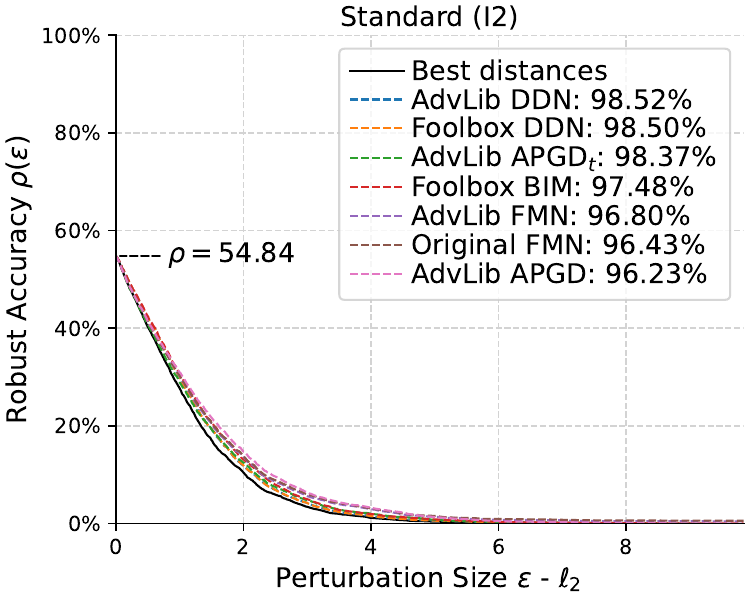}
    \includegraphics[width=0.245\textwidth]{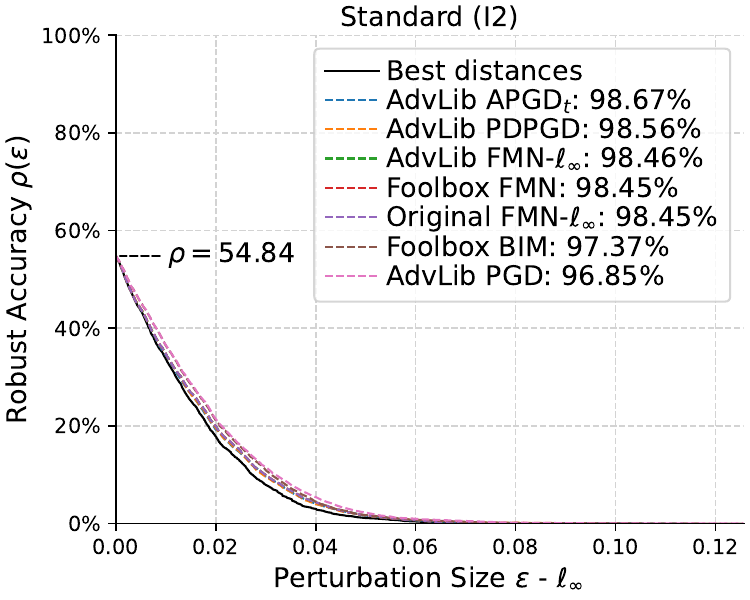}
    \caption{Robustness evaluation curves for the best $\ell_p$-norm attacks against \wong in ImageNet.}
    \label{fig:wong_security_evaluation_curves_imagenet}
\end{figure*}

\begin{figure*}[!htb]
    \centering
    \includegraphics[width=0.245\textwidth]{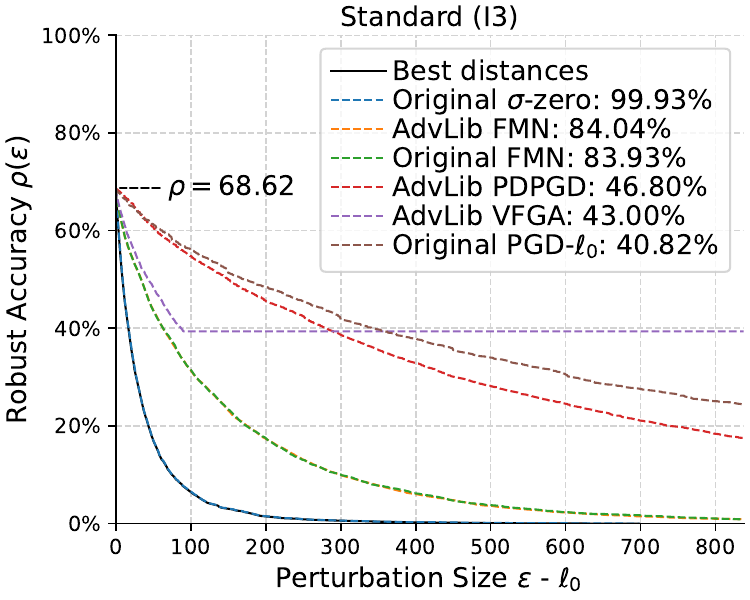}
    \includegraphics[width=0.245\textwidth]{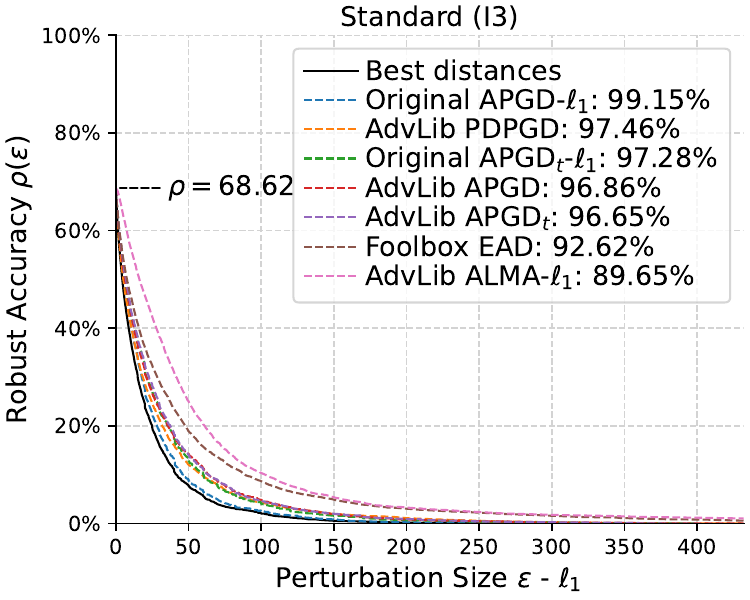}
    \includegraphics[width=0.245\textwidth]{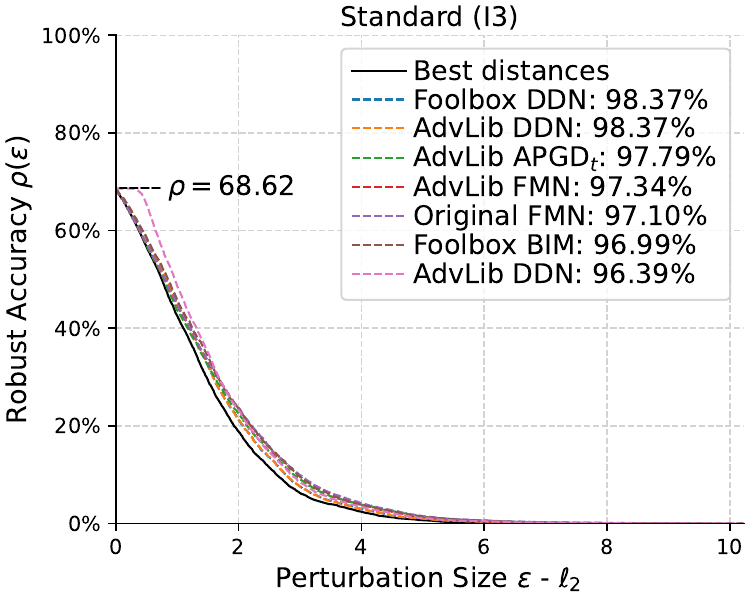}
    \includegraphics[width=0.245\textwidth]{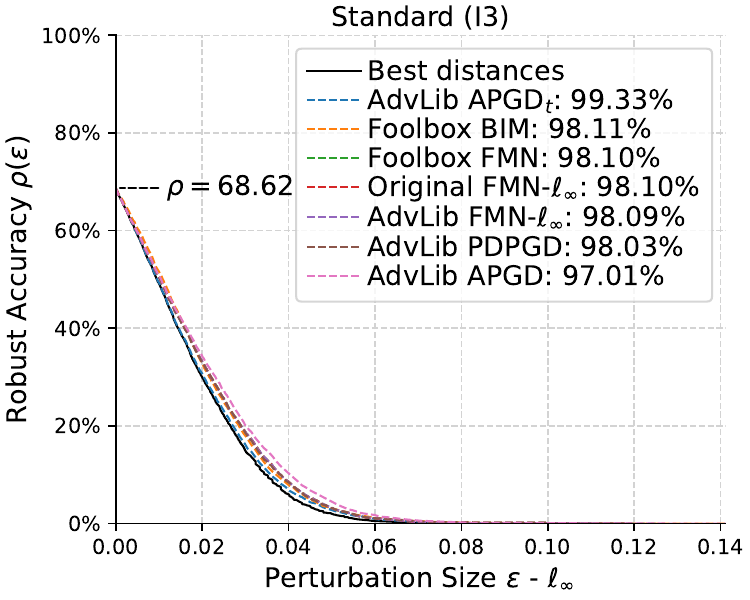}
    \caption{Robustness evaluation curves for the best $\ell_p$-norm attacks against \salman in ImageNet.}
    \label{fig:salman_security_evaluation_curves_imagenet}
\end{figure*}

\begin{figure*}[!htb]
    \centering
    \includegraphics[width=0.245\textwidth]{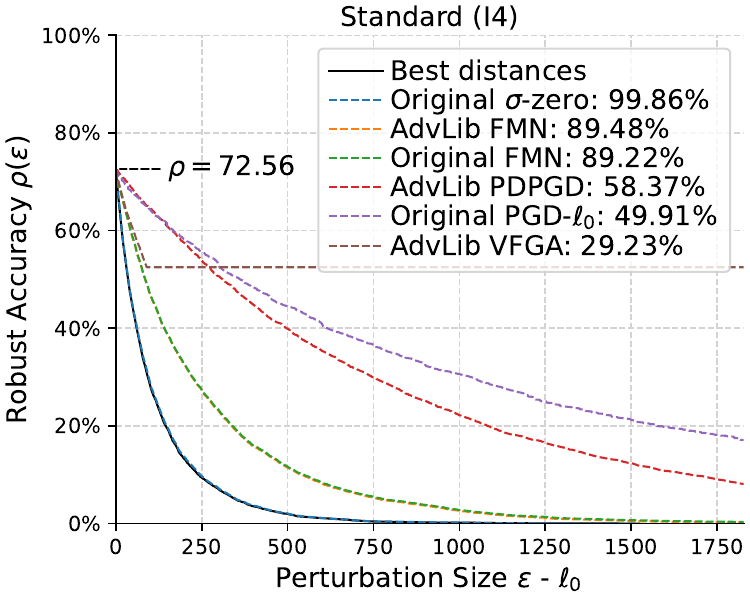}
    \includegraphics[width=0.245\textwidth]{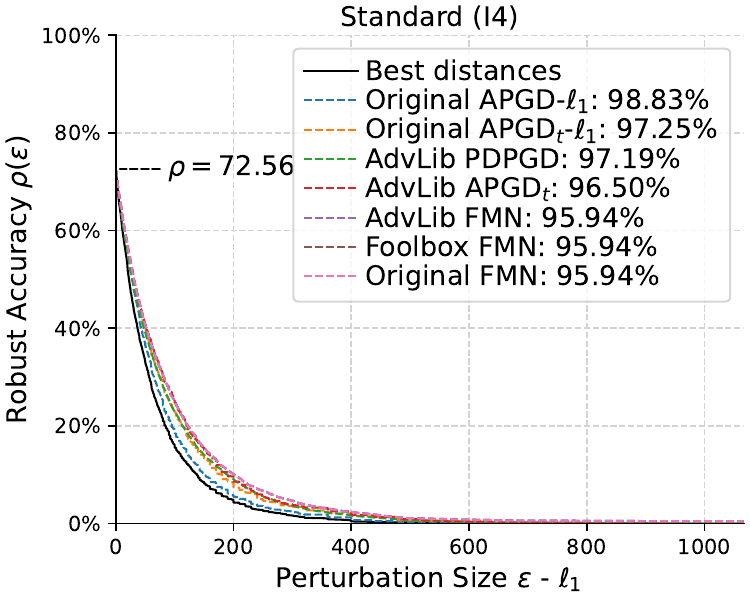}
    \includegraphics[width=0.245\textwidth]{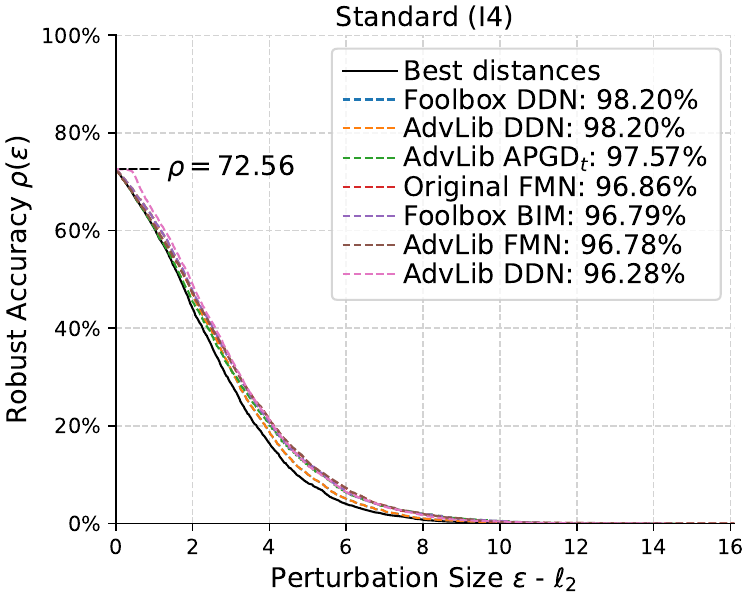}
    \includegraphics[width=0.245\textwidth]{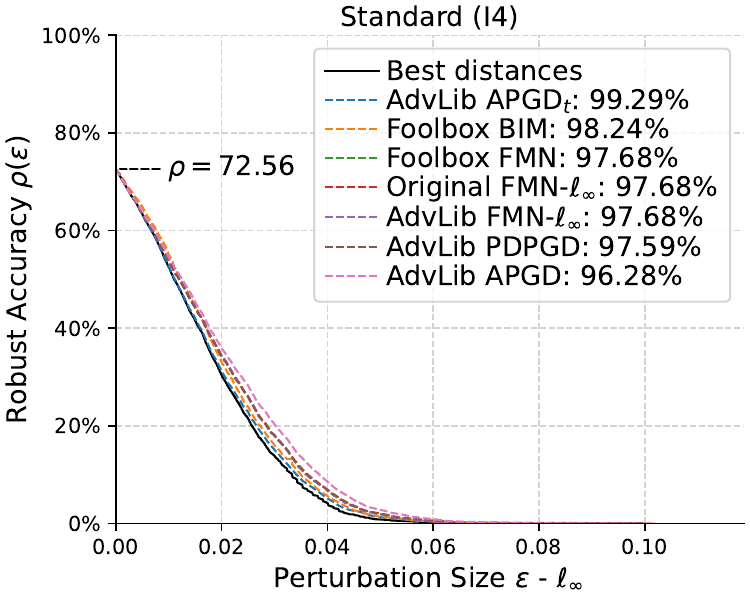}
    \caption{Robustness evaluation curves for the best $\ell_p$-norm attacks against \debenedetti in ImageNet.}
    \label{fig:debenedetti_security_evaluation_curves_imagenet}
\end{figure*}

\input{tables/best_attacks_batch_size1}

\clearpage
\input{tables/appendix/lzero-cifar}
\input{tables/appendix/lone-cifar}
\input{tables/appendix/ltwo-cifar}
\input{tables/appendix/linf-cifar}

%% file: appendix/background.tex
\section{Gradient-based Attacks}
\label{sec:advattacks}

We present the background formalism for \maxloss and \minnorm gradient-based attacks against machine learning models. Furthermore, we list the attacks considered in this study in~\Cref{table:adv_attacks_categorization}.

\paragraph{Fixed-budget Attacks.}  These attacks (\maxloss) minimize the loss and set the perturbation size as the constraint. They optimize the following problem:
\begin{align}
\minimize_{\vct \delta}  && L(\vct x + \vct \delta, y, \vct \theta) \, ,\label{eq:obj_max_conf} \\
\subjectto && \|\vct \delta \|_{p} \leq \epsilon \, ,  \label{eq:constr_max_conf}\\ 
	&&  \vct x + \vct \delta \in [0,1]^\con d \label{eq:bounds_max_conf}\, ,
\end{align}
where $\| \cdot\|_{p}$ indicates the $\ell_p$-norm operator. 
The loss $L$ in \cref{eq:obj_max_conf} is defined accordingly to the goal of the attack, \ie untargeted or targeted misclassification.
For example, this group of attacks include the Projected Gradient Descent (\pgd) attack by \citet{madry18-iclr}. 

\paragraph{Minimum-norm Attacks.} 
These attacks (\minnorm) minimize the perturbation size under the constraint that the sample is misclassified~\cite{carlini17-sp,Rony2020AugmentedLA,rony2019decoupling,pintor2021fast}. They are formulated as follows:
\begin{eqnarray}
    \label{eq:obj_min_distance}	\minimize_{\vct \delta}  && \|\vct \delta \|_{p}  \\
    \label{eq:constr_min_distance}	\subjectto  &&  f(\vct x + \vct \delta, \vct \theta) \neq f(\vct x, \vct \theta)\, , \\
    \label{eq:bounds_min_distance}	&& \vct x + \vct \delta \in [0,1]^\con d \, 
 ,
\end{eqnarray}
where $|| \cdot||_{p}$ indicates the $\ell_p$-norm operator. 
However, this problem cannot be solved directly by gradient descent since the misclassification constraint \eqref{eq:constr_min_distance} is not differentiable.
In contrast to \maxloss attacks, which maximize confidence for predicting a wrong class within a given perturbation budget, \minnorm attacks aim at finding the closest adversarial example to each input. This requires solving a more complex problem, as it amounts not only to finding a valid adversarial example, but also to minimizing its perturbation size~\cite{Rony2020AugmentedLA}.

%% file: appendix/implementation.tex
\section{AttackBench - Implementation details}\label{appendix:implementation}
We provide here specific implementation details that characterize our methodology.

\subsection{Samples Hashing}
\label{appendix:sample_hashing}
\ab uses the SHA-512 hashing algorithm, standardized by NIST~\cite{Dang2012SecureHS}, to associate each sample with a unique compressed representation for accurately tracking and identifying samples throughout the benchmarking process.
The choice of SHA-512 is motivated by its cryptographic strength and resistance to collisions.
Also, the choice of sample hashing seamlessly solves the problem of evaluating attacks on the same set of samples.
Controversy, one common strategy for evaluating attacks is to use the entire testing set of widely-used datasets (like CIFAR-10), but this could be challenging and unnecessary, especially when considering large-scale datasets such as ImageNet.
Thus, researchers often expedite comparisons by relying on arbitrary data subsets~\cite{pintor2021fast}, difficult to reconstruct without curated metadata (\eg the indexes of the samples used). 
Moreover, the content of these datasets may undergo alterations over time~\cite{carlini2023poisoning}, potentially compromising the integrity of the comparison.
Thus, we aim to guarantee that future evaluations will consider the same subset of data to establish a common and unique baseline.
Another trivial solution would be sharing the samples used during the evaluations, but this would expose us to the risk of potential license infringements. 
Thus, \ab employs a hashing algorithm to associate each sample with a unique compressed representation, significantly reducing the risk of sample replacement and collaterally containing the size of saved information, facilitating result sharing.

\subsection{Query-budget Termination.}\label{appendix:query-budget}
Incorporating the query-halting feature while attacks are running introduces the challenge of avoiding any intervention in their implementation. \ab must accept attacks as input without modifying their internal code to avoid introducing potential pitfalls that could compromise the validity of the evaluation.
To overcome this challenge, we introduce a \texttt{BenchModel} class to encapsulate the models under assessment (\autoref{alg:benchmodel}). 
During an attack, this class counts all the backward and forward \textit{hooks}\footnote{\url{https://pytorch.org/docs/stable/generated/torch.nn.modules.module.register_module_forward_hook.html}} operations invoked on the model, without interfering with their output.
Once the query limit is reached, \ab blocks replaces the prediction with a correct one (for an untargeted attack) 
in the forward pass, and zeroing the gradient in the backward pass. 
This ensures that the attack either can continue without causing any run-time error, even if the number of iterations surpasses the query budget, and that the optimization is effectively halted, due to the absence of gradients.
Special care was taken to ensure that these hooks are compatible with batching strategies (some samples might reach the limit before others), and do not disrupt the internal adversarial perturbation tracking mechanisms of the attack being evaluated.
Ultimately, the forward and backward hooks implemented by \ab track the progress of adversarial example optimization and save the best adversarial perturbation found. 
This approach addresses the issue identified by Pintor \etal \cite{pintor2022indicators}, who demonstrated that certain attack implementations return the adversarial perturbation from the last iteration, even though it may not be the best perturbation found during optimization. Thus, within \ab, we ensure to report the best-case performance of the attacks.

\subsection{Optimality Score Development}\label{appendix:optimality_development_appendix}
The development of this measure involves a structured three-step process, highlighted in \autoref{alg:atkbenchmark_code}.
Firstly, we subject a target model to an exhaustive range of attacks $\{\attack^{1}, \dots, \attack^{N}\}$, recording the minimum best distances $d^\star$ required to render each sample adversarial (\autoref{alg:test_attacks}). 
We then use the collected distances to construct an empirically derived curve \Aopt, which effectively captures the model's optimal trade-off between robustness and perturbation size, for the best empirical attack \bestattack (\autoref{alg:compile_results}). This curve maps the minimum perturbation generated by the array of attacks for every validation sample.
Finally, we to quantitatively assess the \textit{optimality} of a specific attack, we propose calculating the difference in the areas under the curves obtained from the empirically derived curve and the curve generated by the attack under evaluation (\autoref{alg:local_optimality}).

\subsection{Including New Attacks into the Benchmark}\label{appendix:new-attacks}
\ab facilitates the inclusion of new attacks thanks to three main characteristics.
First, \ab treats attacks as black boxes, thus preventing contributors from specifically adapting their code to integrate and test the attack with \ab.
Second, the output of \autoref{algo:atkbenchmark_code} is a readable \textit{json} file containing all the considered metrics (e.g., hashed input, the number of forwards and backward passes used, the attack execution time, and the distances of the returned and optimal adversarial perturbation). 
Consequently, attack developers can integrate novel attacks by sharing their \textit{json} files within the collection of benchmark results. 
While the runtime may not be directly comparable unless using the same hardware architecture, the optimality measure will still be valid, as it only considers the distances of the adversarial perturbations. 
Finally, the \textit{optimality} measure proposed in \ab can be continually updated with the results of each attack efficiently. Including each new attack does not require indeed possessing data for all attacks to assess a new one; the current empirically optimal curve alone suffices. 
Once the results of the novel attack are uploaded, \ab updates the empirically optimal curve if the novel attack advances the state of the art, and refreshes the leaderboard for all attacks by recomputing \autoref{eq:local_optimality}.

\subsection{Search Strategy for \maxloss attacks}\label{appendix:search-fixed-budget}
\maxloss attacks, as explained in \autoref{sec:advattacks}, find the perturbation within a maximum budget $\budgetsize$.
For these attacks, we adopt the search strategy implemented in \cite{Rony2020AugmentedLA, Cina2024SigmaZero} to find the smallest budget $\budgetsize^\star$ for which the attack can successfully find an adversarial perturbation. 
Specifically, starting from $\budgetsize^{(0)}$, we run the attack (corresponding to one search step) and multiply (divide) $\budgetsize$ by $2$ if the attack fails (succeeds). 
Once the attack succeeds, we have a lower and upper bound for $\budgetsize^\star$, so we use the remaining search steps to perform a binary search to refine the solution. In practice, we perform $10$ search steps and adapt the number of attack steps to reach the overall $2\,000$ query budget. We use $\budgetsize^{(0)}=100,10,1,$ and $\nicefrac{1}{255}$ for the $\ell_0$, $\ell_1$, $\ell_2$, $\ell_\infty$-norm \maxloss attacks respectively.

%% file: checklist.tex
\section*{Reproducibility Checklist}

This paper:

\begin{itemize}

\item Includes a conceptual outline and/or pseudocode description of AI methods introduced. {Yes. See Figure 1, Algorithms 1 and 2.} 

\item Clearly delineates statements that are opinions, hypothesis, and speculation from objective facts and results. Yes

\item Provides well marked pedagogical references for less-familiare readers to gain background necessary to replicate the paper {Yes, see Section 2 and Appendix A.}

\item Does this paper make theoretical contributions? {No. It provides methodological and practical contributions for the fair assessment of the effectiveness of gradient-based attacks.}

\item Does this paper rely on one or more datasets? {Yes.}
\begin{itemize}
    \item A motivation is given for why the experiments are conducted on the selected datasets. {No, a specific motivation is not provided for the selection of datasets, but we utilize standard and well-known datasets commonly used in this domain.}
    \item All novel datasets introduced in this paper are included in a data appendix. {NA.}
    \item All novel datasets introduced in this paper will be made publicly available upon publication of the paper with a license that allows free usage for research purposes. {NA.}
    \item All datasets drawn from the existing literature (potentially including authors’ own previously published work) are accompanied by appropriate citations. {Yes. See Section 4.1.}
    \item All datasets drawn from the existing literature (potentially including authors’ own previously published work) are publicly available. {Yes.}
    \item All datasets that are not publicly available are described in detail, with explanation why publicly available alternatives are not scientifically satisficing. {NA.}
\end{itemize}

\item Does this paper include computational experiments? Yes.
\begin{itemize}
    \item  Any code required for pre-processing data is included in the appendix. {Yes, preprocessing details are included in the source code and follow standard approaches.}
    \item All source code required for conducting and analyzing the experiments is included in a code appendix. {Yes.}
    \item All source code required for conducting and analyzing the experiments will be made publicly available upon publication of the paper with a license that allows free usage for research purposes. {Yes.}
    \item All source code implementing new methods have comments detailing the implementation, with references to the paper where each step comes from. {Yes.}

    \item If an algorithm depends on randomness, then the method used for setting seeds is described in a way sufficient to allow replication of results. {yes.}

    \item This paper specifies the computing infrastructure used for running experiments (hardware and software), including GPU/CPU models; amount of memory; operating system; names and versions of relevant software libraries and frameworks. {Yes, we provide detailed information in the paper and the source code README on how to replicate the experimental computing infrastructure, including hardware and software specifications.}
    
    \item This paper formally describes evaluation metrics used and explains the motivation for choosing these metrics. {Yes, See Section 3.2}

    \item This paper states the number of algorithm runs used to compute each reported result. {Yes. See Section 4.2}

    \item Analysis of experiments goes beyond single-dimensional summaries of performance (e.g., average; median) to include measures of variation, confidence, or other distributional information. {Yes, the optimality metric looks at distributional information.}

    \item The significance of any improvement or decrease in performance is judged using appropriate statistical tests (e.g., Wilcoxon signed-rank). {No.}
    
    \item This paper lists all final (hyper-)parameters used for each model/algorithm in the paper’s experiments. {Yes, we clarify in Section 4.1 that default hyperparameters were used for the attacks.}
    
    \item This paper states the number and range of values tried per (hyper-) parameter during development of the paper, along with the criterion used for selecting the final parameter setting. {NA.}
\end{itemize}

\end{itemize}

%% file: tables/attacks_libraries.tex
\begin{table*}[!htbp]
\small
    \centering
    \caption{List earlier  adversarial attacks and their implementations we have benchmarked with \ab. For each attack, we mark the supported $\ell_p$ norms and the libraries implementing them. The libraries considered are: Original (\originalshort), AdversarialLib (\advlibshort), FoolBox (\foolboxshort), Art (\artshort), TorchAttacks (\torchattacksshort), DeepRobust (\deeprobustshort), and CleverHans (\cleverhansshort).}
    \label{table:adv_attacks_categorization}
    \setlength\tabcolsep{2pt}
\begin{tabular}{@{}ll@{\hskip 0.3in}llll@{\hskip 0.3in}ccccccc@{}}
\toprule
\multicolumn{1}{l}{\multirow{2}{*}{\textbf{Name}}} & \multicolumn{1}{l}{\multirow{2}{*}{\textbf{Type}}} & \multicolumn{4}{c}{\textbf{Norms}} & \multicolumn{7}{c}{\textbf{Library Implementation}} \\ \cmidrule{3-6}\cmidrule{6-13}
 &  & $\ell_0$ & $\ell_1$ & $\ell_2$ & $\ell_\infty$ & \originalshort & \advlibshort & \foolboxshort & \artshort & \torchattacksshort & \deeprobustshort & \cleverhansshort \\ \hline
\fgsm~\cite{goodfellow15-iclr-explaining}  & \maxloss & \tcheck &  &  & \tcheck &  &  & \tcheck & \tcheck & \tcheck & \tcheck & \tcheck \\
\jsma~\cite{papernot16-sp}  & \minnorm & \tcheck &  &  &  &  &  &  & \tcheck &  &  &  \\
\bim~\cite{kurakin16adversarialexamples}  & \maxloss &  &  &  & \tcheck &  &  & \tcheck & \tcheck &  &  &  \\
\dfool~\cite{moosavi16-deepfool}   & \minnorm &  & \tcheck & \tcheck & \tcheck & \tcheck &  & \tcheck & \tcheck & \tcheck & \tcheck &  \\
\cw~\cite{carlini17-sp}  & \minnorm & \tcheck &  & \tcheck & \tcheck &  & \tcheck & \tcheck & \tcheck & \tcheck & \tcheck & \tcheck \\
\ead~\cite{chen2018ead}  & \minnorm &  & \tcheck & \tcheck & \tcheck &  &  & \tcheck & \tcheck &  &  &  \\
\pgd~\cite{madry18-iclr}  & \maxloss &  &  & \tcheck & \tcheck &  & \tcheck & \tcheck & \tcheck & \tcheck & \tcheck & \tcheck \\
\bb~\cite{brendel2020accurate}  & \minnorm & \tcheck & \tcheck & \tcheck & \tcheck &  &  & \tcheck & \tcheck &  &  &  \\
\ddn~\cite{rony2019decoupling}   & \minnorm &  &  & \tcheck &  &  & \tcheck & \tcheck &  &  &  &  \\
\pgdlzero~\cite{Croce2019SparseAI}   & \maxloss & \tcheck &  &  &  &  &  &  &  &  &  &  \\
\sfool~\cite{Modas2018SparseFoolAF}  & \minnorm &  & \tcheck &  &  &  &  &  &  & \tcheck &  &  \\
\tr~\cite{Yao2018TrustRB}  & \minnorm &  &  & \tcheck & \tcheck & \tcheck & \tcheck &  &  &  &  &  \\
\fab~\cite{croce2020minimally}  & \minnorm &  & \tcheck & \tcheck & \tcheck & \tcheck & \tcheck &  &  & \tcheck &  &  \\
\apgd~\cite{croce2020reliable}  & \maxloss &  &  & \tcheck & \tcheck & \tcheck & \tcheck &  & \tcheck & \tcheck &  &  \\
\apgdt~\cite{croce2020reliable}  & \maxloss &  &  & \tcheck & \tcheck & \tcheck & \tcheck &  &  &  &  &  \\
\alma~\cite{Rony2020AugmentedLA}  & \minnorm &  & \tcheck & \tcheck &  &  & \tcheck &  &  &  &  &  \\
\apgdlone~\cite{Croce2021MindTB}  & \maxloss &  & \tcheck &  &  & \tcheck & \tcheck &  &  &  &  &  \\
\fmn~\cite{pintor2021fast}  & \minnorm & \tcheck & \tcheck & \tcheck & \tcheck & \tcheck & \tcheck & \tcheck &  &  &  &  \\
\pdgd ~\cite{Matyasko2021PDPGDPP}  & \minnorm & \tcheck & \tcheck & \tcheck & \tcheck &  & \tcheck &  &  &  &  &  \\
\pdpgd ~\cite{Matyasko2021PDPGDPP} & \minnorm & \tcheck & \tcheck & \tcheck & \tcheck &  & \tcheck &  &  &  &  &  \\
\vfga~\cite{Hajri2020StochasticSA}   & \minnorm & \tcheck &  &  &  &  & \tcheck &  &  &  &  &  \\ 
\sigmazero~\cite{Cina2024SigmaZero}   & \minnorm & \tcheck &  &  &  &  \tcheck & &  &  &  &  &  \\ 
\bottomrule
\end{tabular}
\end{table*}

%% file: tables/best_attacks_batch_size1.tex
\begin{table*}[htbp]
\caption{Top $\ell_\infty$-norm performing attacks with batch size equals to 1. For each attack, we list the library implementations that offer the same or very similar results. When multiple libraries are present, the runtime is reported for the emphasized one.}
\centering
\small
\label{tab:best_performing_batchsizeone}
\setlength\tabcolsep{3pt}
\setlength{\dashlinedash}{3.pt}
\sisetup{table-auto-round}
\begin{tabular}{lllc*{2}{S[table-format=3.1, fixed-exponent=-2, drop-exponent=true, exponent-mode=fixed, drop-zero-decimal]}S[table-format=4]S[table-format=4]S[table-format=4.1]}
\toprule
\textbf{Dataset} & \textbf{$\ell_p$} & \textbf{Attack} & \textbf{Library} & \textbf{ASR (\%)} & \textbf{Optimality (\%)} & \textbf{\#Forwards} & \textbf{\#Backwards} & \textbf{t(s)} \\ 
\midrule
\multirow{5}{*}{\begin{turn}{0}CIFAR-10\end{turn}} 
& & \apgd & TorchAttacks, Original, AdvLib & 1.0 & 0.985 & 802.3 & 782.3 & 9.99 \\
& & \apgdt & Original, AdvLib & 1.0 & 0.977 & 626.0 & 580.8 & 8.77 \\
& & \bim & FoolBox & 1.0 & 0.951 & 1000.0 & 990.0 & 10.17 \\
& & \pdpgd & AdvLib & 0.998 & 0.946 & 1000.0 & 1000.0 & 12.39 \\
&\multirow{-5}{*}{$\ell_\infty$} & \pgd & AdvLib, FoolBox & 1.0 & 0.936 & 1000.0 & 990.0 & 10.04 \\  \bottomrule
\end{tabular}
\end{table*}

%% file: tables/appendix/lzero-cifar.tex
\begin{table*}[htbp]
\small\centering
\setlength\tabcolsep{7pt}
\caption{Local optimality for $\ell_0$-norm attacks.}
\label{tab:lzero_all}
\sisetup{table-auto-round}
\begin{tabular}{cl*{2}{S[table-format=3.1, fixed-exponent=-2, drop-exponent=true, exponent-mode=fixed, drop-zero-decimal]}ccS[table-format=4.1]}
\toprule\textbf{Model} & \textbf{Attack} & \textbf{ASR} & \textbf{Optimality} & \textbf{\#Forwards} & \textbf{\#Backwards} & \textbf{ExecTime} \\ \cmidrule(l{0.5em}r{0.5em}){1-7}
\multirow{6}{*}{\standardcifar} & Original $\sigma$-zero & 1 & 0.972 & 999 & 999 & 269.682 \\
 & AdvLib \fmn & 1 & 0.875 & 1000 & 1000 & 266.6437 \\
 & Original \fmn & 1 & 0.854 & 892 & 892 & 264.5156 \\
 & AdvLib \vfga & 0.987 & 0.669 & 359 & 17 & 50.8855 \\
 & Original \pgdlzero & 1 & 0.51 & 974 & 955 & 423.4975 \\
 & AdvLib\pdpgd & 1 & 0.284 & 839 & 838 & 264.5755 \\\cmidrule(l{0.5em}r{0.5em}){1-7}
\multirow{6}{*}{\zhang} & Original $\sigma$-zero & 1 & 0.993 & 998 & 998 & 4.3377 \\
 & AdvLib \vfga & 0.98 & 0.943 & 183 & 8 & 1.2691 \\
 & Original \pgdlzero & 1 & 0.869 & 812 & 794 & 69.4237 \\
 & AdvLib \fmn & 0.936 & 0.845 & 1000 & 1000 & 3.4154 \\
 & Original \fmn & 0.936 & 0.845 & 999 & 999 & 4.8291 \\
 & AdvLib\pdpgd & 0.979 & 0.585 & 997 & 997 & 4.9836 \\ \cmidrule(l{0.5em}r{0.5em}){1-7}
\multirow{6}{*}{\stutz} & Original $\sigma$-zero & 1 & 0.992 & 999 & 999 & 38.4524 \\
 & AdvLib \fmn & 1 & 0.834 & 1000 & 1000 & 37.6205 \\
 & Original \fmn & 1 & 0.805 & 842 & 842 & 40.8738 \\
 & AdvLib \vfga & 1 & 0.742 & 136 & 6 & 3.1348 \\
 & Original \pgdlzero & 1 & 0.426 & 953 & 934 & 134.6154 \\
 & AdvLib\pdpgd & 1 & 0.145 & 848 & 848 & 39.9815 \\ \cmidrule(l{0.5em}r{0.5em}){1-7}
\multirow{6}{*}{\xiao} & Original $\sigma$-zero & 1 & 0.967 & 999 & 999 & 46.335 \\
 & AdvLib \fmn & 1 & 0.928 & 999 & 999 & 45.4927 \\
 & Original \fmn & 0.998 & 0.916 & 690 & 690 & 47.2757 \\
 & Original \pgdlzero & 1 & 0.91 & 880 & 861 & 286.1117 \\
 & AdvLib \vfga & 0.867 & 0.875 & 496 & 23 & 15.5488 \\
 & AdvLib\pdpgd & 0.997 & 0.509 & 919 & 919 & 47.1681 \\ \cmidrule(l{0.5em}r{0.5em}){1-7}
\multirow{6}{*}{\wang} & Original $\sigma$-zero & 1 & 0.998 & 999 & 999 & 292.1994 \\
 & AdvLib \fmn & 1 & 0.784 & 999 & 999 & 278.8026 \\
 & AdvLib \vfga & 0.888 & 0.782 & 768 & 36 & 106.1816 \\
 & Original \fmn & 1 & 0.78 & 911 & 911 & 277.2555 \\
 & Original \pgdlzero & 1 & 0.618 & 978 & 959 & 544.9969 \\
 & AdvLib\pdpgd & 0.999 & 0.442 & 962 & 962 & 280.3719 \\ 
 \bottomrule
\end{tabular}
\end{table*}

%% file: tables/appendix/lone-cifar.tex
\begin{table*}[htbp]
\small\centering
\setlength\tabcolsep{5pt}
\caption{Local optimality for $\ell_1$-norm attacks.}
\label{tab:lone_all_1}
\sisetup{table-auto-round}
\begin{tabular}{cl*{2}{S[table-format=3.1, fixed-exponent=-2, drop-exponent=true, exponent-mode=fixed, drop-zero-decimal]}ccS[table-format=4.1]}
\toprule\textbf{Model} & \textbf{Attack} & \textbf{ASR} & \textbf{Optimality} & \textbf{\#Forwards} & \textbf{\#Backwards} & \textbf{ExecTime} \\ \toprule
\multirow{21}{*}{\standardcifar} & AdvLib \alma & 1.00 & 0.97 & 993 & 993 & 262.89 \\
 & AdvLib \fmn & 0.99 & 0.97 & 1,000 & 1,000 & 268.69 \\
 & FoolBox \fmn & 0.99 & 0.97 & 1,000 & 1,000 & 265.75 \\
 & Original \fmn & 0.99 & 0.97 & 1,000 & 1,000 & 264.31 \\
 & Original \apgdlone & 1.00 & 0.96 & 958 & 938 & 523.04 \\
 & AdvLib \pdpgd & 1.00 & 0.96 & 992 & 992 & 263.28 \\
 & AdvLib \apgd & 1.00 & 0.93 & 1,000 & 990 & 274.91 \\
 & Original \apgdt & 1.00 & 0.88 & 667 & 622 & 546.57 \\
 & AdvLib \apgdt & 1.00 & 0.87 & 590 & 570 & 160.77 \\
 & AdvLib \fab & 0.95 & 0.83 & 361 & 1,620 & 267.48 \\
 & Original \fab & 0.94 & 0.83 & 311 & 1,544 & 255.77 \\
 & FoolBox \ead & 1.00 & 0.82 & 405 & 207 & 80.54 \\
 & Art \ead & 1.00 & 0.73 & 334 & 1,665 & 275.16 \\
 & TorchAttacks \fab & 0.05 & 0.61 & 311 & 1,544 & 256.94 \\
 & Art \pgd & 0.84 & 0.48 & 1,010 & 980 & 579.31 \\
 & FoolBox \pgd & 1.00 & 0.47 & 1,000 & 990 & 640.9 \\
 & FoolBox \bim & 1.00 & 0.46 & 220 & 200 & 284.37 \\
 & FoolBox \fgsm & 0.99 & 0.26 & 40 & 20 & 30.32 \\
 & Art \fgsm & 0.99 & 0.26 & 40 & 20 & 31.49 \\
 & FoolBox \bb & 0.05 & 0.24 & 634 & 0 & 96.65 \\
 & Art \apgd & 1.00 & 0.12 & 1,044 & 457 & 456.88 \\ \cmidrule(l{0.5em}r{0.5em}){1-7}
\multirow{21}{*}{\zhang} & AdvLib \pdpgd & 0.99 & 0.91 & 997 & 997 & 5.2 \\
 & Original \apgdlone & 1.00 & 0.83 & 406 & 386 & 143.97 \\
 & Original \apgdt & 1.00 & 0.81 & 306 & 275 & 163.38 \\
 & AdvLib \apgdt & 1.00 & 0.77 & 637 & 617 & 12.69 \\
 & AdvLib \fmn & 0.91 & 0.75 & 1,000 & 1,000 & 4.03 \\
 & FoolBox \fmn & 0.91 & 0.75 & 1,000 & 1,000 & 5.09 \\
 & Original \fmn & 0.91 & 0.75 & 1,000 & 1,000 & 5.32 \\
 & AdvLib \fab & 0.98 & 0.74 & 360 & 1,619 & 3.66 \\
 & Original \fab & 0.97 & 0.71 & 128 & 635 & 2.85 \\
 & FoolBox \bim & 1.00 & 0.67 & 220 & 200 & 21.22 \\
 & Art \pgd & 0.84 & 0.66 & 1,010 & 980 & 35.43 \\
 & FoolBox \ead & 1.00 & 0.61 & 355 & 182 & 1.01 \\
 & FoolBox \pgd & 1.00 & 0.60 & 999 & 989 & 56.96 \\
 & AdvLib \apgd & 1.00 & 0.59 & 999 & 989 & 10.96 \\
 & AdvLib \alma & 1.00 & 0.55 & 841 & 841 & 5.37 \\
 & TorchAttacks \fab & 0.61 & 0.54 & 128 & 635 & 2.85 \\
 & Art \fgsm & 0.98 & 0.48 & 40 & 20 & 2.25 \\
 & FoolBox \fgsm & 0.98 & 0.48 & 40 & 20 & 2.21 \\
 & Art \ead & 0.79 & 0.41 & 334 & 1,665 & 6.39 \\
 & Art \apgd & 1.00 & 0.33 & 441 & 179 & 38.02 \\
 & FoolBox \bb & 0.61 & 0.17 & 637 & 0 & 1.91 \\ \cmidrule(l{0.5em}r{0.5em}){1-7}
\multirow{21}{*}{\xiao} & Original \apgdt & 1.00 & 0.93 & 558 & 518 & 526.82 \\
 & FoolBox \fmn & 1.00 & 0.91 & 999 & 999 & 47.98 \\
 & Original \fmn & 1.00 & 0.91 & 999 & 999 & 46.69 \\
 & AdvLib \fmn & 1.00 & 0.91 & 1,000 & 1,000 & 46.13 \\
 & Original \apgdlone & 1.00 & 0.91 & 668 & 648 & 547.49 \\
 & AdvLib \pdpgd & 1.00 & 0.90 & 998 & 998 & 46.9 \\
 & AdvLib \apgdt & 1.00 & 0.89 & 690 & 670 & 43.81 \\
 & AdvLib \apgd & 1.00 & 0.84 & 998 & 988 & 52.96 \\
 & FoolBox \ead & 1.00 & 0.83 & 292 & 150 & 10.62 \\
 & FoolBox \bim & 1.00 & 0.64 & 220 & 200 & 160.07 \\
 & FoolBox \pgd & 1.00 & 0.61 & 999 & 989 & 317.41 \\
 & Art \pgd & 0.42 & 0.58 & 1,010 & 980 & 169.83 \\
 & TorchAttacks \fab & 0.35 & 0.54 & 155 & 790 & 31.85 \\
 & Art \ead & 0.47 & 0.38 & 334 & 1,665 & 44.5 \\
 & Art \apgd & 1.00 & 0.37 & 773 & 331 & 228.78 \\
 & Art \fgsm & 0.99 & 0.34 & 40 & 20 & 18.39 \\
 & FoolBox \fgsm & 0.99 & 0.34 & 40 & 20 & 16.76 \\
 & AdvLib \alma & 1.00 & 0.17 & 805 & 805 & 47.88 \\
 & Original \fab & 0.92 & 0.10 & 155 & 790 & 32.23 \\
 & FoolBox \bb & 0.35 & 0.07 & 629 & 0 & 20.84 \\
 & AdvLib \fab & 0.96 & 0.02 & 216 & 1,012 & 41.35 \\ 
 \bottomrule
 \end{tabular}
\end{table*}

 \begin{table*}[htbp]
\small\centering
\setlength\tabcolsep{5pt}
\caption{Local optimality for $\ell_1$-norm attacks (pt2).}
\label{tab:lone_all_2}
\sisetup{table-auto-round}
\begin{tabular}{cl*{2}{S[table-format=3.1, fixed-exponent=-2, drop-exponent=true, exponent-mode=fixed, drop-zero-decimal]}ccS[table-format=4.1]}
\toprule\textbf{Model} & \textbf{Attack} & \textbf{ASR} & \textbf{Optimality} & \textbf{\#Forwards} & \textbf{\#Backwards} & \textbf{ExecTime} \\ \toprule
 \cmidrule(l{0.5em}r{0.5em}){1-7}
\multirow{21}{*}{\wang} & Original \apgdt & 1.00 & 0.97 & 726 & 679 & 860.59 \\
 & AdvLib \pdpgd & 1.00 & 0.97 & 998 & 998 & 279.6 \\
 & Original \apgdlone & 1.00 & 0.97 & 935 & 915 & 892.36 \\
 & AdvLib \apgdt & 1.00 & 0.96 & 657 & 637 & 190.46 \\
 & AdvLib \apgd & 1.00 & 0.96 & 1,000 & 990 & 284.24 \\
 & AdvLib \alma & 1.00 & 0.96 & 989 & 989 & 276.72 \\
 & FoolBox \fmn & 1.00 & 0.95 & 1,000 & 1,000 & 281.21 \\
 & Original \fmn & 1.00 & 0.95 & 1,000 & 1,000 & 275.98 \\
 & AdvLib \fmn & 1.00 & 0.95 & 1,000 & 1,000 & 276.7 \\
 & AdvLib \fab & 0.98 & 0.88 & 360 & 1,618 & 288.11 \\
 & FoolBox \ead & 1.00 & 0.86 & 482 & 245 & 96.5 \\
 & Original \fab & 0.96 & 0.85 & 303 & 1,505 & 268.29 \\
 & TorchAttacks \fab & 0.08 & 0.59 & 303 & 1,505 & 275.56 \\
 & Art \ead & 1.00 & 0.58 & 334 & 1,665 & 295.69 \\
 & Art \pgd & 0.24 & 0.55 & 1,010 & 980 & 429.67 \\
 & FoolBox \pgd & 1.00 & 0.52 & 1,000 & 990 & 714.96 \\
 & FoolBox \bim & 1.00 & 0.49 & 220 & 200 & 279.16 \\
 & Art \fgsm & 0.93 & 0.22 & 40 & 20 & 29.53 \\
 & FoolBox \fgsm & 0.93 & 0.22 & 40 & 20 & 27.51 \\
 & FoolBox \bb & 0.09 & 0.05 & 647 & 2 & 119.42 \\
 & Art \apgd & 0.94 & 0.03 & 1,098 & 482 & 303.17 \\ \cmidrule(l{0.5em}r{0.5em}){1-7}
\multirow{21}{*}{\stutz} & AdvLib \fmn & 0.99 & 0.94 & 1,000 & 1,000 & 38.85 \\
 & FoolBox \fmn & 0.99 & 0.94 & 1,000 & 1,000 & 39.04 \\
 & Original \fmn & 0.99 & 0.94 & 1,000 & 1,000 & 39.51 \\
 & AdvLib \pdpgd & 1.00 & 0.93 & 992 & 992 & 38.96 \\
 & AdvLib \fab & 0.97 & 0.90 & 361 & 1,620 & 38.85 \\
 & Original \fab & 0.97 & 0.89 & 295 & 1,465 & 33.77 \\
 & Original \apgdlone & 1.00 & 0.89 & 910 & 890 & 133.94 \\
 & AdvLib \alma & 1.00 & 0.87 & 987 & 987 & 39.49 \\
 & AdvLib \apgd & 1.00 & 0.72 & 1,000 & 990 & 44.81 \\
 & Art \pgd & 1.00 & 0.69 & 1,010 & 980 & 127.18 \\
 & Original \apgdt & 1.00 & 0.68 & 628 & 584 & 231.16 \\
 & FoolBox \bb & 0.81 & 0.66 & 567 & 176 & 51.69 \\
 & TorchAttacks \fab & 0.10 & 0.60 & 295 & 1,465 & 33.17 \\
 & FoolBox \pgd & 1.00 & 0.58 & 1,000 & 990 & 135.1 \\
 & Art \ead & 1.00 & 0.57 & 334 & 1,665 & 42.63 \\
 & FoolBox \bim & 1.00 & 0.56 & 220 & 200 & 137.15 \\
 & AdvLib \apgdt & 1.00 & 0.48 & 572 & 552 & 32.75 \\
 & FoolBox \ead & 1.00 & 0.48 & 389 & 199 & 10.93 \\
 & Art \apgd & 1.00 & 0.42 & 752 & 321 & 253.57 \\
 & Art \fgsm & 1.00 & 0.11 & 40 & 20 & 18.99 \\
 & FoolBox \fgsm & 1.00 & 0.11 & 40 & 20 & 17.54 \\ \bottomrule
\end{tabular}
\end{table*}

%% file: tables/appendix/ltwo-cifar.tex
\begin{table*}[htbp]
\small\centering
\setlength\tabcolsep{5pt}
\caption{Local optimality for $\ell_2$-norm attacks (pt1).}
\label{tab:ltwo_all_1}
\sisetup{table-auto-round}
\begin{tabular}{cl*{2}{S[table-format=3.1, fixed-exponent=-2, drop-exponent=true, exponent-mode=fixed, drop-zero-decimal]}ccS[table-format=4.1]}
\toprule\textbf{Model} & \textbf{Attack} & \textbf{ASR} & \textbf{Optimality} & \textbf{\#Forwards} & \textbf{\#Backwards} & \textbf{ExecTime} \\ \toprule
\multirow{42}{*}{\zhang} & FoolBox \ddn & 1 & 0.8869 & 999 & 999 & 3.6927 \\
 & AdvLib \ddn-NQ & 1 & 0.8866 & 999 & 999 & 3.5727 \\
 & AdvLib \ddn & 0.9932 & 0.8715 & 1000 & 1000 & 3.1607 \\
 & TorchAttacks \fab & 0.9223 & 0.8675 & 128 & 635 & 3.2069 \\
 & AdvLib \fab & 0.9896 & 0.8605 & 360 & 1618 & 3.5711 \\
 & AdvLib \fmn & 0.981 & 0.8515 & 998 & 998 & 3.4182 \\
 & Original \fab & 0.9816 & 0.8405 & 128 & 635 & 2.8992 \\
 & AdvLib \alma & 0.9995 & 0.8281 & 993 & 993 & 5.6367 \\
 & AdvLib \pdgd & 0.9497 & 0.823 & 1000 & 1000 & 3.377 \\
 & Original \apgd & 0.9999 & 0.8111 & 406 & 386 & 74.4073 \\
 & Original \apgdt & 1 & 0.7856 & 317 & 285 & 85.1092 \\
 & AdvLib \apgdt & 0.9982 & 0.7809 & 720 & 700 & 4.694 \\
 & Original \fmn & 0.9207 & 0.7474 & 1000 & 1000 & 4.5912 \\
 & FoolBox \bim & 0.9984 & 0.715 & 1000 & 990 & 62.9453 \\
 & AdvLib \apgd & 0.9997 & 0.7064 & 999 & 989 & 4.1065 \\
 & AdvLib \tr & 0.7786 & 0.7016 & 356 & 187 & 3.0319 \\
 & Original \tr & 0.7786 & 0.7016 & 841 & 187 & 3.6845 \\
 & TorchAttacks \pgd & 0.9991 & 0.6894 & 998 & 988 & 53.3644 \\
 & Art \pgd & 0.8728 & 0.6385 & 1010 & 980 & 50.4203 \\
 & DeepRobust \pgd & 0.6629 & 0.6269 & 1186 & 777 & 10.3979 \\
 & FoolBox \pgd & 0.9999 & 0.6061 & 999 & 989 & 61.5788 \\
 & TorchAttacks \cw & 0.7305 & 0.603 & 379 & 379 & 1.3869 \\
 & TorchAttacks \apgd & 0.61 & 0.5962 & 1 & 0 & 0.0195 \\
 & Art \apgd & 0.9992 & 0.5897 & 412 & 165 & 53.7246 \\
 & FoolBox \cw & 0.9957 & 0.5829 & 784 & 784 & 3.3601 \\
 & AdvLib \cw & 0.8806 & 0.5816 & 846 & 845 & 3.2319 \\
 & Original \deepfool & 0.839 & 0.5802 & 7 & 51 & 8.117 \\
 & FoolBox \deepfool & 0.9279 & 0.5762 & 998 & 997 & 3.2707 \\
 & AdvLib \pdpgd & 0.9908 & 0.5492 & 984 & 984 & 4.9793 \\
 & Original \tr & 0.777 & 0.4999 & 743 & 132 & 3.4808 \\
 & AdvLib \tr & 0.777 & 0.4999 & 370 & 132 & 2.6547 \\
 & Art \cw & 0.7153 & 0.4658 & 485 & 1514 & 9.833 \\
 & FoolBox \fgsm & 0.9817 & 0.4561 & 40 & 20 & 2.0855 \\
 & Cleverhans \fgsm & 0.9817 & 0.4561 & 41 & 20 & 2.0805 \\
 & Art \fgsm & 0.9817 & 0.4561 & 40 & 20 & 2.1947 \\
 & DeepRobust \fgsm & 0.976 & 0.4507 & 39 & 19 & 2.203 \\
 & Art \deepfool & 0.7875 & 0.4485 & 334 & 1666 & 29.0183 \\
 & Cleverhans \cw & 0.7634 & 0.4404 & 1000 & 1000 & 23.4063 \\
 & FoolBox \bb & 0.6109 & 0.4086 & 637 & 0 & 1.8536 \\
 & FoolBox \fmn & 0.9379 & 0.348 & 1000 & 1000 & 4.4161 \\
 & AdvLib \pgd & 0.995 & 0.1948 & 996 & 986 & 2.9746 \\
 & Art \bim & 0.9882 & 0.1361 & 833 & 807 & 34.1254 \\ \bottomrule
\end{tabular}
\end{table*}

\begin{table*}[htbp]
\small\centering
\setlength\tabcolsep{5pt}
\caption{Local optimality for $\ell_2$-norm attacks (pt2).}
\label{tab:ltwo_all_2}
\sisetup{table-auto-round}
\begin{tabular}{cl*{2}{S[table-format=3.1, fixed-exponent=-2, drop-exponent=true, exponent-mode=fixed, drop-zero-decimal]}ccS[table-format=4.1]}
\toprule\textbf{Model} & \textbf{Attack} & \textbf{ASR} & \textbf{Optimality} & \textbf{\#Forwards} & \textbf{\#Backwards} & \textbf{ExecTime} \\ \toprule
\multirow{42}{*}{\xiao} & AdvLib \apgd & 1 & 0.9576 & 999 & 989 & 45.7843 \\
 & Original \apgdt & 1 & 0.9391 & 509 & 468 & 398.5067 \\
 & Original \apgd & 1 & 0.928 & 668 & 648 & 406.5414 \\
 & AdvLib \apgdt & 0.9925 & 0.905 & 613 & 593 & 33.0274 \\
 & AdvLib \pdgd & 1 & 0.8396 & 997 & 997 & 46.1504 \\
 & Original \fmn & 1 & 0.8367 & 999 & 999 & 47.3604 \\
 & AdvLib \pdpgd & 1 & 0.8354 & 979 & 979 & 47.2508 \\
 & FoolBox \ddn & 1 & 0.8324 & 992 & 992 & 46.6346 \\
 & AdvLib \ddn & 1 & 0.8324 & 992 & 992 & 46.6727 \\
 & AdvLib \ddn-NQ & 1 & 0.8321 & 992 & 992 & 45.4467 \\
 & FoolBox \bim & 0.9983 & 0.822 & 995 & 985 & 413.0241 \\
 & AdvLib \fmn & 0.9996 & 0.8123 & 994 & 994 & 45.5755 \\
 & TorchAttacks \pgd & 0.9942 & 0.8031 & 996 & 986 & 421.1803 \\
 & AdvLib \cw & 0.9479 & 0.7902 & 921 & 920 & 42.6894 \\
 & AdvLib \tr & 0.5882 & 0.7342 & 612 & 315 & 26.5676 \\
 & Original \tr & 0.5878 & 0.7336 & 969 & 315 & 35.8658 \\
 & Art \pgd & 0.8469 & 0.7225 & 1010 & 980 & 423.0204 \\
 & FoolBox \pgd & 0.9973 & 0.7058 & 996 & 986 & 402.3861 \\
 & AdvLib \pgd & 0.993 & 0.7005 & 997 & 987 & 44.6087 \\
 & TorchAttacks \cw & 0.5143 & 0.6933 & 163 & 163 & 7.4648 \\
 & FoolBox \cw & 0.9925 & 0.6728 & 301 & 301 & 14.2661 \\
 & DeepRobust \pgd & 0.4819 & 0.6656 & 1184 & 776 & 95.7004 \\
 & Cleverhans \cw & 0.598 & 0.6116 & 1000 & 1000 & 65.5693 \\
 & Art \cw & 0.3685 & 0.5801 & 480 & 1519 & 53.2833 \\
 & FoolBox \fmn & 1 & 0.5792 & 438 & 438 & 47.048 \\
 & TorchAttacks \apgd & 0.3449 & 0.5609 & 1 & 0 & 0.0435 \\
 & Art \apgd & 0.9966 & 0.5283 & 708 & 300 & 291.1963 \\
 & TorchAttacks \fab & 0.3455 & 0.4997 & 85 & 453 & 32.6747 \\
 & AdvLib \tr & 0.5502 & 0.4476 & 732 & 253 & 27.5517 \\
 & Original \tr & 0.5514 & 0.4454 & 984 & 253 & 34.5724 \\
 & Art \bim & 0.9433 & 0.4042 & 720 & 696 & 218.9333 \\
 & FoolBox \fgsm & 0.9907 & 0.386 & 40 & 20 & 17.5344 \\
 & DeepRobust \fgsm & 0.9907 & 0.3859 & 40 & 20 & 19.2763 \\
 & Art \fgsm & 0.9906 & 0.3784 & 40 & 20 & 18.6498 \\
 & Cleverhans \fgsm & 0.9907 & 0.3783 & 41 & 20 & 17.8358 \\
 & AdvLib \alma & 1 & 0.3757 & 978 & 978 & 47.3374 \\
 & FoolBox \bb & 0.3463 & 0.0431 & 630 & 0 & 20.438 \\
 & Original \fab & 0.9364 & 0.0139 & 85 & 453 & 31.8946 \\
 & AdvLib \fab & 0.9821 & 0.0123 & 125 & 617 & 41.5891 \\
 & Original \deepfool & 0.9191 & 0.0119 & 4 & 20 & 19.4411 \\
 & FoolBox \deepfool & 1 & 0.0092 & 51 & 50 & 2.3078 \\
 & Art \deepfool & 0.9258 & 0.0078 & 12 & 63 & 1.7209 \\ \bottomrule
 \end{tabular}
 \end{table*}
 
  \begin{table*}[htbp]
\small\centering
\setlength\tabcolsep{5pt}
\caption{Local optimality for $\ell_2$-norm attacks (pt3).}
\label{tab:ltwo_all_3}
\sisetup{table-auto-round}
\begin{tabular}{cl*{2}{S[table-format=3.1, fixed-exponent=-2, drop-exponent=true, exponent-mode=fixed, drop-zero-decimal]}ccS[table-format=4.1]}
\toprule\textbf{Model} & \textbf{Attack} & \textbf{ASR} & \textbf{Optimality} & \textbf{\#Forwards} & \textbf{\#Backwards} & \textbf{ExecTime} \\ \toprule
\multirow{42}{*}{\stutz} & AdvLib \ddn-NQ & 1 & 0.9796 & 1000 & 1000 & 37.4055 \\
 & FoolBox \ddn & 1 & 0.9794 & 1000 & 1000 & 37.8202 \\
 & AdvLib \pdgd & 1 & 0.9771 & 991 & 991 & 38.475 \\
 & AdvLib \fab & 0.9999 & 0.9741 & 361 & 1620 & 38.8034 \\
 & Original \fab & 0.9947 & 0.9683 & 295 & 1465 & 32.9567 \\
 & TorchAttacks \fab & 0.9947 & 0.9683 & 295 & 1465 & 33.1802 \\
 & Original \apgd & 1 & 0.9655 & 910 & 890 & 61.856 \\
 & Art \apgd & 1 & 0.9572 & 557 & 236 & 49.2121 \\
 & Cleverhans \cw & 1 & 0.9519 & 1000 & 1000 & 57.4647 \\
 & FoolBox \bim & 1 & 0.9501 & 1000 & 990 & 64.653 \\
 & AdvLib \apgd & 1 & 0.9489 & 1000 & 990 & 37.9903 \\
 & Original \fmn & 0.9991 & 0.9391 & 1000 & 1000 & 38.5861 \\
 & AdvLib \fmn & 0.9991 & 0.937 & 1000 & 1000 & 38.0444 \\
 & Original \apgdt & 1 & 0.9195 & 514 & 474 & 64.572 \\
 & AdvLib \apgdt & 0.9864 & 0.9035 & 493 & 473 & 21.5488 \\
 & Art \cw & 0.9935 & 0.8856 & 478 & 1521 & 48.5609 \\
 & FoolBox \pgd & 1 & 0.873 & 1000 & 990 & 72.9501 \\
 & Art \pgd & 1 & 0.8662 & 1010 & 980 & 76.6566 \\
 & AdvLib \pdpgd & 1 & 0.8266 & 928 & 928 & 40.1371 \\
 & AdvLib \ddn & 1 & 0.8239 & 1000 & 1000 & 37.4725 \\
 & AdvLib \cw & 1 & 0.8126 & 999 & 999 & 38.6529 \\
 & AdvLib \tr & 1 & 0.796 & 92 & 55 & 4.7711 \\
 & Original \tr & 1 & 0.796 & 243 & 55 & 7.2187 \\
 & FoolBox \bb & 0.8164 & 0.7854 & 571 & 178 & 17.5174 \\
 & TorchAttacks \pgd & 1 & 0.7711 & 1000 & 990 & 92.5014 \\
 & FoolBox \cw & 1 & 0.7696 & 616 & 616 & 24.0965 \\
 & AdvLib \pgd & 1 & 0.7585 & 1000 & 990 & 37.4227 \\
 & Original \tr & 1 & 0.7404 & 68 & 23 & 2.1319 \\
 & AdvLib \tr & 1 & 0.7403 & 42 & 23 & 1.6804 \\
 & AdvLib \alma & 1 & 0.6247 & 997 & 997 & 39.1032 \\
 & FoolBox \fmn & 1 & 0.6243 & 730 & 730 & 39.4595 \\
 & TorchAttacks \apgd & 0.1008 & 0.6243 & 1 & 0 & 0.0354 \\
 & DeepRobust \pgd & 0.4144 & 0.4515 & 1183 & 774 & 83.5139 \\
 & TorchAttacks \cw & 1 & 0.4483 & 230 & 230 & 8.8744 \\
 & FoolBox \deepfool & 1 & 0.2707 & 91 & 90 & 3.3965 \\
 & Original \deepfool & 0.9551 & 0.2706 & 4 & 24 & 20.0079 \\
 & Art \fgsm & 0.9997 & 0.256 & 40 & 20 & 17.6911 \\
 & FoolBox \fgsm & 0.9997 & 0.256 & 40 & 20 & 16.8035 \\
 & DeepRobust \fgsm & 0.9997 & 0.256 & 40 & 20 & 17.9343 \\
 & Cleverhans \fgsm & 0.9997 & 0.256 & 41 & 20 & 17.0376 \\
 & Art \deepfool & 0.8062 & 0.1975 & 334 & 1666 & 66.6713 \\
 & Art \bim & 0.9315 & 0.0073 & 803 & 777 & 75.2022 \\ \bottomrule
 \end{tabular}
 \end{table*}

 \begin{table*}[htbp]
\small\centering
\setlength\tabcolsep{5pt}
\caption{Local optimality for $\ell_2$-norm attacks (pt4).}
\label{tab:ltwo_all_4}
\sisetup{table-auto-round}
\begin{tabular}{cl*{2}{S[table-format=3.1, fixed-exponent=-2, drop-exponent=true, exponent-mode=fixed, drop-zero-decimal]}ccS[table-format=4.1]}
\toprule\textbf{Model} & \textbf{Attack} & \textbf{ASR} & \textbf{Optimality} & \textbf{\#Forwards} & \textbf{\#Backwards} & \textbf{ExecTime} \\ \toprule
\multirow{42}{*}{\standardcifar} & Original \apgdt & 1 & 0.989 & 579 & 536 & 319.1984 \\
 & AdvLib \alma & 1 & 0.9887 & 999 & 999 & 262.2418 \\
 & AdvLib \apgdt & 0.9964 & 0.9867 & 524 & 504 & 136.9066 \\
 & AdvLib \ddn-NQ & 1 & 0.9802 & 1000 & 1000 & 261.5184 \\
 & FoolBox \ddn & 1 & 0.9802 & 1000 & 1000 & 262.3011 \\
 & Original \apgd & 1 & 0.979 & 958 & 938 & 396.6695 \\
 & AdvLib \pdgd & 1 & 0.9788 & 988 & 988 & 266.3426 \\
 & Original \fmn & 0.996 & 0.9755 & 1000 & 1000 & 267.5347 \\
 & AdvLib \fmn & 0.9956 & 0.9753 & 1000 & 1000 & 262.3283 \\
 & FoolBox \bim & 1 & 0.9744 & 1000 & 990 & 402.4092 \\
 & AdvLib \fab & 0.9996 & 0.9686 & 361 & 1620 & 270.1778 \\
 & Original \fab & 0.9997 & 0.9682 & 311 & 1544 & 255.1561 \\
 & TorchAttacks \fab & 0.9997 & 0.9682 & 311 & 1544 & 256.6718 \\
 & AdvLib \apgd & 1 & 0.9682 & 1000 & 990 & 263.092 \\
 & Art \pgd & 1 & 0.9607 & 1010 & 980 & 428.7291 \\
 & AdvLib \cw & 1 & 0.96 & 999 & 999 & 261.4782 \\
 & FoolBox \cw & 1 & 0.9582 & 780 & 780 & 209.5063 \\
 & Art \apgd & 1 & 0.9578 & 698 & 297 & 289.0286 \\
 & Cleverhans \cw & 1 & 0.955 & 1000 & 1000 & 282.1782 \\
 & TorchAttacks \pgd & 1 & 0.9385 & 1000 & 990 & 408.7166 \\
 & FoolBox \pgd & 1 & 0.9357 & 1000 & 990 & 433.9398 \\
 & AdvLib \ddn & 1 & 0.9331 & 1000 & 1000 & 259.8803 \\
 & AdvLib \pdpgd & 1 & 0.9007 & 940 & 940 & 265.2022 \\
 & AdvLib \tr & 0.9963 & 0.891 & 265 & 141 & 58.8974 \\
 & Original \tr & 0.9963 & 0.891 & 731 & 141 & 117.9695 \\
 & AdvLib \tr & 1 & 0.8763 & 54 & 27 & 11.0964 \\
 & Original \tr & 1 & 0.8763 & 77 & 27 & 13.7297 \\
 & Art \cw & 0.9949 & 0.8483 & 475 & 1524 & 284.2591 \\
 & AdvLib \pgd & 1 & 0.7847 & 1000 & 990 & 260.2534 \\
 & FoolBox \deepfool & 1 & 0.6722 & 81 & 80 & 21.2132 \\
 & Original \deepfool & 0.9629 & 0.6718 & 4 & 24 & 32.8758 \\
 & FoolBox \fmn & 1 & 0.6617 & 938 & 938 & 266.596 \\
 & TorchAttacks \apgd & 0.0522 & 0.6617 & 1 & 0 & 0.1346 \\
 & TorchAttacks \cw & 1 & 0.6525 & 201 & 201 & 52.8653 \\
 & DeepRobust \pgd & 0.2111 & 0.5139 & 1181 & 773 & 318.7292 \\
 & Art \deepfool & 0.7805 & 0.4785 & 334 & 1666 & 296.8559 \\
 & DeepRobust \fgsm & 0.9863 & 0.4554 & 40 & 20 & 31.642 \\
 & FoolBox \fgsm & 0.9863 & 0.4554 & 40 & 20 & 31.648 \\
 & Art \fgsm & 0.9862 & 0.4506 & 40 & 20 & 29.2661 \\
 & Cleverhans \fgsm & 0.9861 & 0.4426 & 41 & 20 & 27.6451 \\
 & Art \bim & 0.9301 & 0.3416 & 813 & 788 & 274.3021 \\
 & FoolBox \bb & 0.0537 & 0.2651 & 634 & 0 & 95.852 \\ \bottomrule
 \end{tabular}
 \end{table*}
 
 \begin{table*}[htbp]
\small\centering
\setlength\tabcolsep{5pt}
\caption{Local optimality for $\ell_2$-norm attacks (pt5).}
\label{tab:ltwo_all_5}
\sisetup{table-auto-round}
\begin{tabular}{cl*{2}{S[table-format=3.1, fixed-exponent=-2, drop-exponent=true, exponent-mode=fixed, drop-zero-decimal]}ccS[table-format=4.1]}
\toprule\textbf{Model} & \textbf{Attack} & \textbf{ASR} & \textbf{Optimality} & \textbf{\#Forwards} & \textbf{\#Backwards} & \textbf{ExecTime} \\ \toprule
\multirow{42}{*}{\wang} & AdvLib \alma & 1 & 0.9839 & 999 & 999 & 280.6575 \\
 & Original \apgdt & 1 & 0.9754 & 693 & 645 & 641.8217 \\
 & AdvLib \apgdt & 0.9989 & 0.9684 & 641 & 621 & 183.6234 \\
 & AdvLib \pdgd & 1 & 0.9677 & 996 & 996 & 279.5777 \\
 & FoolBox \ddn & 1 & 0.9658 & 999 & 999 & 278.315 \\
 & AdvLib \ddn-NQ & 1 & 0.9658 & 999 & 999 & 278.0119 \\
 & AdvLib \fmn & 1 & 0.9642 & 999 & 999 & 275.3351 \\
 & Original \fmn & 1 & 0.9637 & 1000 & 1000 & 280.4277 \\
 & AdvLib \ddn & 1 & 0.9628 & 999 & 999 & 280.2926 \\
 & Original \apgd & 1 & 0.9602 & 935 & 915 & 709.2426 \\
 & AdvLib \apgd & 1 & 0.9567 & 1000 & 990 & 277.2928 \\
 & AdvLib \fab & 1 & 0.9556 & 360 & 1619 & 287.5661 \\
 & Cleverhans \cw & 0.9919 & 0.9553 & 1000 & 1000 & 298.7841 \\
 & FoolBox \bim & 1 & 0.9527 & 1000 & 990 & 707.8158 \\
 & TorchAttacks \pgd & 1 & 0.9486 & 1000 & 990 & 812.2736 \\
 & FoolBox \cw & 1 & 0.9438 & 771 & 771 & 215.04 \\
 & AdvLib \cw & 0.9991 & 0.9435 & 801 & 799 & 222.1641 \\
 & Original \fab & 0.9841 & 0.9361 & 304 & 1506 & 271.7352 \\
 & Art \pgd & 0.9988 & 0.9332 & 1010 & 980 & 764.9297 \\
 & FoolBox \pgd & 1 & 0.8639 & 1000 & 990 & 728.9306 \\
 & TorchAttacks \fab & 0.6187 & 0.8494 & 304 & 1506 & 269.8616 \\
 & TorchAttacks \cw & 0.7294 & 0.8437 & 553 & 553 & 151.9668 \\
 & Original \tr & 0.3751 & 0.7472 & 1190 & 536 & 235.7945 \\
 & AdvLib \tr & 0.3751 & 0.7472 & 1053 & 536 & 221.0055 \\
 & TorchAttacks \apgd & 0.0756 & 0.7091 & 1 & 0 & 0.1319 \\
 & Art \cw & 0.6122 & 0.6875 & 476 & 1523 & 307.3343 \\
 & AdvLib \pdpgd & 1 & 0.685 & 979 & 979 & 281.4783 \\
 & DeepRobust \pgd & 0.0945 & 0.6757 & 1005 & 657 & 261.4744 \\
 & Original \tr & 0.3864 & 0.5907 & 1201 & 361 & 207.8493 \\
 & AdvLib \tr & 0.3863 & 0.5906 & 1056 & 361 & 198.4336 \\
 & Art \apgd & 0.9931 & 0.5425 & 945 & 411 & 583.0781 \\
 & FoolBox \deepfool & 1 & 0.4998 & 57 & 56 & 15.6782 \\
 & Original \deepfool & 0.949 & 0.4995 & 3 & 19 & 24.6384 \\
 & Art \deepfool & 0.9454 & 0.4813 & 329 & 1643 & 317.815 \\
 & FoolBox \fmn & 1 & 0.4804 & 809 & 809 & 276.9579 \\
 & Cleverhans \fgsm & 0.9233 & 0.3598 & 41 & 20 & 28.1363 \\
 & Art \fgsm & 0.9233 & 0.3597 & 40 & 20 & 29.8348 \\
 & FoolBox \fgsm & 0.9233 & 0.3588 & 40 & 20 & 27.9462 \\
 & DeepRobust \fgsm & 0.9233 & 0.3586 & 40 & 20 & 29.4463 \\
 & AdvLib \pgd & 1 & 0.294 & 999 & 989 & 274.2667 \\
 & Art \bim & 0.9912 & 0.2384 & 869 & 841 & 322.2215 \\
 & FoolBox \bb & 0.0862 & 0.0428 & 646 & 2 & 112.0969 \\ \bottomrule
\end{tabular}
\end{table*}

%% file: tables/appendix/linf-cifar.tex
\begin{table*}[htbp]
\footnotesize\centering
\setlength\tabcolsep{10pt}
\caption{Local optimality for $\ell_\infty$-norm attacks.}
\label{tab:linf_all_1}
\sisetup{table-auto-round}
\begin{tabular}{cl*{2}{S[table-format=3.1, fixed-exponent=-2, drop-exponent=true, exponent-mode=fixed, drop-zero-decimal]}ccS[table-format=4.1]}
\toprule\textbf{Model} & \textbf{Attack} & \textbf{ASR} & \textbf{Optimality} & \textbf{\#Forwards} & \textbf{\#Backwards} & \textbf{ExecTime} \\ \toprule
\multirow{33}{*}{\xiao} & AdvLib \apgd & 1 & 0.9577 & 1000 & 990 & 45.60903595 \\
 & Original \apgdt & 1 & 0.9472 & 572 & 529 & 295.424865 \\
 & Original \apgd & 1 & 0.9316 & 668 & 648 & 306.0540897 \\
 & AdvLib \apgdt & 0.9855 & 0.927 & 723 & 703 & 37.31342022 \\
 & AdvLib \pdpgd & 1 & 0.8671 & 979 & 979 & 50.74953581 \\
 & Art \pgd & 0.9949 & 0.8655 & 1006 & 976 & 272.3489933 \\
 & FoolBox \bim & 0.9966 & 0.8353 & 996 & 986 & 273.6509239 \\
 & AdvLib \pgd & 1 & 0.8186 & 999 & 989 & 44.54171901 \\
 & DeppRobust \pgd & 0.9945 & 0.8174 & 1007 & 987 & 255.5108982 \\
 & AdvLib \cw & 0.9975 & 0.8141 & 594 & 594 & 36.82547415 \\
 & FoolBox\fmn& 1 & 0.8114 & 996 & 996 & 46.24670772 \\
 & Original \fmn & 1 & 0.8114 & 996 & 996 & 46.41489238 \\
 & AdvLib \fmn & 1 & 0.8106 & 996 & 996 & 45.49834233 \\
 & TorchAttacks \pgd & 0.9939 & 0.8093 & 996 & 986 & 253.3254305 \\
 & FoolBox \pgd & 0.9962 & 0.7604 & 997 & 987 & 271.0859351 \\
 & Art \apgd & 0.8895 & 0.7595 & 902 & 434 & 184.9918291 \\
 & Original \tr & 0.9931 & 0.6415 & 944 & 104 & 35.62499423 \\
 & AdvLib \tr & 0.993 & 0.6414 & 191 & 104 & 18.05543258 \\
 & Original \tr & 0.9697 & 0.6216 & 1127 & 89 & 42.19395773 \\
 & AdvLib \tr & 0.9694 & 0.6211 & 239 & 89 & 20.85629815 \\
 & Art \cw & 0.6344 & 0.5794 & 1324 & 645 & 1893.669929 \\
 & Art \bb & 0.3449 & 0.5639 & 13 & 0 & 12.10203703 \\
 & TorchAttacks \apgd & 0.3449 & 0.5639 & 1 & 0 & 0.03618708242 \\
 & TorchAttacks \fgsm & 0.9739 & 0.5183 & 40 & 20 & 1.420788827 \\
 & Art \fgsm & 0.974 & 0.5183 & 40 & 20 & 16.33913708 \\
 & Cleverhans \fgsm & 0.974 & 0.5183 & 41 & 20 & 14.45655176 \\
 & DeppRobust \fgsm & 0.974 & 0.5183 & 40 & 20 & 18.35881087 \\
 & FoolBox \fgsm & 0.974 & 0.5183 & 40 & 20 & 15.57287393 \\
 & FoolBox \bb & 0.4276 & 0.0661 & 717 & 0 & 26.06701577 \\
 & TorchAttacks \fab & 0.3507 & 0.0434 & 80 & 501 & 32.64473853 \\
 & FoolBox \deepfool & 0.9179 & 0.0094 & 396 & 394 & 64.13559374 \\
 & Original \fab & 0.9739 & 0.0076 & 80 & 501 & 32.9417186 \\
 & AdvLib \fab & 0.9988 & 0.0057 & 126 & 726 & 41.98098306 \\
\bottomrule
\end{tabular}
\end{table*}

 \begin{table*}[htbp]
\footnotesize\centering
\setlength\tabcolsep{10pt}
\caption{Local optimality for $\ell_\infty$-norm attacks (pt2).}
\label{tab:linf_all_2}
\sisetup{table-auto-round}
\begin{tabular}{cl*{2}{S[table-format=3.1, fixed-exponent=-2, drop-exponent=true, exponent-mode=fixed, drop-zero-decimal]}ccS[table-format=4.1]}
\toprule\textbf{Model} & \textbf{Attack} & \textbf{ASR} & \textbf{Optimality} & \textbf{\#Forwards} & \textbf{\#Backwards} & \textbf{ExecTime} \\ \toprule
\multirow{33}{*}{\zhang} & Original \apgd & 1 & 0.9777 & 406 & 386 & 34.17364362 \\
 & Original \apgdt & 1 & 0.9763 & 373 & 338 & 40.9238866 \\
 & AdvLib \fab & 0.999 & 0.9663 & 362 & 1627 & 4.019505774 \\
 & AdvLib \apgdt & 0.9978 & 0.966 & 930 & 910 & 5.038323688 \\
 & TorchAttacks \fab & 0.9873 & 0.9548 & 131 & 647 & 3.20221179 \\
 & Original \fab & 0.9931 & 0.9535 & 131 & 647 & 3.140642734 \\
 & FoolBox \bim & 0.9998 & 0.9506 & 999 & 989 & 28.55142492 \\
 & Art \pgd & 0.9997 & 0.9499 & 1009 & 979 & 35.06045258 \\
 & DeppRobust \pgd & 0.9999 & 0.9472 & 1009 & 989 & 30.80463352 \\
 & TorchAttacks \pgd & 0.9999 & 0.9445 & 999 & 989 & 29.47137476 \\
 & AdvLib \pgd & 1 & 0.9422 & 999 & 989 & 2.84793673 \\
 & FoolBox \pgd & 1 & 0.9339 & 999 & 989 & 28.33815072 \\
 & AdvLib \apgd & 1 & 0.9083 & 999 & 989 & 4.364918633 \\
 & AdvLib \pdpgd & 0.9913 & 0.8836 & 890 & 890 & 8.543939697 \\
 & Original \tr & 0.9963 & 0.8806 & 689 & 55 & 3.060740377 \\
 & AdvLib \tr & 0.9963 & 0.8806 & 92 & 55 & 2.769245939 \\
 & Original \tr & 0.9865 & 0.8664 & 901 & 32 & 4.847534236 \\
 & AdvLib \tr & 0.9865 & 0.8664 & 70 & 32 & 3.448026721 \\
 & Art \fgsm & 0.9896 & 0.8408 & 40 & 20 & 2.011855843 \\
 & Cleverhans \fgsm & 0.9896 & 0.8408 & 41 & 20 & 1.719804761 \\
 & DeppRobust \fgsm & 0.9896 & 0.8408 & 40 & 20 & 2.051072473 \\
 & FoolBox \fgsm & 0.9896 & 0.8408 & 40 & 20 & 1.760548855 \\
 & TorchAttacks \fgsm & 0.9896 & 0.8408 & 40 & 20 & 0.108176486 \\
 & FoolBox\fmn& 0.973 & 0.8003 & 984 & 984 & 4.852477346 \\
 & Original \fmn & 0.973 & 0.8003 & 984 & 984 & 4.847741032 \\
 & AdvLib \fmn & 0.973 & 0.8003 & 984 & 984 & 3.702594381 \\
 & FoolBox \deepfool & 0.9986 & 0.7954 & 101 & 100 & 0.4710133579 \\
 & AdvLib \cw & 0.9588 & 0.7696 & 402 & 402 & 5.580304638 \\
 & Art \cw & 0.7846 & 0.6921 & 1311 & 636 & 392.2011165 \\
 & Art \apgd & 0.9507 & 0.6204 & 555 & 264 & 50.8354446 \\
 & Art \bb & 0.61 & 0.5915 & 13 & 0 & 2.213246743 \\
 & TorchAttacks \apgd & 0.61 & 0.5915 & 1 & 0 & 0.01055272058 \\
 & FoolBox \bb & 0.6109 & 0.393 & 743 & 0 & 2.182113868 \\
\bottomrule
\end{tabular}
\end{table*}
 
 \begin{table*}[htbp]
\footnotesize\centering
\setlength\tabcolsep{10pt}
\caption{Local optimality for $\ell_\infty$-norm attacks (pt3).}
\label{tab:lone_all_3}
\sisetup{table-auto-round}
\begin{tabular}{cl*{2}{S[table-format=3.1, fixed-exponent=-2, drop-exponent=true, exponent-mode=fixed, drop-zero-decimal]}ccS[table-format=4.1]}
\toprule\textbf{Model} & \textbf{Attack} & \textbf{ASR} & \textbf{Optimality} & \textbf{\#Forwards} & \textbf{\#Backwards} & \textbf{ExecTime} \\ \toprule
\multirow{33}{*}{\stutz} & Original \apgd & 1 & 0.9898 & 910 & 890 & 197.708097 \\
 & AdvLib \fab & 1 & 0.9835 & 363 & 1629 & 39.22042385 \\
 & FoolBox \bim & 1 & 0.9828 & 1000 & 990 & 210.8213104 \\
 & Original \fab & 0.9988 & 0.9813 & 301 & 1492 & 33.61925404 \\
 & TorchAttacks \fab & 0.9988 & 0.9813 & 301 & 1492 & 33.99166793 \\
 & AdvLib \apgd & 1 & 0.9756 & 1000 & 990 & 37.83721955 \\
 & AdvLib \pgd & 1 & 0.9731 & 1000 & 990 & 37.07504553 \\
 & FoolBox \pgd & 1 & 0.965 & 1000 & 990 & 202.5737616 \\
 & Original \apgdt & 1 & 0.9623 & 680 & 633 & 260.5282985 \\
 & AdvLib \apgdt & 0.9899 & 0.9545 & 644 & 624 & 27.93517878 \\
 & AdvLib \fmn & 1 & 0.9508 & 1000 & 1000 & 37.71431865 \\
 & FoolBox\fmn& 1 & 0.9505 & 1000 & 1000 & 38.43173173 \\
 & Original \fmn & 1 & 0.9505 & 1000 & 1000 & 38.49708351 \\
 & Art \apgd & 1 & 0.9259 & 1212 & 590 & 122.7880404 \\
 & FoolBox \bb & 0.9203 & 0.9072 & 1038 & 634 & 132.9615622 \\
 & AdvLib \pdpgd & 1 & 0.8779 & 872 & 872 & 42.62813491 \\
 & DeppRobust \pgd & 1 & 0.8536 & 1010 & 990 & 319.5648904 \\
 & Art \pgd & 1 & 0.8534 & 1010 & 980 & 326.1730099 \\
 & TorchAttacks \pgd & 1 & 0.8511 & 1000 & 990 & 284.606721 \\
 & Original \tr & 1 & 0.7758 & 13 & 12 & 0.544000877 \\
 & AdvLib \tr & 1 & 0.7758 & 8 & 12 & 0.4670626291 \\
 & Original \tr & 1 & 0.7747 & 10 & 12 & 0.4794204197 \\
 & AdvLib \tr & 1 & 0.7747 & 6 & 12 & 0.4039631172 \\
 & TorchAttacks \fgsm & 1 & 0.6213 & 40 & 20 & 1.090870292 \\
 & Art \fgsm & 1 & 0.6213 & 40 & 20 & 17.99850274 \\
 & Cleverhans \fgsm & 1 & 0.6213 & 41 & 20 & 17.30247086 \\
 & DeppRobust \fgsm & 1 & 0.6213 & 40 & 20 & 18.39927492 \\
 & FoolBox \fgsm & 1 & 0.6213 & 40 & 20 & 16.88673786 \\
 & Art \cw & 0.997 & 0.6166 & 1291 & 622 & 1393.188229 \\
 & TorchAttacks \apgd & 0.1008 & 0.6166 & 1 & 0 & 0.02830320366 \\
 & FoolBox \deepfool & 1 & 0.2696 & 37 & 36 & 1.374453488 \\
 & Art \bb & 0.1368 & 0.0066 & 20 & 3 & 11.18960129 \\
 & AdvLib \cw & 1 & 0.0061 & 307 & 306 & 17.62803388 \\
\bottomrule
\end{tabular}
\end{table*}
 
 \begin{table*}[htbp]
\footnotesize\centering
\setlength\tabcolsep{10pt}
\caption{Local optimality for $\ell_\infty$-norm attacks (pt4).}
\label{tab:lone_all_4}
\sisetup{table-auto-round}
\begin{tabular}{cl*{2}{S[table-format=3.1, fixed-exponent=-2, drop-exponent=true, exponent-mode=fixed, drop-zero-decimal]}ccS[table-format=4.1]}
\toprule\textbf{Model} & \textbf{Attack} & \textbf{ASR} & \textbf{Optimality} & \textbf{\#Forwards} & \textbf{\#Backwards} & \textbf{ExecTime} \\ \toprule
\multirow{33}{*}{\standardcifar} & Original \apgdt & 1 & 0.9953 & 707 & 659 & 626.1366574 \\
 & AdvLib \apgdt & 0.9967 & 0.994 & 641 & 621 & 169.0567646 \\
 & Original \apgd & 1 & 0.99 & 958 & 938 & 711.4984119 \\
 & FoolBox \bim & 1 & 0.9838 & 1000 & 990 & 692.2857321 \\
 & AdvLib \apgd & 1 & 0.9789 & 1000 & 990 & 263.7414422 \\
 & AdvLib \fab & 0.9999 & 0.977 & 363 & 1629 & 268.0842136 \\
 & Original \fab & 0.9999 & 0.9767 & 317 & 1573 & 262.0676951 \\
 & TorchAttacks \fab & 0.9999 & 0.9767 & 317 & 1573 & 257.8745368 \\
 & FoolBox\fmn& 1 & 0.9695 & 1000 & 1000 & 262.798728 \\
 & Original \fmn & 1 & 0.9695 & 1000 & 1000 & 264.4311096 \\
 & AdvLib \fmn & 1 & 0.9694 & 1000 & 1000 & 271.1564848 \\
 & AdvLib \pgd & 1 & 0.9686 & 1000 & 990 & 260.8865639 \\
 & AdvLib \pdpgd & 1 & 0.9635 & 970 & 970 & 269.7126066 \\
 & FoolBox \pgd & 1 & 0.9543 & 1000 & 990 & 727.3190633 \\
 & Art \apgd & 0.9966 & 0.9321 & 1274 & 623 & 389.9829526 \\
 & TorchAttacks \pgd & 1 & 0.8859 & 1000 & 990 & 696.4475099 \\
 & DeppRobust \pgd & 1 & 0.8603 & 1010 & 990 & 707.046313 \\
 & AdvLib \tr & 1 & 0.8493 & 10 & 14 & 3.474250101 \\
 & Original \tr & 1 & 0.8493 & 24 & 14 & 5.234910014 \\
 & AdvLib \tr & 1 & 0.8381 & 12 & 13 & 3.53266671 \\
 & Original \tr & 1 & 0.8381 & 19 & 13 & 4.430017738 \\
 & Art \pgd & 1 & 0.7743 & 1010 & 980 & 798.572302 \\
 & FoolBox \deepfool & 1 & 0.6697 & 48 & 47 & 12.91581292 \\
 & TorchAttacks \apgd & 0.0522 & 0.6601 & 1 & 0 & 0.1566745683 \\
 & AdvLib \cw & 1 & 0.6101 & 758 & 756 & 206.5772174 \\
 & TorchAttacks \fgsm & 0.9979 & 0.562 & 40 & 20 & 7.760535824 \\
 & Art \fgsm & 0.9979 & 0.562 & 40 & 20 & 31.69086154 \\
 & Cleverhans \fgsm & 0.9979 & 0.562 & 41 & 20 & 30.5000193 \\
 & DeppRobust \fgsm & 0.9979 & 0.562 & 40 & 20 & 31.92062016 \\
 & FoolBox \fgsm & 0.9979 & 0.562 & 40 & 20 & 30.21744849 \\
 & Art \cw & 1 & 0.4233 & 1346 & 651 & 2269.622471 \\
 & FoolBox \bb & 0.0773 & 0.0391 & 751 & 17 & 110.9280726 \\
 & Art \bb & 0.1032 & 0.005 & 24 & 5 & 29.03957141 \\
\bottomrule
\end{tabular}
\end{table*}

 \begin{table*}[htbp]
\footnotesize\centering
\setlength\tabcolsep{10pt}
\caption{Local optimality for $\ell_\infty$-norm attacks (pt5).}
\label{tab:lone_all_5}
\sisetup{table-auto-round}
\begin{tabular}{cl*{2}{S[table-format=3.1, fixed-exponent=-2, drop-exponent=true, exponent-mode=fixed, drop-zero-decimal]}ccS[table-format=4.1]}
\toprule\textbf{Model} & \textbf{Attack} & \textbf{ASR} & \textbf{Optimality} & \textbf{\#Forwards} & \textbf{\#Backwards} & \textbf{ExecTime} \\ \toprule
\multirow{33}{*}{\wang} & Original \apgdt & 1 & 0.998 & 813 & 760 & 416.0199557 \\
 & AdvLib \apgdt & 0.9982 & 0.9969 & 783 & 763 & 222.9686551 \\
 & AdvLib \apgd & 1 & 0.9872 & 1000 & 990 & 278.2009331 \\
 & FoolBox\fmn& 1 & 0.9868 & 999 & 999 & 279.7639111 \\
 & Original \fmn & 1 & 0.9868 & 999 & 999 & 277.2227874 \\
 & AdvLib \fmn & 1 & 0.9868 & 999 & 999 & 275.2100922 \\
 & AdvLib \fab & 1 & 0.984 & 362 & 1628 & 289.1614056 \\
 & Original \apgd & 1 & 0.9839 & 935 & 915 & 446.5897341 \\
 & FoolBox \bim & 1 & 0.9798 & 1000 & 990 & 451.0010057 \\
 & DeppRobust \pgd & 1 & 0.9794 & 1010 & 990 & 453.5602451 \\
 & TorchAttacks \pgd & 1 & 0.9781 & 1000 & 990 & 449.0419818 \\
 & Art \pgd & 1 & 0.9683 & 1010 & 980 & 474.3129772 \\
 & Original \fab & 0.9832 & 0.9657 & 309 & 1534 & 273.47272 \\
 & TorchAttacks \fab & 0.9832 & 0.9657 & 309 & 1534 & 274.8412449 \\
 & AdvLib \pgd & 1 & 0.9591 & 1000 & 990 & 281.7804681 \\
 & AdvLib \cw & 1 & 0.949 & 911 & 911 & 264.5508421 \\
 & AdvLib \pdpgd & 1 & 0.9486 & 899 & 899 & 284.6375123 \\
 & FoolBox \pgd & 1 & 0.9455 & 1000 & 990 & 450.379862 \\
 & AdvLib \tr & 1 & 0.9209 & 108 & 63 & 26.63460432 \\
 & Original \tr & 1 & 0.9209 & 279 & 63 & 47.75373745 \\
 & AdvLib \tr & 1 & 0.9142 & 42 & 23 & 9.651357454 \\
 & Original \tr & 1 & 0.9142 & 68 & 23 & 12.69500252 \\
 & Art \cw & 0.8948 & 0.8113 & 1331 & 648 & 2314.353344 \\
 & Art \apgd & 0.8883 & 0.6353 & 1243 & 611 & 367.1050834 \\
 & Art \bb & 0.0756 & 0.6319 & 21 & 0 & 20.55260322 \\
 & TorchAttacks \apgd & 0.0756 & 0.6319 & 1 & 0 & 0.1526896127 \\
 & TorchAttacks \fgsm & 0.9167 & 0.6031 & 40 & 20 & 7.927395867 \\
 & Art \fgsm & 0.9167 & 0.6031 & 40 & 20 & 25.04086819 \\
 & Cleverhans \fgsm & 0.9167 & 0.6031 & 41 & 20 & 24.13534286 \\
 & DeppRobust \fgsm & 0.9167 & 0.6031 & 40 & 20 & 24.94051932 \\
 & FoolBox \fgsm & 0.9167 & 0.6031 & 40 & 20 & 23.83053865 \\
 & FoolBox \deepfool & 1 & 0.597 & 62 & 61 & 17.35682661 \\
 & FoolBox \bb & 0.1091 & 0.1928 & 782 & 23 & 139.0334847 \\ \bottomrule
\end{tabular}
\end{table*}